\documentclass[serif]{wiley-article}
\usepackage[super]{natbib}
\makeatletter
\renewcommand{\@biblabel}[1]{#1.}
\makeatother

\usepackage{siunitx}
\usepackage{moreverb}
\usepackage{amsmath} 
\usepackage{algorithm,algpseudocode}
\usepackage{mathtools}
\usepackage{subcaption}
\usepackage{graphicx}
\usepackage{adjustbox}
\usepackage{tabularx}
\usepackage[hidelinks]{hyperref}

\title{Interpretable Machine Learning for Survival Analysis}

\author[1,2]{Sophie Hanna Langbein}
\author[4]{Mateusz Krzyziński}
\author[4]{Mikołaj Spytek}
\author[4,5]{Hubert Baniecki}
\author[4,5]{Przemysław Biecek}
\author[1,2,3]{Marvin N. Wright}

\affil[1]{Leibniz Institute for Prevention Research and Epidemiology – BIPS, Bremen, Germany}
\affil[2]{Faculty of Mathematics and Computer Science, University of Bremen, Germany}
\affil[3]{Department of Public Health, University of Copenhagen, Denmark}
\affil[4]{Faculty of Mathematics and Information Science, Warsaw University of Technology, Poland}
\affil[5]{Faculty of Mathematics, Informatics and Mechanics, University of Warsaw, Poland}

\corraddress{Marvin N. Wright, Leibniz Institute for Prevention Research and Epidemiology – BIPS, Bremen, Germany}
\corremail{wright@leibniz-bips.de}

\fundinginfo{German Research Foundation (DFG), Grant Numbers: 437611051, 459360854; Polish National Science Centre, Grant Number: 2019/34/E/ST6/00052;
Polish Ministry of Education and Science, Project Number: PN/01/0087/2022}

\runningauthor{Sophie Hanna Langbein et al.}

\begin{document}
\vspace*{-2cm}
\begin{frontmatter}
\maketitle

\begin{abstract}
With the spread and rapid advancement of black box machine learning models, the field of interpretable machine learning (IML) or explainable artificial intelligence (XAI) has become increasingly important over the last decade. This is particularly relevant for survival analysis, where the adoption of IML techniques promotes transparency, accountability and fairness in sensitive areas, such as clinical decision making processes, the development of targeted therapies, interventions or in other medical or healthcare related contexts. More specifically, explainability can uncover a survival model's potential biases and limitations and provide more mathematically sound ways to understand how and which features are influential for prediction or constitute risk factors. However, the lack of readily available IML methods may have deterred practitioners from leveraging the full potential of machine learning for predicting time-to-event data. We present a comprehensive review of the existing work on IML methods for survival analysis within the context of the general IML taxonomy. In addition, we formally detail how commonly used IML methods, such as individual conditional expectation (ICE), partial dependence plots (PDP), accumulated local effects (ALE), different feature importance measures or Friedman’s H-interaction statistics can be adapted to survival outcomes.  An application of several IML methods to data on breast cancer recurrence in the German Breast Cancer Study Group (GBSG2) serves as a tutorial or guide for researchers, on how to utilize the techniques in practice to facilitate understanding of model decisions or predictions.

\end{abstract}
\end{frontmatter}

\section{Introduction}\label{introduction}
Survival analysis is a statistical subfield which analyzes and interprets time-to-event data, i.e., examining the duration until an event of interest (e.g., death, failure) occurs in a study population, while considering the effects of censoring. Censoring refers to the event of interest not being observed for some subjects before the study is terminated. For an extensive period of time, classical parametric and semi-parametric statistical methods, exemplified by the Cox proportional hazards (CoxPH) model\cite{cox1972}, have held a pervasive influence over the field of survival analysis. This dominance is largely attributable to their ability to handle censored data and yield reliable inferences in the form of hazard ratios. Hazard ratios are widely considered interpretable, clarifying underlying event dynamics, rendering the CoxPH model an indispensable tool in diverse scientific domains, including epidemiology, medicine, and engineering. However, in practice, the strong assumptions of the CoxPH model are often not met, with the noncollapsibility problem potentially leading to biased crude hazard ratio estimates.\cite{martinussen2013,rulli2018} In recent years, the ease of interpretability of the hazard ratio, which is readily assumed by a majority of domains regularly applying the CoxPH model, has been called into question, even under the condition that all basic CoxPH model assumptions are met.\cite{martinussen2022} Challenges of classical statistical methods such as the CoxPH model are amplified by the rapid development of big data technologies, leading to the collection of a wide variety and volume of survival data in many different disciplines.\cite{herrmann2021,wiegrebe2024} In this light, machine learning models proof a viable alternative to classic statistical approaches, due to their enhanced prediction performance and flexibility, especially when using high-dimensional, large datasets.\cite{wang2019} Popular machine learning survival models include survival tree ensemble methods\cite{ishwaran2008,hothorn2006}, such as random survival forests and boosted trees, and survival neural networks\cite{katzman2018,lee2018,zhao2020}.

While there is considerable promise in using machine learning models for survival analysis, their inherent opacity has prompted legitimate concerns, since in fields such as the life sciences, results and model outputs must be interpretable to provide a solid basis for highly sensitive decision-making.\cite{rahman2022,vellido2020} The lack of interpretability of these so-called black box models\cite{rudin2019} may render it problematic to downright impossible to identify and understand the influence of specific risk factors on survival. There are further increasingly high regulatory standards for decisions made by artificial intelligence in a healthcare or biomedical context, which survival applications predominantly focus on.\cite{panigutti2023} For instance, the European General Data Protection Regulation (GDPR), emphasizes the importance of transparency and accountability in the processing of personal data, in particular granting patients the right to understand how and why certain diagnoses or recommendations are made by artificial intelligence.\cite{vellido2020} Moreover, biases in survival data are easily concealed, even though they may adversely impact the analysis and decision making especially in medical domains, which may unfairly affect the patients in minority groups.\cite{chen2019,manrai2016,baniecki2023hospital} Specifically, for common survival analysis data, models, and methods, recent papers have proven the existence of unintentional biases toward protected attributes such as age, race, gender, and/or ethnicity.\cite{mhasawade2021,rajkomar2018,gianfrancesco2018} Research has shown that censoring is oftentimes not independent of protected attributes (e.g. marginalized minority groups often experience high exposure to censoring), potentially leading to an overestimation of survival predictions.\cite{barrajon2020,bland2004,schulman1999} These issues can potentially be addressed by means of interpretable machine learning or explainable artificial intelligence methods.\cite{krzyzinski2023shap,molnar2022}

So far, there are few interpretable machine learning methods applicable to or specifically developed for complex survival models. In general, the taxonomy of explanation methods as established by Molnar\cite{molnar2022} and Biecek and Burzykowski\cite{biecek2021} divides interpretable machine learning methods into six different groups. The most important differentiation is between local and global methods, model-specific and model-agnostic methods, and post-hoc methods and ante-hoc interpretable models. In their most basic form, the most important local model-agnostic interpretability methods have been adapted to survival analysis, this includes \emph{SurvLIME}\cite{kovalev2020}, \emph{SurvSHAP}\cite{alabdallah2022} and \emph{SurvSHAP(t)}\cite{krzyzinski2023shap} and survival counterfactual explanations\cite{kovalev2021c}. Moreover, some specialized model-agnostic approaches, like \emph{SurvNAM}\cite{utkin2022} or \emph{UncSurvEx}\cite{utkin2021unc} have been developed, parallel to a set of interpretable machine learning survival models\cite{rahman2021,xu2023,hao2019PAGE,hao2019CoxPAS}. While there are some reviews for explainable artificial intelligence in healthcare\cite{bharati2023,srinivasu2022,nazar2021}, there is no comprehensive review for IML in survival analysis. Moreover, many foundational interpretable machine learning techniques, widely utilized in traditional settings have not yet been formally adapted to survival analysis.

\begin{figure}[htb!]
  \centering
  \includegraphics[width = 0.8\linewidth]{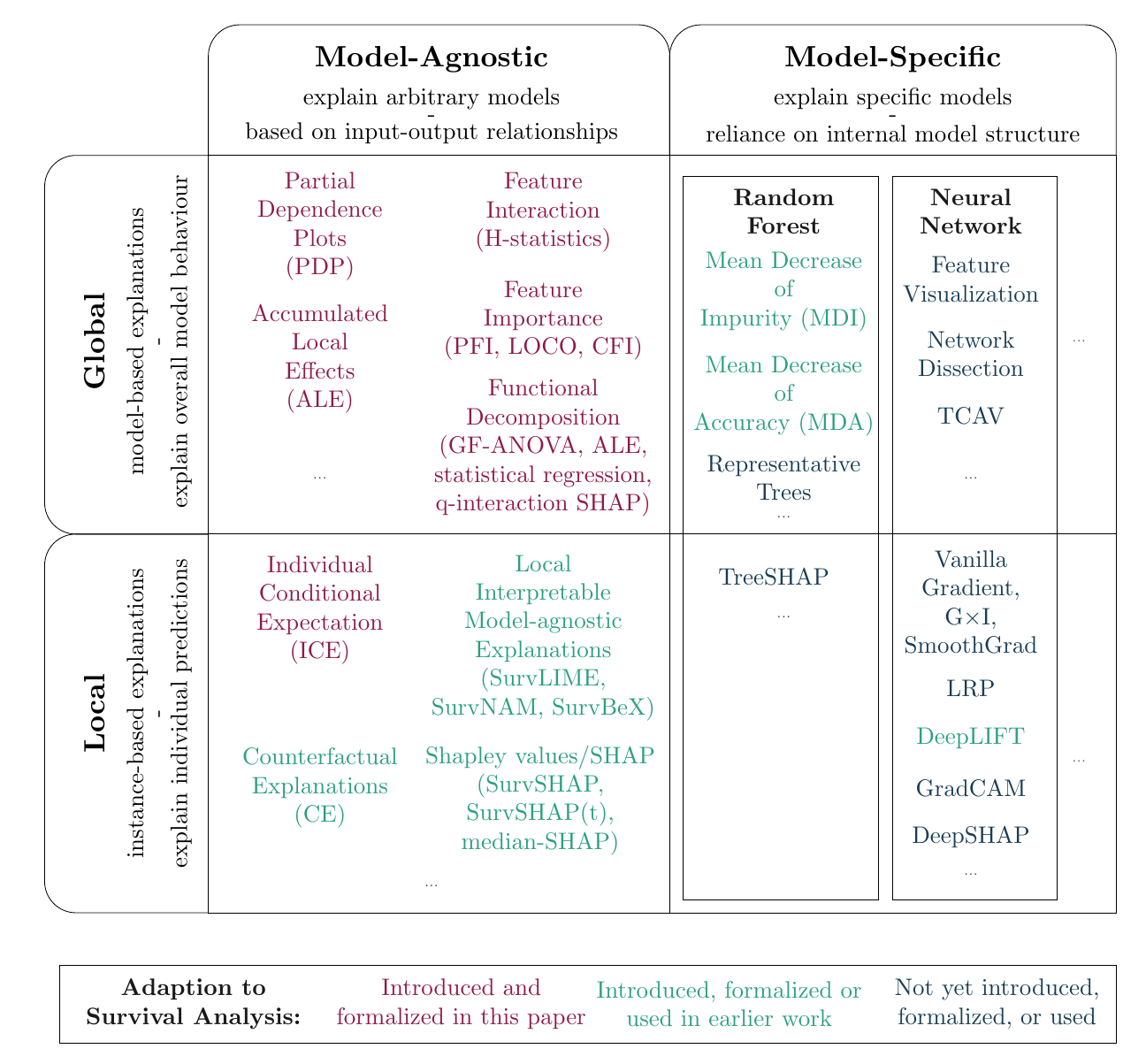}
  \caption{Overview over the interpretable machine learning methods reviewed and newly introduced in this paper in the context of the taxonomy popularized by Molnar\cite{molnar2022} and Biecek and Burzykowski\cite{biecek2021}.}
  \label{fig:overview_taxonomy}
\end{figure}

The paper is structured into three main parts to address the aforementioned concerns. The first focus of the paper lies on a comprehensive, focus-driven review of relevant contributions to survival interpretable machine learning, with a particular emphasis on model-agnostic explainability methods. Existing contributions are examined on a conceptual level, with attention directed toward their limitations with respect to real-world data and applications. The second focus of the paper is to mathematically and conceptually translate a set of fundamental interpretable machine learning methods from the classical machine learning setting to survival analysis. For model-agnostic local methods, this encompasses individual conditional expectation curves, while global model-agnostic methods are comprised of partial dependence curves, accumulated local effects, Friedman's H-statistic for feature interaction, permutation feature importance, conditional feature importance, leave-one-covariate-out, and functional decomposition methods. For an overview of the reviewed and newly adapted methods, consider Figure~\ref{fig:overview_taxonomy}. The third and final focus is put on providing a tutorial that showcases both existing and newly adapted methods using real-world data on cancer recurrence in the well-known GBSG2 dataset from the German Breast Cancer Study Group. The tutorial aims to guide practitioners in utilizing interpretability methods to elucidate the opaque mechanisms of black box machine learning survival models and gain a broader perspective on interpretation beyond simple survival curves and hazard ratios, also for classical statistical models.

\subsection{Notation and Key Concepts of Survival Analysis}\label{survival_analysis}
The objective of survival analysis is to model distribution of the time to (survival) event $T$ given a set of training data $D$. In the standard right-censored setting, the training set $D$ consists of $n$ triplets $(\mathbf{x}^{i},\delta^{i},t^{i})_{i=1}^{n}$, where $\mathbf{x}^{i} = (x_1^{i},\dots,x_p^{i})^{\intercal}$ is a vector of the $p$ covariates or features. We refer to a specific, undefined feature as $x_j$ with $j \in \{1,\dots,p\}$. If a selected instance $i$ is observed, $t^{i}$ represents the true duration between the baseline time and the event occurrence for that instance and the censoring indicator $\delta^{i}$ takes the value 1. In contrast, if the event is not observed, $t^{i}$ signifies the time elapsed between the baseline time until either the end of the observation period or until time of dropout and the true time to event is not observed. Then $\delta^{i} = 0$ is indicating a censored observation and $\boldsymbol{\delta} = (\delta^0,\dots,\delta^n)^{\intercal}$ is the censoring vector. The set of observed times to event (including censoring times) is defined as $\mathbf{t}^{\text{obs}} = (t^{1},\dots,t^n)^{\intercal}$, with $t^{u} \in \{t^{1},\dots,t^{n}\}$; the ordered, unique, observed times to event are $\mathbf{t}^{\text{ord}} = (t^{min},\dots,t^{max})^{\intercal}$, with $t^{o} \in \{t^{min},\dots,t^{max}\}$. 

Event time distributions are usually described by a set of common functions, such as survival or hazard functions. We generalize them as $f:\{\mathcal{X},\mathcal{T}\} \rightarrow \mathbb{R}$, a set of functions that map value combinations from the feature space $\mathcal{X}$ and the time space $\mathcal{T}$ to a one-dimensional outcome. The survival function $S(t|\mathbf{x}) = \mathbb{P}(T > t|\mathbf{x})$ gives the probability of surviving beyond a specific timepoint $t$. The hazard function $h(t|\mathbf{x})$ describes the instantaneous rate of occurrence of the event of interest at a specific time $t$, given that no event occurred before time $t$. If $\text{p}(t|\mathbf{x})$ is the probability density function of the event of interest, for which $\text{p}(t|\mathbf{x}) = -\frac{\text{d}S(t|\mathbf{x})}{\text{d}t}$ holds, then $h(t|\mathbf{x})$ is defined as
\begin{align}\label{eq:hazard_function}
    h(t|\mathbf{x}) = \lim_{\Delta t \rightarrow 0} \frac{\mathbb{P}(t \leq T \leq t + \Delta t | T \geq t, \mathbf{x})}{\Delta t} = \frac{\text{p}(t|\mathbf{x})}{S(t|\mathbf{x})} = -\frac{\text{d}}{\text{d}t} \ln S(t|\mathbf{x}) \; \text{.}
\end{align}
\noindent
The integral of the hazard function up to a specific point in time $t$ is the cumulative hazard function (CHF) $\Lambda(t|\mathbf{x})$. It represents the cumulative risk of experiencing the event of interest up to a specific time $t$, i.e.,
\begin{align}\label{eq:CHF}
    \Lambda(t|\mathbf{x}) = \int_{0}^{t} h(z|\mathbf{x}) \text{d}z \; \text{.}
\end{align}
Using the hazard function and the cumulative hazard function, the survival function can be expressed as
\begin{align}\label{eq:survival-function}
    S(t|\mathbf{x}) =  \mathbb{P}(T > t|\mathbf{x}) = \exp \left( - \int_0^{t} h(z|\mathbf{x}) \text{d}z \right) = \exp(-\Lambda(t|\mathbf{x}))  \; \text{.}
\end{align}

The Kaplan-Meier estimator of the survival function is given by
\begin{align}\label{eq:KM}
    \hat{S}_{KM}(t) &= \prod_{t^o \leq t} \left( 1 - \frac{d^o}{n^o}\right) \text{,}
\end{align}
where $t^{o} \in \{t^{min},\dots,t^{max}\}$ represents one element of the ordered, observed event times, $d^o$ denotes the count of events at time $t^{o}$, while $n^o$ is the number of individuals who have not yet experienced an event or been censored by the specified time $t^o$.

The Cox regression model or Cox proportional hazards (CoxPH) model\cite{cox1972} defines the survival function $S(t|\mathbf{x},\mathbf{b})$ and the hazard function $h(t|\mathbf{x},\mathbf{b})$ at time $t$ given a set of features $\mathbf{x}$ as: 
\begin{align}
    \hat{S}(t|\mathbf{x},\mathbf{b}) &= (\hat{S}_0(t))^{\exp(\varphi(\mathbf{x},\mathbf{b}))}, \label{eq:coxph1}\\ \hat{h}(t|\mathbf{x},\mathbf{b}) &= \hat{h}_0(t) \exp(\varphi(\mathbf{x},\mathbf{b})), \label{eq:coxph2}\\ \varphi(\mathbf{x},\mathbf{b}) &= \mathbf{b}^{\intercal}\mathbf{x} = \sum_{j=1}^{p} b_j x_j  \; \label{eq:coxph3}\text{.}
\end{align}
Here $\mathbf{b}^{\intercal} = (b_1,\dots,b_p)$ is an unknown vector of regression coefficients, $S_0(t)$ is the estimated baseline survival function and $h_0(t)$ the estimated baseline hazard function. Key properties of the CoxPH model are the linearity of the function $\varphi(\mathbf{x},\mathbf{b})$ and its independence of time, as well as the independence of $h_0(t)$ and $S_0(t)$ from $\mathbf{x}$ and $\mathbf{b}$. 

\section{Review of Interpretable Machine Learning Methods}

\subsection{Local Methods}

\subsubsection{Individual Conditional Expectation (ICE)}\label{sec:ice}

Originally introduced by Goldstein et al.\cite{goldstein2015}, individual conditional expectation (ICE) curves visualize changes in prediction for one observation induced by feature changes. So far, they have not yet been formally adapted to survival data applications, which we present in the following. 
To formally define the ICE curves, let $A \in \{1,\dots,p \}$ represent an element from the index set of all features of interest and let $-A$ be the complement set of $A$. The function predicted by the survival machine learning model is denoted as $\hat{f}(\cdot)$. For each observation, the feature matrix $\mathbf{X}$ is partitioned into $\mathbf{X}_A$ and $\mathbf{X}_{-A}$. Since $|A| = 1$, the feature matrix partition $\mathbf{X}_A$ is a vector $\mathbf{x}_A$, which is also the most common setting for visualization purposes. However, all subsequent definitions are directly generalizable to subsets $A \subset \{1,\dots,p \}$ with $|A| > 1$. To obtain the ICE curves, a set of grid values $\mathbf{x}_A^{\text{grid}} = \{ x_A^{1}, \dots, x_A^{g} \}$ is defined for the feature of interest. Then, for one specific $i=1,\dots,n$ a prediction is obtained by varying over the elements in $\mathbf{x}_A^{\text{grid}}$ and $\mathbf{t}^{\text{ord}}$, while holding the observed feature values corresponding to the index set $-A$ constant, such that
\begin{align}\label{eq:ICE}
\hat{f}_{\text{ICE},A}^{i}(\mathbf{t}^{\text{ord}}) = \hat{f}_{\text{ICE}}^{i}(\mathbf{t}^{\text{ord}}|\mathbf{x}_A) = \hat{f}(\mathbf{t}^{\text{ord}}|\mathbf{x}_A^{\text{grid}},\mathbf{x}_{-A}^{i}) \; \text{.} 
\end{align}
From this a set of $\{1,\dots,g\} \times \{\min,\dots,\max\}$ ordered pairs is obtained: $\left\{\left\{ (x_A^{k}, t^{o}, \hat{f}_{\text{ICE}}^{i}(t^{o}|x_A^{k})) \right\}_{k=1}^{g}\right\}^{\max}_{o=\min}$. 

Common choices for grid values are equidistant grid values within feature range, randomly sampled values or quantile values of the observed feature values. The latter two options approximately preserve the marginal distribution of $X_A$, thus avoiding unrealistic feature values for distributions with outliers. Which timepoints are of particular interest depends on the specific purpose of the analysis, in many cases an equidistant grid of few representative values within the range of $[t^{\min}, \dots, t^{\max}]$ may suffice.

To demonstrate how ICE curves can be used to improve understanding of the predictions of survival models we simulate data for $N=3000$ patients with a maximum follow up time of five years under a standard Weibull survival model that incorporates both a time-independent and a time-dependent effect (i.e. non-proportional hazards). The time-dependent effect is modeled using a single binary feature (e.g., a treatment indicator), which initially has a time-independent protective effect (i.e., a negative log hazard ratio), but is eventually overshadowed by a time-dependent adverse effect. A real world example for this could be an effective, but highly invasive treatment such as radio-, or chemotherapy. Additionally, two time-independent features are included: one without any effect (\texttt{x}$_2$) and one with a negative effect on the survival outcome (\texttt{x}$_1$). Further details on the simulation set-up are provided in Appendix~\ref{app1:sim1}. The simulated data is split into training and test set and a CoxPH model (\texttt{coxph}) and a random survival forest model (\texttt{ranger} from \texttt{ranger} package\cite{wright2017ranger}) are fit to the training set. 

\begin{figure}[htb!]
  \centering
  \includegraphics[width = 0.9\linewidth]{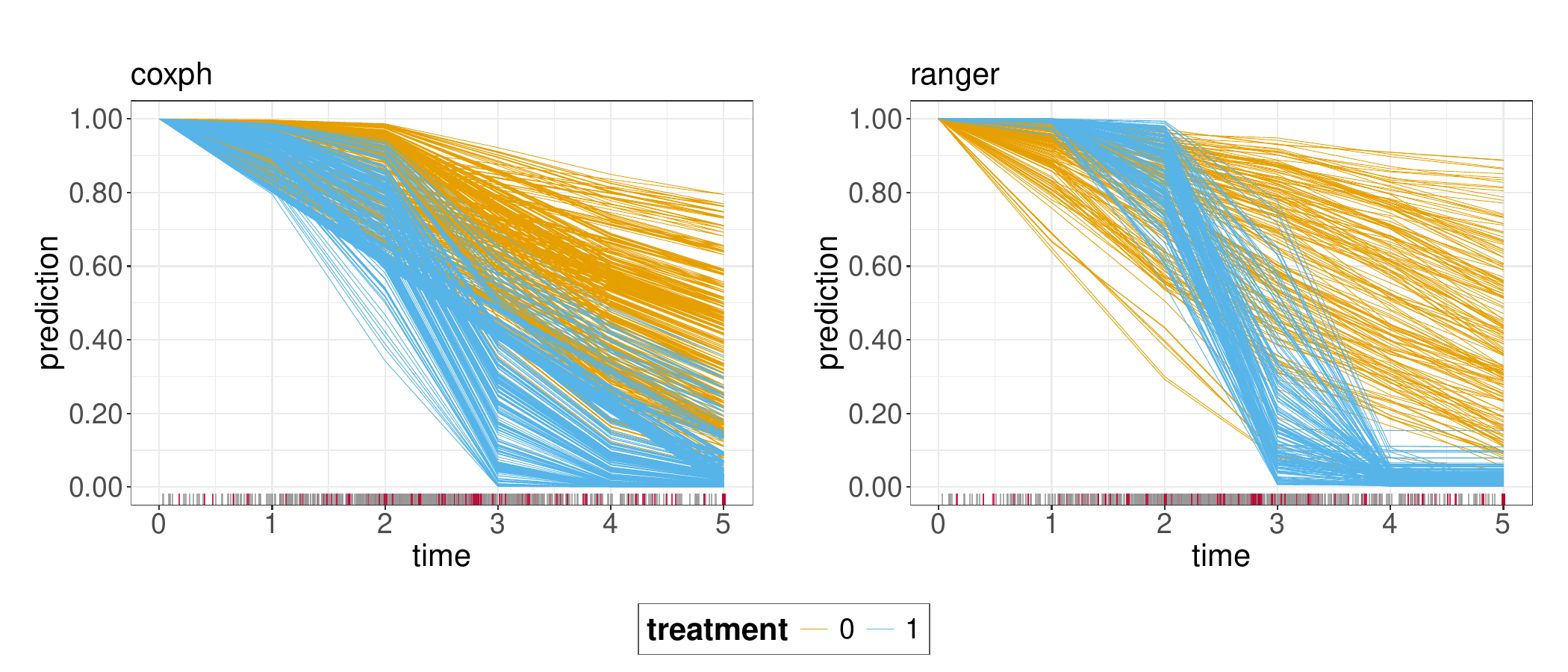}
  \caption{ICE curves for the \texttt{coxph} model (left) and the \texttt{ranger} model (right) for the time-dependent \texttt{treatment} feature. One ICE curve shows how the model's prediction varies over time for one observation for a fixed \texttt{treatment} strategy. The different line colors correspond to different \texttt{treatment} strategies (0 = no \texttt{treatment}, 1 = \texttt{treatment}). Therefore, each observation from the simulated test dataset is represented by one orange line (\texttt{treatment} = 0) and one blue line (\texttt{treatment} = 1). The rug on the x-axis shows the survival time distribution with the grey bars indicating observed survival times and the red bars indicating censoring.}
  \label{fig:ICE_uc_sim}
\end{figure}

Figure~\ref{fig:ICE_uc_sim} shows the ICE curves obtained according to Equation~\ref{eq:ICE}. In theory, each ICE curve reflects the predicted value as a function of the selected set of features in $A$ and a set of time values, conditional on the observed values for all features in $-A$ and for each element in $\mathbf{t}^{\text{ord}}$. In case the feature of interest is binary or categorical, as for the given simulation example, the choice of grid values is trivial, simply consisting of the feature levels. For ease of visualization, the time is then denoted on the x-axis and different colors are used for the different feature values (orange = no \texttt{treatment}; blue = \texttt{treatment}). For a continuous feature this is interchangeable.

\begin{figure}[htb!]
  \centering
  \includegraphics[width = 0.9\linewidth]{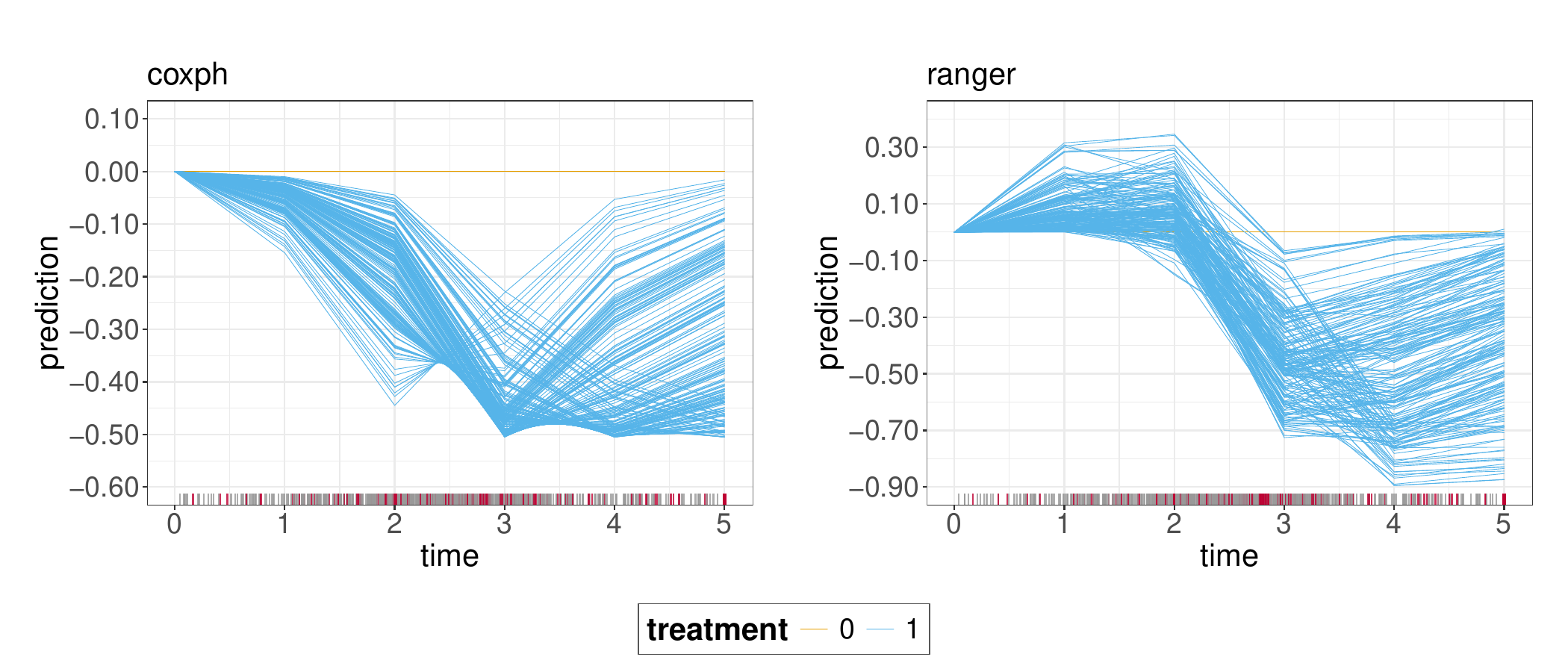}
  \caption{c-ICE curves for the \texttt{coxph} model (left) and the \texttt{ranger} model (right) for the time-dependent treatment feature. One ICE curve shows how the model's prediction varies over time for one observation for a fixed treatment strategy relative to a no treatment strategy for each individual patient. As a result, the c-ICE curves are 0 constants for treatment = 0 for both models (for the \texttt{ranger} model, the orange curves are simply covered by the blue curves.)}
  \label{fig:ICE_c_sim}
\end{figure}

Generally speaking, different shapes of ICE curves at a fixed point in time $t$, suggest interaction between $\mathbf{x}_A$ and $\mathbf{X}_{-A}$ (i.e. between \texttt{treatment} and $\texttt{x}_2$), while different shapes of ICE curves for a fixed value of $\mathbf{x}_A$, suggest time-dependency of the feature of interest. The latter is particularly visible for the $\texttt{treatment}=1$ ICE curves generated from the \texttt{ranger} model in Figure~\ref{fig:ICE_uc_sim}, while the other curves are largely homogeneous within their treatment group for $\texttt{treatment}=0$ for both models. However, if ICE curves have different intercepts, i.e. stacked curves, heterogeneity in the model can be difficult to identify. Since the ground truth in our simple simulated example does not involve any interaction between the \texttt{treatment} and other features, stacked ICE curves are not present in Figure~\ref{fig:ICE_uc_sim}. Centering the ICE curves (so-called "c-ICE" curves) to remove level effects at a fixed reference value $X^{\prime} \sim \mathbb{P}(\mathbf{X}_A)$ simplifies discerning the heterogeneous shapes. The points $(x^{\prime},\hat{f}(\mathbf{t}^{\text{ord}}|x^{\prime},\mathbf{x}_{-A}^{i}))$ serve as reference points for every curve. The $g \times \max$ ordered c-ICE curve triplets, for every $i=1,\dots,n$ are then given by: $\left\{\left\{ (x_A^{k}, t^{o}, \hat{f}^{i}_{\text{c-ICE}}(t^{o}|x_A^{k})) \right\}_{k=1}^{g}\right\}^{\max}_{o = \min}$ with
\begin{align}\label{eq:c-ICE}
\hat{f}^{i}_{\text{c-ICE},A}(\mathbf{t}^{\text{ord}}) = \hat{f}^{i}_{\text{c-ICE}}(\mathbf{t}^{\text{ord}}|\mathbf{x}_A) = \hat{f}_A^{i}(\mathbf{t}^{\text{ord}}|\mathbf{x}_A) - \hat{f}_A^{i}(\mathbf{t}^{\text{ord}}|x^{\prime}) = \hat{f}(\mathbf{t}^{\text{ord}}|\mathbf{x}_A^{\text{grid}},\mathbf{x}_{-A}^{i}) - \hat{f}(\mathbf{t}^{\text{ord}}|x^{\prime},\mathbf{x}_{-A}^{i}) \; \text{.}
\end{align}
The most common choices for $x^{\prime}$ are the minimum or maximum values of $\mathbf{x}_A$. To obtain c-ICE curves for the simulated we center at $x^{\prime} = \min(\mathbf{x}_A) = \texttt{treatment} = 0$, the results are shown in Figure~\ref{fig:ICE_c_sim}. By setting $x^{\prime} = \min(\mathbf{x}_A)$ all curves originate at $0$, thereby eliminating level differences caused by varying $\mathbf{x}_{-A}^{i}$ values.\footnote{For $x^{\prime} = \max(\mathbf{x}_A)$,  the level of each centered curve reflects the cumulative effect of $\mathbf{x}_A$ on $\hat{f}(\cdot)$ relative to the reference point $x^{\prime}$. In this way, the isolation of the combined effect of $\mathbf{x}_A$ on $\hat{f}(\cdot)$ is improved, holding the values of all other features $-A$ fixed.} The centering operation helps to clearly isolate the effect of \texttt{treatment} $=1$. Only the \texttt{ranger} model successfully captures the ground truth, accurately reflecting the initial positive effect of the treatment that transitions into a negative effect over time. 

It is further possible to reduce the dimensionality to match classic (non-survival) ICE curves by marginalizing over time; we refer to this as ICE-T curves. This simplification helps visualize the relationship between the feature of interest and the predicted survival outcome while retaining the temporal dynamics. In the simplest form we can average over the set $\{t^{min}, \dots, t^{max}\}$, such that for the ICE-T curves a set of $g$ ordered pairs is obtained for every $i=1,\dots,n$: $\left\{ (x_A^{k}, \frac{1}{|\mathbf{t^{\text{ord}}}|} \sum_{o = \min}^{\max} \hat{f}_A^{i}(t^o|x_A^{k})) \right\}_{k=1}^{g}$. Alternatively, to obtain a more established identifier, we can simply sum over the observed survival times $t^1,\dots,t^m$, where $t^m$ is a particular (user selected) timepoint of interest, such that the ordered pairs $\left\{ (x_A^{k}, \sum_{u=1}^{m} \hat{S}_A^{i}(t^u|x_A^{k})) \right\}_{k=1}^{g}$ estimate the restricted mean survival time (RMST) for an individual $i=1,\dots,n$ conditional on the feature value $x_A^{k}$ up to time $t^m$.

ICE curves offer an intuitive and transparent way of assessing the consistency and robustness of model predictions across different observations, aiding in the detection of potential biases or inconsistencies in the model's behavior. They further play a central role in uncovering heterogeneous relationships between features and predictions on an individual-level, drawing attention to both feature interactions and time-varying feature effects.

\newpage
\subsubsection{Counterfactual Explanations (CE)}\label{sec:counterfactuals}
A counterfactual $\mathbf{c}^{i} = (c_1^{i},\dots,c_p^{i})^{\intercal}$ or "what-if" explanation, identifies the minimum change required in the original feature values $\mathbf{x}^{i} = (x_1^{i},\dots,x_p^{i})^{\intercal}$ of a particular instance $i$, such that the model's prediction is modified to align with a predefined outcome.\citep{molnar2022}
Kovalev et al.\cite{kovalev2021c} propose a method for counterfactual explanations of machine learning survival models, which is derived from the method introduced by Wachter et al.\cite{wachter2017}. They define an optimization problem to be solved for finding counterfactual $\mathbf{c}$ by minimizing the sum of a loss function $L^{\text{dist}}(\cdot)$ and the Euclidean distance between the original observation and the counterfactual~$\lVert\mathbf{c} - \mathbf{x} \rVert_2$:
\begin{align}\label{counterfactual}
    \min_{\mathbf{c} \in \mathbb{R}^p} L(\mathbf{c}) = \min_{\mathbf{c} \in \mathbb{R}^p} \{ L^{\text{dist}} \left( \mathbb{E}(\mathbf{c}),\mathbb{E}(\mathbf{x}) \right) + C \lVert\mathbf{c} - \mathbf{x} \rVert_2 \}  \; \text{.}
\end{align}
The parameter $C > 0$ determines the regularization strength. For $L^{\text{dist}}$, Kovalev et al. consider a hinge-loss of the following form:
\begin{align}\label{counterfactual_hinge_loss}
    &L^{\text{dist}} \left( \mathbb{E}(\mathbf{c}),\mathbb{E}(\mathbf{x}) \right) = \max \left(0, r - \left(\mathbb{E}(\mathbf{c}) - \mathbb{E}(\mathbf{x}) \right) \right)\text{,  with} \\ &\mathbb{E}(\mathbf{x}) = \int_0^{\infty} S(t|\mathbf{x}) \text{d} t \; \; \text{and} \; \; \mathbb{E}(\mathbf{c}) = \int_0^{\infty} S(t|\mathbf{c}) \text{d} t  \; \text{.}
\end{align}
For some $r \geq 0$, minimizing the hinge loss prompts the identification of counterfactuals $\mathbf{c}$, which increase the distance in mean times to events for $\mathbf{c}$ and $\mathbf{x}$ up to $r$. In this context, $r$ requires explicit definition by the modeler, as a smallest distance in mean times to events characterizing distinct groups of instances (e.g., patients). The rationale for configuring the loss function in this manner is to produce counterfactual statements such as the following: “Your treatment was not successful because of a small dose of the drug (one tablet). If your dose had been three tablets, the predicted mean time of recession would have been increased to a required value”. In general, Equation~\ref{counterfactual} poses a non-convex optimization problem, which Kovalev et al. propose to solve using the particle swarm optimization algorithm\cite{eberhart1995}.

To evaluate the counterfactual explanation proposed by Kovalev et al., we can consider five concrete criteria for a good counterfactual explanation defined by Dandl et al.\cite{dandl2020} and Molnar\cite{molnar2022}: (1) the prediction associated with $\mathbf{c}$ is close to the desired outcome, (2) $\mathbf{c}$ is close to $\mathbf{x}$ in the feature space $\mathcal{X}$, (3) $\mathbf{c}$ differs from $\mathbf{x}$ only in a few features, (4) $\mathbf{c}$ is a plausible data point according to the probability distribution $\mathbb{P}(\mathcal{X})$, and (5) the underlying algorithm is able to generate multiple diverse counterfactual explanations. The optimization problem in Equation~\ref{counterfactual} minimizes two terms. The Euclidean distance between counterfactual $\mathbf{c}$ and original input $\mathbf{x}$ acts as a penalty term for deviations of $\mathbf{c}$ from $\mathbf{x}$, effectively fulfilling criterion (2). Minimizing the hinge loss as delineated in Equation~\ref{counterfactual_hinge_loss} functions to satisfy criterion (1). Kovalev et al. argue that, e.g., doctors rarely think in basic concepts of survival analysis and that medical practitioners usually exhibit a preference for scalar metrics, such as mean time to event, when describing, for instance, at-risk cohorts. Nonetheless, it can be posited that representing the counterfactual by a difference in mean time to event holds limited practical significance and even engenders misleading interpretations in real-world applications. This is exemplified by the illustrative interpretation provided by Kovalev et al., which links the reduction in mean recession time to a drug dose in a causal manner (see prior paragraph). While the method encourages this causal conclusion, its validity is highly questionable. In general, collapsing the survival distribution or hazard function into one number is associated with a significant loss of information. The mean time to event is particularly problematic, due to its inability to account for censoring, cohort dynamics and time-dependent effects. Oftentimes in clinical or biomedical research, specific percentiles (e.g., the 25th or 75th percentile or the median) or other summary measures may be more clinically relevant than the mean. The main argument in favor of the mean time to event definition is computational convenience. Equation~\ref{counterfactual} simplifies to a convex optimization problem if the Cox regression model is chosen as a prediction model. Both in their simulation studies as well as their real data examples, Kovalev et al. place significant emphasis on substantiating computational accuracy, notably with regard to the particle swarm optimization algorithm. However, they do not engage in a substantive discourse pertaining to the practical utility and interpretability of counterfactual outcomes when applied to real-world data. Criteria (3)-(5) are not at all considered in their survival counterfactual algorithm. Moreover, the optimization problem is unable to account for categorical data. In essence, the counterfactual algorithm proposed by Kovalev et al. serves as a robust foundational framework upon which to develop more advanced algorithms. These advanced algorithms can be carefully calibrated to accommodate the nuances and distinct requisites for counterfactual explanations that hold practical utility across diverse domains of survival analysis. In the future, more meaningful ways of defining groups of instances to differentiate counterfactuals from the original inputs need to be considered, as well as augmentations to the original algorithm by Wachter et al.\cite{wachter2017}, such as the multi-objective algorithm proposed by Dandl et al.\cite{dandl2020}.  

\subsubsection{Local Interpretable Model-Agnostic Explanations (LIME)}\label{sec:LIME}
Kovalev et al.\cite{kovalev2020} have proposed an extension of the well-known local interpretable model-agnostic explanations (LIME) algorithm\cite{ribeiro2016} called \emph{SurvLIME}. LIME and \emph{SurvLIME} are specific variants of the idea of local surrogate models. These are designed to approximate the behavior of black box models in a local region of the input space, usually to provide insights into the decision-making process for one individual prediction. The \emph{SurvLIME} explanation algorithm approximates the CHF generated by the the black box model using the Cox regression model's CHF for the same input example $\mathbf{x}$. This idea is based on the fact that the Cox regression is considered an interpretable model, as discussed in Section~\ref{interpretable_models}. In a first step, a point of interest $\mathbf{x}$ is selected around which a set of nearby points $\mathbf{x}^k$ with $k=1,\dots,g$ is generated. Then the survival machine learning model is used to predict cumulative hazard functions $\Lambda(t|\mathbf{x}^k), \;  \;  \forall k \in \{1,\dots,g\}$. In a next step, the optimal coefficients $\mathbf{b}$ are determined such that the average Euclidean distance between every pair $\Lambda(t|\mathbf{x}^k)$ and $\Lambda_{\text{Cox}}(t|\mathbf{x}^k,\mathbf{b})$ is minimized over all generated points $\mathbf{x}^1,\dots,\mathbf{x}^g$ for all ordered event times $t = t_{\min},\ldots,t_{\max}$. This mean distance between every CHF pair over all generated points is considered, since survival models produce functional outputs rather than pointwise outputs as in the classical setting. Each CHF distance is further assigned a weight $w^k$, which is larger for smaller distances between $\mathbf{x}$ and $\mathbf{x}^k$ and vice versa. A considerable simplification of the resulting minimization problem is induced by the independence of $\varphi(\mathbf{x},\mathbf{b})$ from time, implying that time and features can be considered separately. Kovalev et al. show that the resulting optimization problem is unconstrained and convex, which is beneficial, because it guarantees a unique global optimum and can be solved efficiently using well-established algorithms. The computation of \emph{SurvLIME} is summarized in Algorithm~\ref{alg:LIME} taken directly from Kovalev et al.\cite{kovalev2020}. 

\begin{algorithm}[!htb]
\caption{\emph{SurvLIME}: algorithm for computing vector $\mathbf{b}$ for point $\mathbf{x}$}
\label{alg:LIME}
\begin{algorithmic}[1]
\Require Training set $D$; point of interest $\mathbf{x}$; the number of generated points $g$; the black box survival model $f(\cdot)$
\State Compute the baseline cumulative hazard function $\Lambda_0(t)$ of the approximating CoxPH model on dataset $D$ by using the Breslow estimator.\footnotemark
\State Generate $g$ random nearest points, in a local area around $\mathbf{x}$ (including $\mathbf{x}$), such that $\mathbf{x}^k$ for every $k = 1, \dots, g$
\State Get the prediction of the cumulative hazard function $\Lambda(t|\mathbf{x}^k)$, for every $k = 1, \dots, g$, by using the black box survival model
\State Compute weights $w_k = K(\mathbf{x}, \mathbf{x}^k)$ of perturbed points, $k = 1, \dots, g$, where $K(\cdot,\cdot)$ is a kernel function
\State Find vector $\mathbf{b}$ by solving the convex optimization problem
\end{algorithmic}
\end{algorithm}
\footnotetext{The original algorithm from the Kovalev et al. paper states that the Nelson-Aalen estimator is used. However, the existing software implementations (\texttt{survex}\cite{spytek2023} and \texttt{SurvLIMEpy}\cite{pachon2024} use the Breslow estimator, which is commonly used to estimate the baseline hazard in the CoxPH model.}

Kovalev and Utkin et al. have also developed modifications to \emph{SurvLIME}. \emph{SurvLIME-Inf}\cite{utkin2020inf} uses the Chebyshev distance $L_{\infty}$ to define the distance between the CHFs. This choice results in a straightforward linear programming problem and \emph{SurvLIME-Inf} is shown to outperform \emph{SurvLIME} with small training sets. Additionally, in Kovalev et al.\cite{kovalev2020ks} \emph{SurvLIME-KS} is proposed, which uses Kolmogorov-Smirnov bounds to construct sets of predicted CHFs. They are obtained by transforming CHFs into cumulative distributions functions. The optimization problem is then formulated as a \emph{maximin} problem. The minimum is defined for the average distance between logarithms of CHFs from the black box survival model and the surrogate Cox regression model. The maximum is taken over logarithms of all CHFs restricted by the obtained KS-bounds. \emph{SurvLIME-KS} promotes robustness to small training sets and outliers. Utkin et al.\cite{utkin2023survbex} contribute \emph{SurvBeX}, which uses a modification of the Beran estimator\cite{beran1981}, which produces local feature importance values, to approximate the survival function predicted by the black box model. The Beran estimator, akin to kernel regression, computes the survival function with greater flexibility compared to the linearly constrained CoxPH approximation utilized in the original \emph{SurvLIME}. 
Furthermore, building up one the ideas of \emph{SurvLIME}, Utkin et al.\cite{utkin2021unc} have proposed \emph{UncSurvEx} a method for uncertainty interpretation of black box survival model predictions. The technique aims to approximate the uncertainty measure of the black box prediction by the uncertainty measure of the Cox model prediction. 

\emph{SurvLIME} and its children inherit all limitations from the original LIME method. Like LIME, \emph{SurvLIME} is a purely data-driven, empirical approach that lacks theoretical foundations and guarantees. As a consequence, the quality and stability of the generated explanations is heavily dependent on the selected sample, the chosen perturbation process and the hyperparameters.\cite{slack2020,zhou2021} For instance, in their paper, Kovalev et al. assign weights to every point using the Epanechnikov kernel $w_k = 1 - \sqrt{\frac{\lVert\mathbf{x}-\mathbf{x}_k\rVert_2}{r}}$ with $r = 0.5$, providing no further justification on the kernel choice or comparisons to results using different kernel functions. Lastly, the explanations obtained from \emph{SurvLIME} are heavily dependent on the chosen surrogate model suffering from its innate restrictions. Employing the Cox regression as a surrogate, implies assuming a local linear relationship between the logarithm of the hazard and the predictors. Furthermore, as discussed in Section~\ref{interpretable_models}, the proportional hazards and time-independence are assumed to hold locally, which are often not met in real-world datasets. The simulation studies and experiments on real data performed by Kovalev et al. provide little substantiation of the validity of the local explanations provided by the CoxPH model. The focus of the simulation studies relies on proving the similarity of the true underlying $\mathbf{b}$ coefficients and the coefficients of the approximating CoxPH model, when assuming the data generating process to follow a CoxPH model. This emphasis primarily underscores the efficacy of the convex programming algorithm employed, with a lesser focus on evaluating \emph{SurvLIME's} efficacy as an interpretability method. Notably, the simulation studies exclusively pertain to data conforming to an underlying CoxPH model. Additionally, the analysis refrains from presenting further comparative assessments between the outcomes of \emph{SurvLIME} and alternative local interpretability methodologies. While certain limitations, such as the examination of varying hyperparameter selections or the formulation of an extension to accommodate time-dependent explanations, may be addressed, it is important to note that \emph{SurvLIME} is bound by the fundamental constraints inherent in the original LIME explanation methodology. 

To alleviate some of the drawbacks of \emph{SurvLIME}, Utkin et al. have proposed \emph{SurvNAM}\cite{utkin2022}. The basic idea behind the explanation algorithm \emph{SurvNAM} is to approximate the black box model by a generalized additive CoxPH model, in which $\varphi(\mathbf{x},\mathbf{b})$ is replaced with
\begin{align}\label{eq:SurvNAM}
    \varphi_{\text{NAM}}(\mathbf{x},\gamma) = \gamma_1(\mathbf{x}_1) + \dots + \gamma_p(\mathbf{x}_p)  \; \text{.}
\end{align}
The Cox generalized additive model (CoxGAM) is in turn approximated in the form of a neural additive model (NAM), used to learn the shape functions $g_j(\mathbf{x}_j), \; \forall j \in \{1,\dots,p\}$. One shape function corresponds to one network. These subnetworks may exhibit different architectures, but are trained jointly using backpropagation\cite{rumelhart1986} and have the capacity to learn arbitrarily complex shape functions. To generate local explanations, again synthetic points are randomly generated around an instance of interest $\mathbf{x}$, for which black box predictions of the CHF are produced by the model to be explained. To train the neural network, the weighted Euclidean distance between the CHFs of the black box model $\Lambda(t|\mathbf{x})$ and the CHF of the approximating NAM $\Lambda_{\text{NAM}}(t|\mathbf{x},\gamma)$ is minimized for all generated points. The resulting expected loss function is convex. In contrast to \emph{SurvLIME}, the authors specifically encourage the use of \emph{SurvNAM} as a global surrogate model, by simply replacing the synthetic points generated around a local example of interest with the whole training set $D$. Due to the additive nature of the model, the marginal effect of each feature is independent of the rest, hence the shape functions correspond to feature contribution functions, which can be visualized for explainability purposes. Overall, \emph{SurvNAM} surpasses \emph{SurvLIME} in terms of flexibility and the richness of its output, while also harboring potential for future advancements. The authors have, for example, already proposed several extensions. The first employs the Lasso-based regularization for the shape functions from GAM\cite{hastie1986} to produce a sparse representation, which prevents the neural network from overfitting. The second modification adds a linear component to each shape function with some additional weights to be trained, mitigating the vanishing gradient problem. The third extension combines \emph{SurvNAM} with \emph{SurvBeX} to create the Survival Beran-based Neural Importance Model \emph{SurvBeNIM}, which integrates importance functions into the kernels of the Beran estimator. These functions, akin to GAM shape functions albeit without additivity, are estimated using neural networks. The authors consider \emph{SurvNAM} and its extensions sufficiently flexible to adequately capture the black box model results. This assumption of comparable complexity to the black box model heavily depends on the specific black box model and data analyzed and has so far not been proven in many practical applications. It would, in fact, warrant the question, as to why the interpretable model should not entirely replace the black box model, which is a general flaw inherent in the idea of global surrogate models. The absence of robust guarantees regarding their fidelity to the original model's behavior and the risk of introducing bias due to assuming an additive model structure strike as additional limitations of the \emph{SurvNAM} concept.

\subsubsection{Shapley Additive Explanations (SHAP)}\label{shapley}

The application of the popular SHAP algorithm \cite{lundberg2017} to the survival analysis context has been explored by Alabdallah et al.\cite{alabdallah2022}, Krzyziński et al.\cite{krzyzinski2023shap} and Ter-Minassian et al.\cite{ter-minassian2024}, who propose distinct approaches for its implementation. SHAP computes the Shapley values\cite{shapley1953} of the features, which quantify the contribution of each feature to the difference between the model prediction for a given sample and the average prediction of the model. Global explanations can be conveniently derived by aggregating Shapley values across a multitude of instances.\cite{lundberg2020} 

Alabdallah et al. propose \emph{SurvSHAP}, an algorithm enabling the derivation of those global explanations based on the original SHAP algorithm, without directly modifying it for survival analysis. Instead, SHAP is used to explain a proxy classification model that learns the mapping from the input space $\mathcal{X} \in \mathbb{R}^{n \times p}$ to a set of survival patterns. These survival patterns are homogeneous sub-populations with similar survival behavior according to the machine learning survival model’s predictions. To achieve this, principal components analysis is employed to reduce the dimensionality of the survival curves to a lower-dimensional representation, denoted as $\mathcal{D} \in \mathbb{R}^{n \times d}$. Subsequently, the curves in the $\mathcal{D}$ space undergo an iterative clustering process, and the resulting sub-populations are compared using log-rank tests. The fraction of significantly different groups is calculated from the total number of comparisons. Finally, the largest number of sub-populations that maximizes the percentage of distinct patterns is selected to cluster the curves in the $\mathcal{D}$ space, resulting in the final set of clusters. In their experimental evaluation involving both real and synthetic data, Alabdallah et al. propose K-means and a random forest classifier as a default clustering and proxy classification method, respectively. Additionally, they use the TreeSHAP\cite{lundberg2020} algorithm as a default base explainer, albeit without performing sensitivity analyses for these choices. Despite being termed \emph{SurvSHAP}, the algorithm allows for generalization as a methodology rendering any global model-agnostic (or model-specific in the case of the proxy classifier) explanation technique applicable to explain survival patterns of homogeneous sub-populations. Comparable levels of explanatory efficacy can be attained with simple feature importance algorithms, as the unique capabilities of SHAP to yield local feature-level attributions grounded on game-theoretical foundations, remains underutilized.

Krzyziński et al. introduce \emph{SurvSHAP(t)}, which is a direct modification of SHAP designed for generating time-dependent explanations for any functional output of a (machine learning) survival model. \emph{SurvSHAP(t)} extends the foundational SHAP properties rooted in game theory, including local accuracy, missingness, and consistency, to accommodate the time-dependent nature of the explanation. Explanations are both computed and evaluated based on the survival function. The algorithm generates the \emph{SurvSHAP(t)} functions $[\phi_{t^{min}}(\mathbf{x}^{\star},j),\dots,\phi_{t^{max}}(\mathbf{x}^{\star},j)]$ for every predictor $j \in \{1,\dots,p\}$, pertaining to the observation of interest $\mathbf{x}^{\star}$ across all timepoints $t^o \in \{t^{\min},\dots,t^{\max}\}$. These functions measure the time-dependent feature attributions with regard to the model`s prediction. \emph{SurvSHAP(t)} estimation is implemented by means of the Shapley sampling values algorithm and KernelSHAP. In the Shapley sampling values algorithm the contribution of predictor $j$ in timepoint $t^o$ is calculated as
\begin{align}\label{eq:SHAP-sampling-values}
    \phi_{t^o}(\mathbf{x}^{\star},j) = \frac{1}{|\Pi|} \sum_{\pi \in \Pi} e_{t^o,\mathbf{x}^{\star}}^{\pi(j)} - e_{t^o,\mathbf{x}^{\star}}^{\pi(j-1)} \; \text{,}
\end{align}
\noindent
where $\Pi$ is a set of all permutations of feature index set $\{1,\dots,p \}$, $\Pi(j-1)$ denoting a subset of features occurring before $j$ in the ordering $\pi \in \Pi$ and $\Pi(j)$ the equivalent subset including $j$.  The term $e_{t,\mathbf{x}^{\star}}^A = \mathbb{E}_{\mathbf{X}_{-A}}[\hat{S}(t|\mathbf{X}_A=\mathbf{x}_A^{\star}, \mathbf{X}_{-A})]$ signifies the expected value corresponding to a conditional distribution with the conditioning applied to some feature set $A$. More specifically, conditioning is applied to all features corresponding to the index set $A$. For KernelSHAP, sample coalitions $z_j \in \{0,1\}^{p}, j \in \{1,2,\dots,l\}$ must be defined, such that $Z$ is the matrix of all binary vectors. A value of $1$ indicates the presence of the corresponding feature $j$ in a coalition. Additionally, a mapping function $h_x : \{0, 1\}^p \rightarrow \mathbb{R}^p$ is employed to convert binary vectors into the original input space ($1$ represents the original value). \emph{SurvSHAP(t)} is estimated as
\begin{align}\label{eq:KernelSHAP}
    \Phi = (Z^{\intercal}WZ)^{-1}Z^{\intercal}WY  \; \text{,}
\end{align}
where $W$ is the diagonal matrix containing Shapley kernel weights for each binary vector $z$.

Another addition to the survival SHAP universe is \emph{median-SHAP}\cite{ter-minassian2024}, a straightforward adaption of traditional SHAP values to non-inferential survival models. Non-inferential survival models are essentially equivalent to a classic machine learning setting, since a one-dimensional output (usually survival time) is predicted, hence the original SHAP value methodology can be used. Conventional SHAP values are defined in relation to the average prediction, which the authors argue obstructs interpretability in a survival setting, as the average prediction is usually not a plausible instance (see the discussion in Section~\ref{sec:counterfactuals}). Therefore they propose to use the median prediction as a reference value, which by definition is plausible due to aligning with a real instance from the given dataset. Likewise, both summarizing the distribution over model outcomes using expected values (see Equation~\ref{eq:SHAP-sampling-values}) as well as using the sample means as Monte-Carlo estimators for these expected values is criticized in the survival setting, as both can be problematic in case of a skewed outcome distribution leading to misleading interpretations. Outcome distributions are often right-skewed for survival time predictions models, since there are usually observations experiencing the event after study time. These issues can be mitigated by replacing the mean of the expected change in the model outcome for a coalition from the conventional SHAP value definition with the median. The \emph{median-SHAP} approach is  severely limited due to only being applicable to models predicting (median) survival times, which are rarely used in practice. Furthermore, the desirable properties of conventional interventional SHAP values, i.e. local accuracy, missingness, and consistency, no longer hold after the proposed adjustments. 

\subsubsection{Model-Specific Local Methods}

This paper puts its main emphasis on discussing model-agnostic explanation methods. An equally extensive exploration of model-specific techniques is beyond the scope of this paper. In this section, model-specific methods for the two most important machine learning model classes for survival analysis -- tree-based models and neural networks -- are reviewed briefly. 

In the context of survival data, the most pertinent IML techniques for neural networks are feature attribution methods. These methods consolidate various local IML approaches, assigning to each feature the contribution to a specified target variable in the model.\footnote{Note, that this definition is not limited to neural network specific-methods, SHAP, Shapley values or LIME are also considered feature attribution methods.} One of the first methods for neural network interpretability is the (vanilla) gradient method or, especially in the computer vision domain, famously knows as saliency maps.\cite{simonyan2013} It simply computes the gradients of the chosen prediction with respect to the input features. This computation yields a feature-specific indication of the degree to which a slight alteration in the input variable affects the prediction. Hence vanilla gradient can be termed a feature influence or feature importance method, but not a feature attribution method, which would require the sum of the computed gradients to result in the original prediction. The gradient × input (G×I) method\cite{shrikumar2017} was introduced as a straightforward extension to vanilla gradient, in which the gradient is multiplied with the corresponding input value. It takes into consideration the magnitudes of the input features reflecting their relative importance with regard to the prediction, whereas the vanilla gradient treats all features equally in terms of their impact on the model's output. The mathematical motivation behind using the G×I for feature attribution is a first-order Taylor approximation applied to the prediction. Thereby, it is decomposed into its feature-wise effects by multiplying the gradients with the corresponding input values. In turn, summing these  effects over all features, approximately leads to the prediction. Therefore it is considered one of the first feature attribution methods for neural networks. An additional extension of the conventional gradient methodology are smoothed gradients (SmoothGrad)\cite{smilkov2017}. Their primary objective is to mitigate potential fluctuations or abrupt alterations in gradients resulting from various non-linear activation functions. SmoothGrad achieves this by computing gradients from stochastically perturbed input replicas and subsequently averaging them to derive the mean gradient. The rationale behind adding  noise in the input is that it helps to smooth out variations in the gradients, providing a more reliable estimate of feature relevance. The layer-wise relevance propagation (LRP) method\cite{bach2015} is a backpropagation-based method to compute layer-wise relevance scores. LRP performs a systematic layer-wise redistribution of upper layer relevance, commencing from the output layer and iteratively progressing until the input layer is reached. This process relies on the weights and intermediate values of the network. The summation of all LRP relevances approximately mirrors the model's prediction, resembling G×I. There are different rules to distribute the upper-layer relevances to the preceding layers, such as the simple rule, the $\alpha-\beta$-rule or the $\epsilon$-rule. DeepLIFT\cite{shrikumar2017} (deep learning important features) is another backpropagation-based approach. It represents the contribution of each feature to the difference in a target neuron's activation between the actual input and the reference input. The target neuron is usually chosen as the output neuron corresponding to the prediction to be explained. The reference input typically represents a baseline or neutral state. It could be an all-zero input or an input chosen based on domain knowledge. As with LRP, the contribution scores are distributed to individual input neurons using specific rules (e.g. path-specific, rescale or linear rules) in a backward propagation process. DeepLIFT offers the flexibility to separately account for the impacts of positive and negative contributions within nonlinearities.

Cho et al.\cite{cho2023} have successfully used DeepLIFT for a meta-learning algorithm with Cox hazard loss for multi-omics cancer survival analysis, discovering correlations between DeepLIFT feature importance scores and gene co-enrichment. The integrated gradients method\cite{sundararajan2017} combines the desirable properties of gradient based methods (sensitivity\footnote{Sensitivity: Non-zero attribution to feature, if two inputs differ only in that feature but produce different predictions.}) with those of backpropagation based methods (implementation invariance\footnote{Implementation invariance: Attributions of two functionally equivalent networks should be equivalent. Two networks are functionally equivalent if their outputs are equal for equal inputs regardless of implementation specifics.}). They are defined as the path integral of the gradients along a straight-line path from some baseline (usually chosen such that the network prediction is close to zero) to a selected input. 

Beyond feature attribution methods, there are other techniques for extracting insights from neural network models. Perturbation-based approaches discern prediction-sensitive regions within input images by rendering individual or small pixel regions uninformative, e.g. by masking, altering, or conditional sampling followed by the quantification of resultant changes in prediction probabilities. Moreover, there are both feature attribution and perturbation-based methods that work only on convolutional neural networks, which are primarily tailored for image analysis, for instance gradient-weighted class activation maps (Grad-CAMs)\cite{selvaraju2017}, meaningful perturbation\cite{fong2017} or guided backpropagation\cite{springenberg2015}. In addition, neural network specific methods for Shapley value computation have been developed, such as expected gradients\cite{erion2021}, a SHAP-based version of integrated gradients, or DeepSHAP\cite{lundberg2017}, which uses the per node attribution rules from DeepLIFT to approximate Shapley values.

With the exception of DeepLIFT, to the best of our knowledge, none of these techniques has hitherto demonstrated successful application in survival neural networks. The most relevant local method for tree interpretability is a model-specific version of KernelSHAP, on which \emph{SurvSHAP(t)} is based, as discussed in Section~\ref{shapley}. TreeSHAP is a variant of the Shapley values algorithm, specifically designed for the computation of feature attributions in tree-based models\cite{lundberg2020}. The formal extension to survival outcomes is a direction for future work.  

\subsection{Global Methods}

\subsubsection{Partial Dependence Plots (PDPs)}\label{sec:pdp}
The partial dependence function quantifies the average changes in prediction induced by feature changes.\cite{friedman2001} In this section, we present a formal adaption of the partial dependence concept to survival analysis. For a subset of feature corresponding to the indices $A \subset \{1,\dots,p\}$ (again for notational simplicity we assume $|A| = 1$) and for prediction function $f(\cdot)$ it is defined as 
\begin{align}\label{eq:PDF-theoretical}
    f_{\text{PDP},A}(\mathbf{t}^{\text{ord}}) = f_{\text{PDP}}(\mathbf{t}^{\text{ord}}|\mathbf{x}_A) = \mathbb{E}_{X_{-A}}[f(\mathbf{t}^{\text{ord}}|\mathbf{x}_A, X_{-A}] = \int f(\mathbf{t}^{\text{ord}}|\mathbf{x}_A, X_{-A}) \ d\mathbb{P}(X_{-A}) \; \text{.}
\end{align}
It computes the average outcome for a fixed subset of features ($\mathbf{x}_A$) with the remaining features ($\mathbf{x}_{-A}$) varying over their marginal distribution ($d\mathbb{P}(X_{-A})$). Thereby, the partial dependence function quantifies the relationship between the feature set of interest and the outcome over each timepoint in $\mathbf{t}^{\text{ord}}$. 

In practice, both the true outcome and the marginal distribution $d\mathbb{P}(X_{-A})$ are unknown. The integral over $X_{-A}$ can be estimated by averaging over the $n$ observed training data values for a fixed value of $t$. If $\{\mathbf{x}_{-A}^{1}, \dots, \mathbf{x}_{-A}^{n}\}$ represent the different values of $X_{-A}$ observed in the training data, then for $\mathbf{t}^{\text{ord}} = \{t^{\min},\ldots,t^{\max}\}$, and a set of grid values $\mathbf{x}_A^{\text{grid}} = \{ x_A^{1}, \dots, x_A^{g} \}$ equivalent to Section~\ref{sec:ice}. Equation~\ref{eq:PDF-theoretical} can be estimated as 
\begin{align} \label{eq:PDP}
    \hat{f}_{\text{PDP},A}(\mathbf{t}^{\text{ord}}) = \hat{f}_{\text{PDP}}(\mathbf{t}^{\text{ord}}|\mathbf{x}_A) = \frac{1}{n} \sum_{i=1}^{n} \hat{f}(\mathbf{t}^{\text{ord}}|\mathbf{x}_A^{\text{grid}}, \mathbf{x}_{-A}^{i}) \; \text{.}
\end{align}
To obtain a PDP, a set of $g \times |t^{ord}|$ ordered triplets, $\left\{\left\{ (x_A^{k},t^{o}, \hat{f}_{\text{PDP}}(t^{o}|x_A^{k})) \right\}_{k=1}^{g}\right\}_{o=\min}^{\max}$ needs to be computed. Note that, depending on the analysis, an equidistant grid of few representative values within the range of the set $\mathbf{t}^{\text{ord}}$ may suffice. The resulting PDP, reflects the average prediction over the $n$ observations as a function of a feature of interest and time. Therefore, it is simply the aggregation of the ICE curves over the $n$ observations at every feature value grid point. Ideally, ICE and PDP plots should be plotted together to avoid obfuscation of heterogeneous effects and interactions. 

We use the same simulation setting as in Section~\ref{sec:ice} and Appendix~\ref{app1:sim1} to add the corresponding PD curves by averaging over the ICE curves plotted in Figure~\ref{fig:ICE_uc_sim}. Although both the \texttt{coxph} and \texttt{ranger} models perform similarly in terms of Brier score, as illustrated in Figure~\ref{fig:brier_sim} in Appendix~\ref{app1:sim1}, only the more flexible \texttt{ranger} model is capable of accurately capturing the non-proportional hazards induced by the time-dependent \texttt{treatment} feature. This is evident from the PD curves, where the curve for the \texttt{treatment} group crosses that of the no \texttt{treatment} group at $t<3$, indicating a higher predicted probability of survival for treated patients before the cutpoint, and a lower predicted probability thereafter.  In contrast, the \texttt{coxph} model, constrained by its proportional hazards assumption, fails to capture this time-varying effect, as reflected in the shape of its PD curves. This example highlights how the use of IML methods can help prevent researchers from relying on biased models that may arise from an over-reliance on performance score comparisons alone.
\begin{figure}[htb!]
  \centering
  \includegraphics[width = 0.9\linewidth]{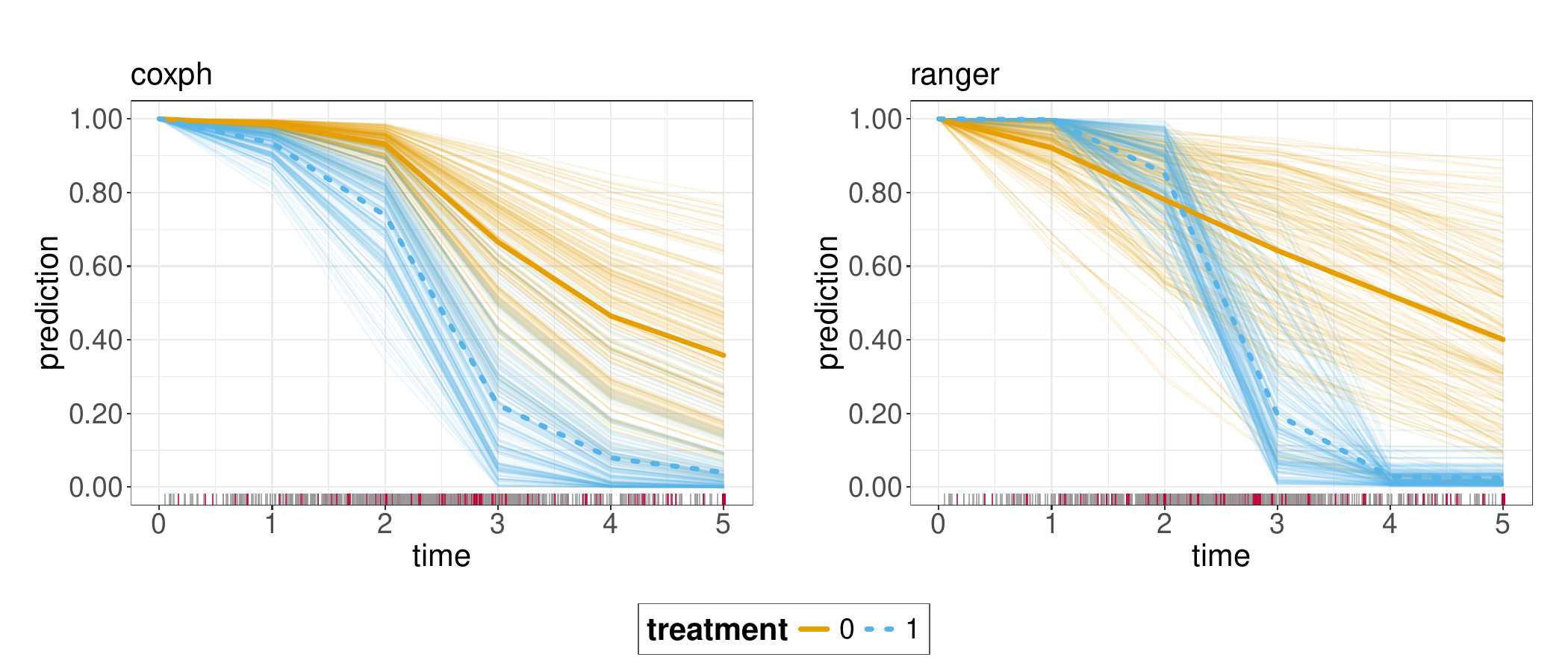}
  \caption{Uncentered PDPs and ICE curves for the \texttt{coxph} model (left) and the \texttt{ranger} model (right) for the time-dependent \texttt{treatment} feature. ICE curves are depicted as thin colored lines, while PDPs are thick colored lines. The different line colors correspond to different \texttt{treatment} strategies (0 = no \texttt{treatment}, 1 = \texttt{treatment}). The PDPs depict the average predicted survival probability within one \texttt{treatment} arm over time. The rug on the x-axis shows the survival time distribution with the grey bars indicating observed survival times and the red bars indicating censoring.}
  \label{fig:pdp_c_sim}
\end{figure}
As for the ICE curves, standard PDPs use one feature of interest. The grid value definition follows the same rationale as for the ICE plots. The c-PDP curves corresponding to the c-ICE curves shown in Figure~\ref{fig:ICE_c_sim} can be found in Figure~\ref{fig:pdp_c_sim} in Appendix~\ref{app1:sim1}. Due to the lack of feature interaction in the given simple simulation example little additional information is obtained from centering the PDPs. Centered PDP plots, i.e. c-PDPs, are obtained by aggregating the c-ICE curves over the $n$ observations:
\begin{align} \label{eq:c-PDP}
    \hat{f}_{\text{c-PDP},A}(\mathbf{t}^{\text{ord}}) = \hat{f}_{\text{c-PDP}}(\mathbf{t}^{\text{ord}}|\mathbf{x}_A) = \frac{1}{n} \sum_{i=1}^{n} \left [ \hat{f}(\mathbf{t}^{\text{ord}}|\mathbf{x}_A^{\text{grid}}, \mathbf{x}_{-A}^{i}) - \hat{f}(\mathbf{t}^{\text{ord}}|x^{\prime}, \mathbf{x}_{-A}^{i}) \right ] \; \text{.}
\end{align}
It is further possible to reduce the dimensionality to match classic PDPs and ICE-T curves by marginalizing over time, analogously we refer to these as PDP-T curves. In the simplest form we can average over $t^o \in  \{t^{\min}, \dots, t^{\max}\}$:
\begin{align}\label{eq:PDP-T}
    \hat{f}_{\text{PDP-T}}(\mathbf{x}_A) = \frac{1}{|\mathbf{t}^{\text{ord}}|} \sum_{o=\min}^{\min} \frac{1}{n} \sum_{i=1}^{n} \hat{f}(t^o|\mathbf{x}_A^{\text{grid}}, \mathbf{x}_{-A}^{i}) \; \text{.}
\end{align}
In combination with ICE plots, PDP's offer an intuitive and easily interpretable unification of both local and global feature effects. This facilitates insights into the direction and magnitude of the effects of features on the predicted survival outcome, allows for the detection of complex patterns and interactions and helps evaluating the reliability of the model and detecting potential biases or overfitting.\citep{jakobsen2020applying,petch2022opening,zeldin2023exposure}

\subsubsection{Accumulated Local Effects (ALE) Plots}

\begin{figure}[htb!]
  \centering
  \includegraphics[width = 0.9\linewidth]{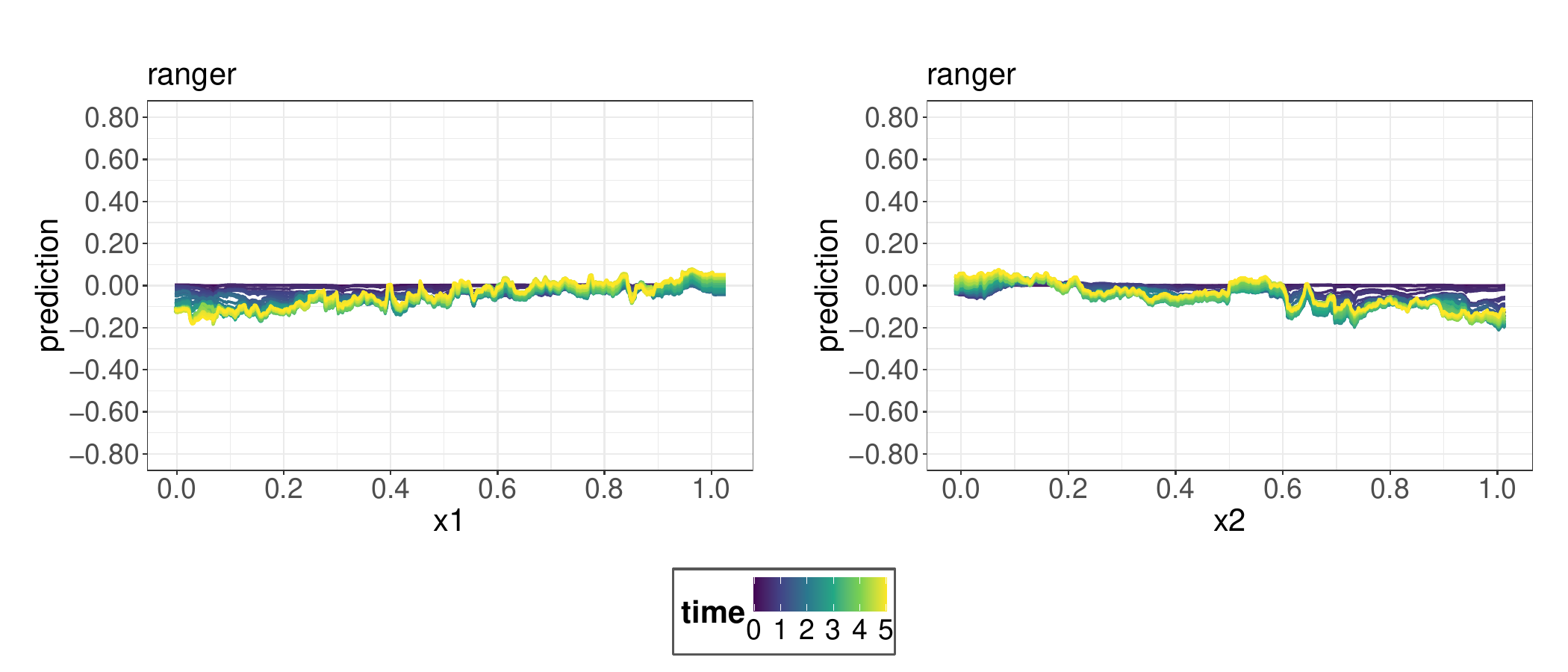}
  \caption{PDPs for the \texttt{ranger} model for feature $\texttt{x}_1$ (left) and feature $\texttt{x}_2$ (right). The different line colors correspond to different survival times, while the feature values are shown on the x-axes. Contrary to the ground truth, the PDPs suggest that the average predicted survival probability barely differs for different feature values, for both feature $\texttt{x}_1$ and feature $\texttt{x}_2$.}
  \label{fig:pdp_comp_sim}
\end{figure}

PDPs implicity assume the independence of the feature of interest of all other features for quantifying and visualizing its main effect on the target. If correlated features are present, the marginal effects shown in the PDP may not accurately reflect the true relationship between target and feature of interest.\cite{hooker2007} This is because, in this case, the integral over the marginal distribution of $X_{-A}$ estimated in Equation~\ref{eq:PDF-theoretical} requires reliable extrapolation beyond the scope of the training data, which cannot be guaranteed by nature of most supervised machine learning models. Consider a straightforward simulated example using a standard Weibull survival model with two time-independent, numerical features (i.e. $\texttt{x}_1$ has a negative log hazard ratio and $\texttt{x}_2$ with a positive log hazard ratio).  These features are highly correlated and exert equally strong, yet opposing, effects on the hazard. A practical scenario for this type of simulation could be a medical study examining the survival of patients for a chronic disease, such as cancer, where specific biomarkers or clinical measures, which in practice are often highly correlated, play crucial, but opposing roles. We simulate data for $N = 3000$ patients with a maximum follow up time of five years and additional random censoring, for further details on the simulation, see Appendix~\ref{app1:sim2}. After fitting a \texttt{ranger} model to the simulated data, PDPs are generated on the test set for both features, as shown in Figure~\ref{fig:pdp_comp_sim}. Although the data generating process establishes strong opposing effects of the features on survival, the PDPs indicate that the respective positive and negative impact of $\texttt{x}_1$ and $\texttt{x}_2$ on the average predicted survival probability is rather small (i.e., $<0.2$ in absolute value). This occurs because, when computing the PDP, to calculate the effect of $\texttt{x}_1$ at a specific value, averaging over predictions for all possible values of $\texttt{x}_2$ is required. However, due to the strong correlation between the features, many of these value combinations are not naturally observed in the data. 
Marginal plots (M-plots)\cite{friedman2008,apley2020} substitute the marginal density with the conditional density to avoid such extrapolation:
\begin{align}\label{eq:Mplot-theoretical}
    f_{M,A}(\mathbf{t}^{\text{ord}}) = f_{M}(\mathbf{t}^{\text{ord}}|\mathbf{x}_A) = \mathbb{E}_{X_{-A}}[f(\mathbf{t}^{\text{ord}}|X_A, X_{-A}) | X_A = \mathbf{x}_A] = \int_{-\infty}^{\infty} f(\mathbf{t}^{\text{ord}}|\mathbf{x}_A, X_{-A}) \ d\mathbb{P}(X_{-A}|X_{A} = \mathbf{x}_A) \; \text{,}
\end{align}
where $f_{M,A}(\mathbf{t}^{\text{ord}}|\mathbf{x}_A)$ can be crudely estimated as 
\begin{align}\label{eq:Mplot-estimate}
    \hat{f}_{M,A}(\mathbf{t}^{\text{ord}}) = \hat{f}_{M}(\mathbf{t}^{\text{ord}}|\mathbf{x}_A) = \frac{1}{n(\mathbf{x}_A)} \sum_{i \in N(\mathbf{x}_A)} \hat{f}(\mathbf{t}^{\text{ord}}|\mathbf{x}_A, \mathbf{x}_{-A}^{i}) \; \text{.}
\end{align}
For $\mathbf{t}^{\text{ord}}$ and $N(\mathbf{x}_A) \subset \{1,\dots, n\}$, the subset of instances for which $x_A^{i}$ falls into some small neighborhood of $\mathbf{x}_A$ and $n(\mathbf{x}_A)$ are the number of observations in that neighborhood. While M-plots solve the extrapolation issue, conditioning on $\mathbf{x}_A$ introduces the so-called omitted variable bias\cite{apley2020}, if there are at least two correlated features which both are associated with the target. In this case, $\hat{f}_{M,A}(\mathbf{t}^{\text{ord}}|\mathbf{x}_A)$ estimates the combined effect of the correlated features on the target at different observed survival timepoints. 

\begin{figure}[htb!]
  \centering
  \includegraphics[width = 0.9\linewidth]{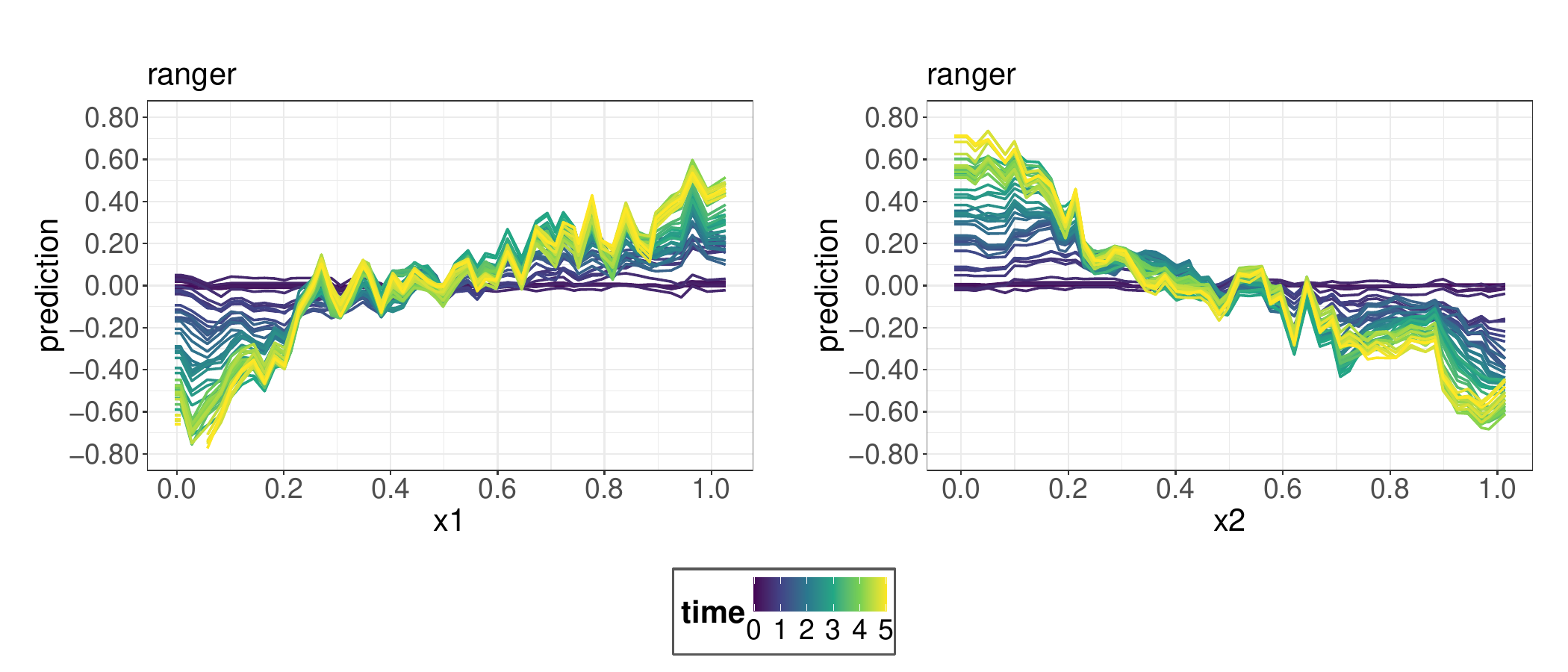}
  \caption{ALE plots for the \texttt{ranger} model for feature $\texttt{x}_1$ (left) and feature $\texttt{x}_2$ (right). The different line colors correspond to different survival times, while the feature values are shown on the x-axes. The ALE plots correctly identify that feature $\texttt{x}_1$ has a strong positive and feature $\texttt{x}_2$ a strong negative effect on the average predicted survival probability.}
  \label{fig:ale_comp_sim}
\end{figure}

Accumulated local effects plots (ALE)\cite{apley2020} extend M-plots to obtain an unbiased estimate for the main effects of a feature of interest. The basic idea consists of three parts. The first part constitutes removing unwanted effects of other features by first taking partial derivatives (local effects)  of the prediction function with regard to the feature of interest. Next, the local effects are averaged over the conditional distribution similar to M-plots to avoid the extrapolation issue. Finally, the averaged local effects are integrated (accumulated) with regard to the same feature of interest to estimate its global main effect, obviating the omitted variable bias. Figure~\ref{fig:ale_comp_sim} presents the ALE plots for the previously discussed simulation example. Contrary to the complementary PDPs (Figure~\ref{fig:pdp_comp_sim}), the ALE procedure accurately captures the respective strong positive and negative effects of features $\texttt{x}_1$ and $\texttt{x}_2$ on the predicted probability of survival. Additionally, we observe that these effects intensify with increasing survival times, which is an artifact of the natural shape of the survival curves.
This section constitutes the first formal demonstration of M-plots and ALE plots in the context of survival analysis. Let the support of $\mathbf{x}_A$ be the interval $\mathcal{S}_A = [\min(\mathbf{x}_A),\max(\mathbf{x}_{A})]$. For each $k = 1,\dots, g$ let $\mathcal{P}_A \equiv \{z_{A}^k: k = 1,\dots,g \}$ be a partition of $\mathcal{S}_A$ into $g$ intervals with $q_{A}^0 = \min(\mathbf{x}_A)$ and $q_{A}^g = \max(\mathbf{x}_A)$. For any $x^{\star} \in \mathcal{S}_A$, define $k_A(x^{\star})$ to be the index of the interval of $\mathcal{P}_A$ into which $x^{\star}$ falls, i.e. $x^{\star} \in (q^{k-1}_A,q^k_A]$ for $k = k_A(x^{\star})$. Assuming $f(\cdot)$ is differentiable, the uncentered first order ALE main effect for the set $\mathbf{t}^{\text{ord}}$ is defined as
\begin{align}\label{eq:ALE-uncentered-theoretical}
    \tilde{f}_{\text{ALE},A}(\mathbf{t}^{\text{ord}}|x^{\star}) = \int_{q^0_A}^{x^{\star}} \mathbb{E}_{X_{-A}|X_A = \mathbf{x}_A} \left(\frac{\partial \hat{f}(\mathbf{t}^{\text{ord}}|X_A,X_{-A})}{\partial X_A} \ \big| \ X_A = \mathbf{q}_A\right) \ d\mathbf{q}_A \; \text{.}
\end{align}
Usually the centered ALE is considered, which has a zero mean regarding the marginal distribution of the feature of interest:
\begin{align}\label{eq:ALE_centered_theoretical}
    f_{\text{ALE},A}(\mathbf{t}^{\text{ord}}|x^{\star}) = \tilde{f}_{\text{ALE},A}(\mathbf{t}^{\text{ord}}|x^{\star}) - \int_{-\infty}^{\infty} \tilde{f}_{\text{ALE},A}(\mathbf{t}^{\text{ord}}|x^{\star}) \ d \mathbb{P}(X_A) \; \text{.}
\end{align}
To estimate the first order ALE effects, a sufficiently fine fixed partition $\{N_A(k) = (q^{k-1}_A,q_{A}^k]: k = 1,\dots,g\}$ of the sample range of the support $\mathcal{S}_A$ is chosen. Customarily, the quantile distribution of $X_A$ is used to create $g$ intervals, with $g - 1$ quantiles as interval bounds $q_{A}^0, \dots, q^{g-1}_A,q_{A}^g$. The lower bound $q_{A}^0$ is set just below the minimum observation $\min(\mathbf{x}_A)$, and $q_{A}^g$ is set as the maximum observation $\max(\mathbf{x}_A)$. The local effect of $\mathbf{x}_A$ within each interval is estimated by averaging all observation-wise finite differences. For some $\mathbf{x}^{i} = (x^{i}_A,\mathbf{x}^{i}_{-A})$, for which $x_A^{i} \in (q^{k-1}_A,q_{A}^k]$ the finite difference is computed as $\hat{f}(\mathbf{t}^{\text{ord}}|q_{A}^k,\mathbf{x}_{-A}^{i}) - \hat{f}(\mathbf{t}^{\text{ord}}|q_{A}^{k-1},\mathbf{x}_{-A}^{i})$. In this way, the conditional expectations (inner integrals over the local effects) are approximated using sample averages across $\{\mathbf{x}_{-A}^{i}: i = 1,\dots,n \}$ conditional on $x_{A}^{i}$ falling into the corresponding interval of the partition. To estimate the outer integral, the local effects are accumulated up to the point of interest $x^{\star}$. 

Let $n^i_j(k)$ be the number of training observations located in interval $N^i_j(k)$, such that $\sum_{k=1}^g n^i_j(k) = n$. Then the formula to estimate the uncentered first order ALE at point $x^{\star}$ for the set $\mathbf{t}^{\text{ord}}$ is denoted as
\begin{align}\label{eq:ALE-uncentered-estimate}
    \hat{\tilde{f}}_{\text{ALE},A}(\mathbf{t}^{\text{ord}}|x^{\star}) = \sum_{k=1}^{k_A(x^{\star})} \frac{1}{n_A^i(k)} \sum_{i:x_A^{i} \in N^i_A(k)} \left[\hat{f}(q_{A}^k,\mathbf{x}_{-A}^{i}) - \hat{f}(q^{k-1}_A,\mathbf{x}_{-A}^{i})\right] \; \text{.}
\end{align}
Analogously to Equation~\ref{eq:ALE_centered_theoretical}, the centered first order ALE effect for a fixed $t$ is obtained by subtracting an estimate of the average uncentered first order ALE
\begin{align}\label{eq:ALE-centered-estimate}
    \hat{f}_{\text{ALE},A}(\mathbf{t}^{\text{ord}}|x^{\star}) = \hat{\tilde{f}}_{\text{ALE},A}(\mathbf{t}^{\text{ord}}|x^{\star}) - \frac{1}{n} \sum_{i=1}^{n} \hat{\tilde{f}}_{\text{ALE},A}(\mathbf{t}^{\text{ord}}|x_A^{i}) \; \text{.}
\end{align}
Similar to ICE and PDP, ALE plots can be marginalized over time either by averaging or summing over the unique, ordered observed survival times. For instance, the uncentered first order ALE-T effect would then be estimated as follows
\begin{align}\label{ALE-uncentered-estimate}
    \hat{\tilde{f}}_{\text{ALE-T},A}(x^{\star}) = \sum_{k=1}^{k_A(x^{\star})} \frac{1}{n_A^i(k)} \sum_{i:x_A^{i} \in N^i_A(k)} \left[ \frac{1}{|\mathbf{t}^{\text{ord}}|} \sum_{o=\min}^{\max} \hat{f}(t^o|q^k_A,\mathbf{x}_{-A}^{i}) - \hat{f}(t^o|q^{k-1}_A,\mathbf{x}_{-A}^{i})\right] \; \text{.}
\end{align}
Essentially, PDP and ALE offer two different methods of quantifying the main effect of a feature of interest at a certain value. Using a set of grid values, the corresponding ALE values $\hat{f}_{\text{ALE},A}(\mathbf{t}^{\text{ord}}|\mathbf{x}^{\text{grid}}_A)$ can be plotted just like the PDP. The values $\hat{f}_{\text{ALE},A}(\mathbf{t}^{\text{ord}}|\mathbf{x}^{\text{grid}}_A)$ are estimates of the change in the prediction, when the feature of interest take the values $\mathbf{x}^{\text{grid}}_A$ compared to the average predicted survival probability at all different ordered observed survival times $\mathbf{t}^{\text{ord}}$. For interpreting the ALE-T curves in case $f(\cdot) = S(\cdot)$, the difference between average predicted probability of survival until time $t^{o}$ and over all observations $n$ needs to be kept in mind. When averaging over time, the ALE values are interpreted as the change in average predicted survival probability over time, compared to the average predicted survival probability over time and over observations. Correspondingly, when summing over time, the ALE values represent the change in average predicted survival time with respect to the average predicted survival time. 

Since ALE plots accumulate effects in a certain direction, the feature values of the feature of interest need to have a natural order. This is the case for numerical features and ordinal categorical features, but not all categorical features in general. The most commonly used method of instilling an order to categorical features is ranking them according to their similarity based on the remaining features.\cite{apley2020} Typically, the feature-wise distances are computed using the Kolmogorov-Smirnov distance (for numerical features) or relative frequency tables (for categorical features). After obtaining distances between all categories, multi-dimensional scaling is used reduce the distance matrix to a one-dimensional distance measure based on which the order of the categories is determined.\cite{molnar2022} 

\subsubsection{Feature Interaction}

\begin{figure}[htb!]
  \centering
  \includegraphics[width = 0.9\linewidth]{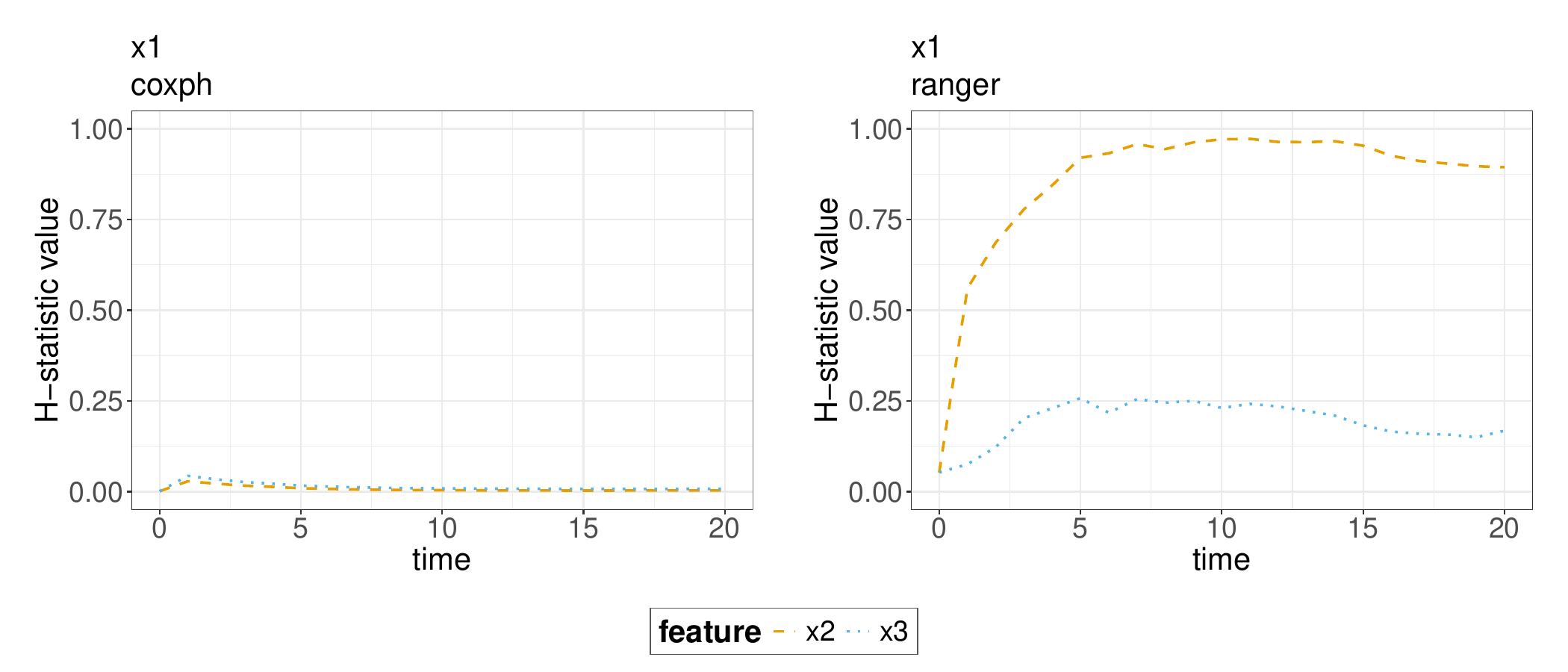}
  \caption{The 2-way interaction strengths ($H_{jk}(t)$-statistics) curves for the \texttt{coxph} model (left) and the \texttt{ranger} model (right) for the feature $\texttt{x}_1$. The interaction at each timepoint is the proportion of variance explained by the interaction. No interactions are found for the \texttt{coxph} model, while for the \texttt{ranger} model strong interactions are found between features $\texttt{x}_1$ and $\texttt{x}_2$.}
  \label{fig:inter_sim}
\end{figure}

Feature interaction refers to the the collaborative effect of two or more features on the prediction after removing their individual main effects. A common way to estimate interaction strength are Friedman's H-statistics\cite{friedman2008}. In this section we focus on two-way interaction measures and a total interaction measure, which we formally adapt to survival analysis. These two measures are of primary interest in most applications, yet the H-statistic can be generalized to any type of interaction. The two-way interaction H-statistics measures the interaction strength between two features $a$ and $b$ for the vector of ordered observed survival times $\mathbf{t}^{\text{ord}}$:
\begin{align}\label{eq:H-2}
    H_{ab}^2(\mathbf{t}^{\text{ord}}) = \frac{\sum_{i=1}^{n} \left[ \hat{f}_{\text{PDP},\{a,b\}}(\mathbf{t}^{\text{ord}}|x_a^{i},x_b^{i}) - \hat{f}_{\text{PDP},a}(\mathbf{t}^{\text{ord}}|x_a^{i}) - \hat{f}_{\text{PDP},b}(\mathbf{t}^{\text{ord}}|x_b^{i}) \right]^2}{\sum_{i=1}^{n} \hat{f}_{\text{PDP},\{a,b\}}^2(\mathbf{t}^{\text{ord}}|x_a^{i},x_b^{i}) } \; \text{.}
\end{align}
This is simply an estimator for the amount of the  variance of the output of the partial dependence explained by the two-way interaction. The partial dependence functions in this case are assumed to be centered at zero and normalized. If $\mathbf{x}_a$ and $\mathbf{x}_b$ do not interact, $H_{ab}^2(\mathbf{t}^{\text{ord}})$ is zero, since the two-way partial dependence function can be decomposed as $\hat{f}_{\text{PDP},\{a,b\}}(\mathbf{t}^{\text{ord}}|x_a^{i},x_b^{i}) = \hat{f}_{\text{PDP},a}(\mathbf{t}^{\text{ord}}|x_a^{i}) + \hat{f}_{\text{PDP},b}(\mathbf{t}^{\text{ord}}|x_b^{i})$. If the interaction statistic is one, the individual partial dependence functions are constant $\hat{f}_{\text{PDP},a}(\mathbf{t}^{\text{ord}}|x_a^{i}) = \hat{f}_{\text{PDP},b}(\mathbf{t}^{\text{ord}}|x_b^{i})$, meaning only the interaction affects the predicted survival probability. Typically, $H_{ab}^2(t) \in [0,1]$, yet in rare cases the H-statistic can be greater than 1. This occurs if the variance of the two-way interaction exceeds the variance of the two-dimensional partial dependence. To illustrate how the the 2-way interaction H-statistic can be used to identify interactions in fitted machine learning models, we simulate data from a standard exponential survival model, with three numerical features of interest: $\texttt{x}_1$, $\texttt{x}_2$, $\texttt{x}_3$. Each of these features has a moderately strong negative effect on the hazard, while features $\texttt{x}_1$ and $\texttt{x}_2$ exhibit a strong positive interaction effect on the hazard. A practical example for such an interaction could involve biomarkers for chemotherapy sensitivity and radiation sensitivity with each marker individually improving cancer survival (lowering hazard), but a strong positive interaction leading to increased hazard due to heightened toxicity when both therapies are highly effective. We simulate data for $N=3000$ patients with a maximum follow-up time of 20 months and additional random censoring (for further details on the simulation, see Appendix~\ref{app1:sim3}). We compute the 2-way interaction strengths ($H_{jk}(t)$-statistics) for the \texttt{coxph} model and the \texttt{ranger} model for the feature $\texttt{x}_1$, as shown in Figure~\ref{fig:inter_sim}. Since the \texttt{coxph} model was fitted without specifying interactions, the H-statistic values over time are consistently close to zero (not exactly zero due to estimation error). In contrast, machine learning models such as the \texttt{ranger} model do not require explicit interaction specification. The H-statistic plot indicates, that the  $\texttt{ranger}$ model correctly detects a strong interaction between $\texttt{x}_1$ and $\texttt{x}_2$, as the curve corresponding to $\texttt{x}_2$ rapidly increases to one, remaining constant at a high level. Furthermore, we can detect from the plot, that the \texttt{ranger} model erroneously identifies an interaction effect between $\texttt{x}_1$ and $\texttt{x}_3$; however, the strength of the effect detected is much smaller in absolute size ($< 0.25$).

The total interaction H-statistic measures the interaction strength between feature $a$ and all other features in set $-a$ over $\mathbf{t}^{\text{ord}}$. Let $P$ be the total set of features $\{1,\dots,p\}$, then the total interaction H-statistics is defined as: 
\begin{align}\label{eq:H-all}
    H_{a}^2(\mathbf{t}^{\text{ord}}) = \frac{\sum_{i=1}^{n} \left[ \hat{f}_{\text{PDP},P}(\mathbf{t}^{\text{ord}}|\mathbf{x}^{i}) - \hat{f}_{\text{PDP},a}(\mathbf{t}^{\text{ord}}|x_a^{i}) - \hat{f}_{\text{PDP},-a}(\mathbf{t}^{\text{ord}}|\mathbf{x}_{-a}^{i}) \right]^2}{\sum_{i=1}^{n} \hat{f}^2_{\text{PDP},P}(\mathbf{t}^{\text{ord}}|\mathbf{x}^{i}) } \; \text{.}
\end{align}
This now is an estimator for the amount of the variance of the output of the entire function explained by the total interaction. Again the numerator makes use of the fact that in case of no interaction between $x_a$ and $\mathbf{x}_{-a}$: $\hat{f}_{\text{PDP},P}(t|\mathbf{x}^{i}) = \hat{f}_{\text{PDP},a}(\mathbf{t}^{\text{ord}}|\mathbf{x}_a^{i}) + \hat{f}_{\text{PDP},-a}(\mathbf{t}^{\text{ord}}|\mathbf{x}_{-a}^{i})$. 

To evaluate the interaction strength at different timepoints, interaction plots of $|\mathbf{t}^{\text{ord}}|$ ordered pairs: $\big\{ (t^o, H^2(t^o)) \big\}_{o=\min}^{\max}$, can be generated. Generalized measures of interaction over time can be obtained by marginalizing the H-statistics over time, e.g. $H_{ab}^2 = \frac{1}{|\mathbf{t}^{\text{ord}}|} \sum_{o=\min}^{\max} H_{ab}^2(t^o)$ and $H_{a}^2 = \frac{1}{|\mathbf{t}^{\text{ord}}|} \sum_{o=\min}^{\max} H_{a}^2(t^o)$. Note, that the H-statistics are merely measures of interaction strength. To learn more about the type and shape of interactions, ICE, PDP and ALE plots with more than one feature of interest can be analyzed. 

Friedman's H statistic is a versatile and robust tool for assessing interaction effects, offering meaningful insights due to being founded on partial dependence decomposition. Therefore its utility extends beyond identifying solely linear interactions, as it is adept at detecting interactions in various forms. Furthermore, its dimensionless nature facilitates comparisons across features and even different models. Practical implementations often involve the calculation of H-statistics on a random sample of $m$ observations with $m<n$, introducing potential instability in the H-statistic estimate. Additionally, a noteworthy limitation of the H-statistic lies in the lack of a model-agnostic statistical test for determining whether an interaction is significantly greater than zero. The current testing procedure entails the reestimation of the underlying model, excluding the relevant interactions, a practice not always feasible, in particular for complex machine learning models. Furthermore, no guidelines for evaluating the strength of interactions are established, i.e., at which thresholds an interaction is considered strong, moderate or weak is ambiguous. This is amplified by the lack of a definitive upper bound of the H-statistic, which can exceed a value of 1.  

\subsubsection{Feature Importance}\label{feature_importance}

Complementary to feature effects methods, such as ICE, PDP and ALE, so-called feature importance methods aim to quantify the extent of which each feature contributes to the predictive performance of a machine learning model. In this section, the three common feature importance techniques, permutation feature importance (PFI)\cite{breiman2001}, Leave-One-Covariate-Out (LOCO)\cite{lei2018} and conditional feature importance (CFI)\cite{strobl2008,hooker2021,watson2021,blesch2023} are adapted to survival analysis.

\textbf{Permutation feature importance (PFI)}\cite{fisher2019} quantifies the importance of a specific feature of interest $j$ as the change in the model's prediction error induced by permuting $\mathbf{x}_j$ and therefore breaking its association with the target. Small or no changes in the model error imply that the model does not rely heavily on the feature for prediction, thus characterizing a feature of low importance. Important features induce large changes in the model error, since the model does depend strongly on the feature for making predictions. 

In the permutation step of permutation feature importance, the feature of interest $j$ is replaced with an independent sample from the marginal distribution $\mathbb{P}(X_j)$. Hence, if correlated features are present in the data, potentially unrealistic data points are created, leading to the model extrapolating its predictions outside the range of the original data. The main idea of \textbf{conditional feature importance (CFI)} is to resample the feature of interest from the conditional distribution, such that the joint distribution is preserved, i.e. $\mathbb{P}(X_j|X_{-j})\mathbb{P}(X_{-j}) = \mathbb{P}(X_{j},X_{-j})$. One concrete implementation of a conditional feature importance measure is the conditional predictive impact (CPI), originally developed by Watson and Wright\cite{watson2021} and extended to mixed data by Blesch et al.\cite{blesch2023}. In CPI, the conditional resampling step is executed by means of knockoff sampling algorithms. The feature of interest $j$ is replaced with a knockoff, retaining the covariance structure of the input features, while being independent of the target, conditional on $-j$, i.e., all other features apart from $j$. The algorithm is similar to that of PFI, but instead of permuting $\mathbf{x}_j$, it is replaced by a knockoff copy. The choice of the knockoff sampler depends on the type of input data, for further information see Blesch et al.\cite{blesch2023}. Furthermore, the conditional predictive impact method offers the added benefit of an integrated conditional independence test, that can be executed within the framework of paired t-tests or Fisher exact test for small samples.

\textbf{Leave-one-covariate-out (LOCO)}\cite{lei2018} is a technique used to assess the importance of individual features by fitting two separate models, one on the full feature vector $\mathbf{X}$ and one on the feature vector excluding the feature of interest $\mathbf{X}_{-j} = \mathbf{X} \setminus \{\mathbf{x}_{-j}\}$ and comparing their prediction errors. The steps of all three feature importance algorithms adapted for survival analysis are summarized and denoted in Algorithm~\ref{alg:all_FI}. 

\begin{algorithm}[h]
\caption{Feature Importance Algorithms: PFI, CPI, LOCO}
\label{alg:all_FI}
\begin{algorithmic}[1]
\Require Trained model $\hat{f}(\cdot)$, feature matrix $\mathbf{X}$, observed survival times vector $\mathbf{t}^{\text{obs}} = (t^1,\dots,t^n)^{\intercal}$, censoring vector $\boldsymbol{\delta}^{\text{obs}} = (\delta^1,\dots,\delta^n)^{\intercal}$, loss function  $L(\mathbf{t}^{\text{obs}}, \boldsymbol{\delta}^{\text{obs}}, \hat{f}(\cdot))$
\State Estimate the model loss of the original model $L(\mathbf{t}^{\text{obs}}, \boldsymbol{\delta}^{\text{obs}}, \hat{f}(\mathbf{t}^{\text{obs}}|\mathbf{X}))$
\For{each feature $j \in \{1,\dots,p\}$}
    \State  \textbf{PFI:} Generate a permuted feature vector $\tilde{\mathbf{x}}_j$ to replace $\mathbf{x}_j$ inducing the permuted feature matrix $\tilde{\mathbf{X}}_j$  \par \textbf{CPI:} Generate a knockoff $\tilde{\mathbf{x}}_j$ to replace $\mathbf{x}_j$ inducing the augmented feature matrix $\tilde{\mathbf{X}}_j$ \par
    \textbf{LOCO:} Generate feature matrix $\mathbf{X}_{-j}$ by removing feature $j$ from the original feature matrix $\mathbf{X}$ \par
    \State  \textbf{PFI \& CPI:} Compute the loss based on the predictions of the permuted feature matrix $L(\mathbf{t}^{\text{obs}}, \boldsymbol{\delta}^{\text{obs}}, \hat{f}(\mathbf{t}^{\text{obs}}|\mathbf{\tilde{X}}_j))$ \par \textbf{LOCO:} Fit a new model $\hat{f}_{-j}(\mathbf{t}^{\text{obs}}|\mathbf{X}_{-j})$ on the feature matrix $\mathbf{X}_{-j}$
    \State Calculate the FI as a difference $\text{FI}_{j}^{d} = L(\mathbf{t}^{\text{obs}}, \boldsymbol{\delta}^{\text{obs}}, \hat{f}_{-j}(\mathbf{t}^{\text{obs}}|\mathbf{X}_{-j})) - L(\mathbf{t}^{\text{obs}}, \boldsymbol{\delta}^{\text{obs}}, \hat{f}(\mathbf{t}^{\text{obs}}|\mathbf{X}))$ or quotient $\text{FI}_{j}^{q} = \frac{L(\mathbf{t}^{\text{obs}}, \boldsymbol{\delta}^{\text{obs}}, \hat{f}_{-j}(\mathbf{t}^{\text{obs}}|\mathbf{X}_{-j}))}{L(\mathbf{t}^{\text{obs}}, \boldsymbol{\delta}^{\text{obs}}, \hat{f}(\mathbf{t}^{\text{obs}}|\mathbf{X}))}$
\EndFor
\State Sort features by descending importance $\text{FI}_{j}^{d}$ or $\text{FI}_{j}^{q}$
\end{algorithmic}
\end{algorithm}

\begin{figure}[htb!]
    \centering
    \includegraphics[width = 1\linewidth]{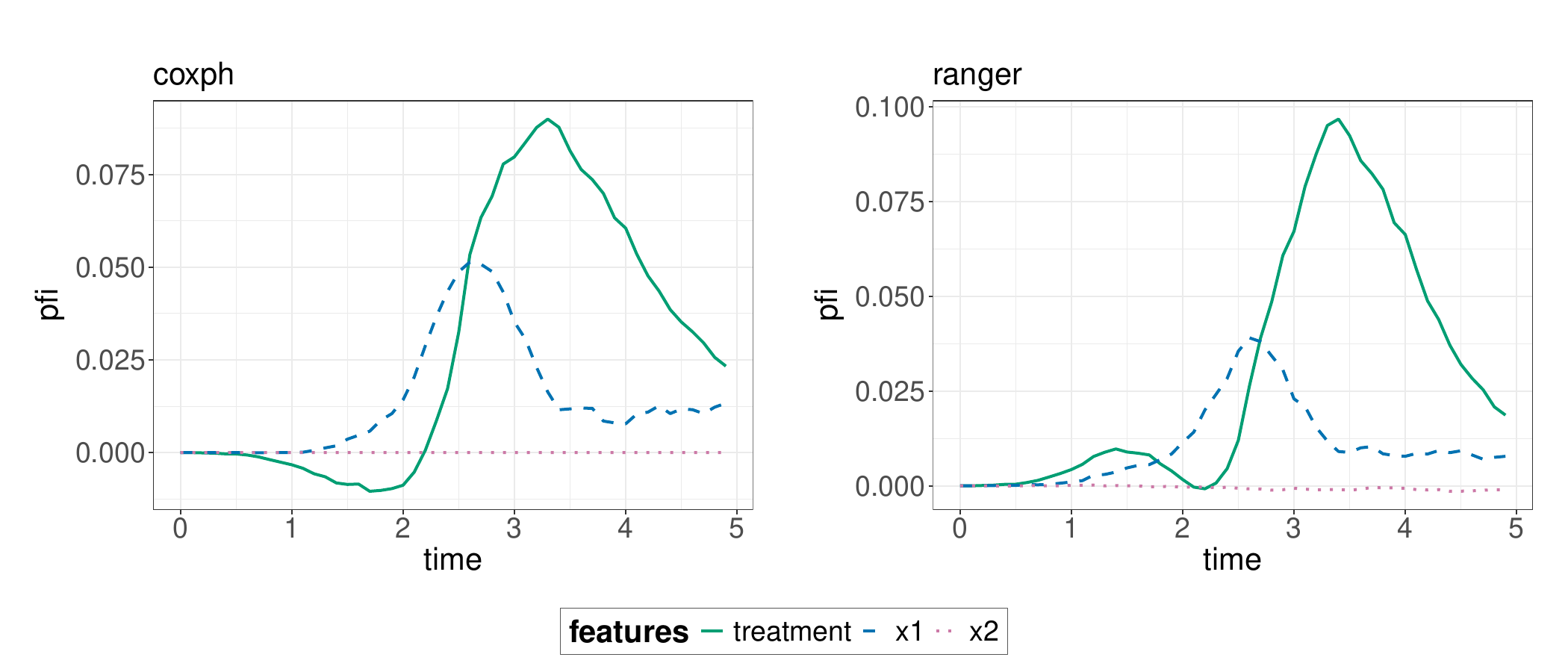}
    \caption{Permutation feature importance for the \texttt{coxph} model (left) and the \texttt{ranger} model (right). The y-axis denotes the Brier score loss after permutations of each separate feature with the loss of the full model including all features without permutations subtracted.}
    \label{fig:pfi_sim}
\end{figure}

For statistical inference, one-sided hypothesis tests of the form: 
\begin{align}\label{eq:FI-test}
    \Lambda_0 \, \text{:}  \; \text{FI}_j \leq 0 \; \; \text{versus} \; \; \Lambda_1 \, \text{:}  \; \text{FI}_j > 0
\end{align}
can be performed for the feature importance values obtained from either of the three methods presented. Originally proposed to mitigate the bias of PFI towards categorical features with many categories, the permutation importance (PIMP) algorithm\cite{altmann2010} provides p-values for the test in Equation~\ref{eq:FI-test}. The distribution of the PFI values under $\Lambda_0$ is computed by permuting the target to break the association with the features and fitting a distribution to the PFI scores computed under $\Lambda_0$. Lei et al.\cite{lei2018} rely on the asymptotic normal distribution of the mean-based LOCO FI scores to construct confidence intervals and hypothesis tests corrected for multiple testing for $j \in \{1,\dots,p\}$. To improve stability, they alternatively suggest nonparametric tests based on LOCO FI scores calculated as median decreases in loss, in the form of sign tests or the Wilcoxon signed-rank tests. For CPI scores, Watson and Wright\cite{watson2021} show that by application of the central limit theorem, one-sided t-tests can be used to evaluate Equation~\ref{eq:FI-test} when samples are sufficiently large. For small sample inference, Fisher exact tests are proposed. For further details on statistical inference, refer to the respective papers\cite{lei2018,fisher2019,watson2021}.   

The key to adapting all of the above feature importance measures for survival analysis is the use of appropriate error metrics suitable for survival models. Time-dependent performance measures such as the Brier score, the cumulative/dynamic AUC or the log-loss can be used to obtain $|\mathbf{t}^{\text{ord}}|$ ordered pairs: $\big\{ (t^o, FI_j(t^o)) \big\}_{o=\min}^{\max}$, such that the feature importance can be visualized over time. We illustrate this using computing the PFI based the same simulation setting as described in Section~\ref{sec:ice} and \ref{sec:pdp}. The importance values in Figure~\ref{fig:pfi_sim} are derived as the difference between the Brier score loss at a specified time point for the model fitted with a permuted feature matrix (for the feature of interest) and the Brier score loss of the model without any feature permutation. The (relative) importance of the features is correctly identified by both models. The permutation of $\texttt{x}_1$ does not have any effect on the Brier score, the permutation of $\texttt{x}_2$ has a moderate effect and the permutation of $\texttt{treatment}$ has the strongest effect over all timepoints. Single-value performance measures, such as integrated Brier score, log-loss or AUC as well as the concordance index produce outcomes akin to classic feature importance plots, with one feature being associated with one feature importance value.

\subsubsection{Functional Decomposition}\label{sec:functional_decomposition}

Functional decomposition is a core concept in interpretable machine learning. It refers to the decomposition of complex, high-dimensional machine learning model output functions into more understandable functional components, including feature main effects and lower-level interactions, often for the purpose of visualization. One and two-dimensional components are one and two-dimensional real-valued functions that can be plotted. In this section we delineate how some of the most popular functional decomposition approaches can be  adapted to suit survival outcomes.

Let $A \subseteq \{1,\dots,p\}$ be a feature combination. The general decomposition is then denoted as the sum of the components of all possible feature combinations,,
\begin{align}\label{eq:decomposition}
    \hat{f}(t|\mathbf{X}) = \sum_{A \subseteq \{1,\dots,p\}} \hat{f}_A(t|\mathbf{X}_A) \; \text{,}
\end{align}
where a component $\hat{f}_A(t|\mathbf{X}_A)$ only includes the features in $A$. Note, that the right-hand side of Equation~\ref{eq:decomposition} is not identified, i.e., it can be changed without changing the left-hand side. Different constraints can be imposed to uniquely identify and best reflect the contribution of each feature subset to $\hat{f}(t|\mathbf{X})$. In theory, $\hat{f}(t|\mathbf{X})$ can be decomposed into $\sum_{j=1}^p \binom{p}{j} = 2^p$ possible functions. In practice however, we are often only interested in the intercept term ($A = \emptyset$), main effects ($\left|A \right| = 1$) and two-way interactions ($\left|A \right| = 2$). 
Different methods exists for obtaining the decomposed functional terms. These include (generalized) functional ANOVA\cite{hooker2004,hooker2007}, accumulated local effects\cite{apley2020}, statistical regression models and $q$-interaction SHAP\cite{hiabu2023}. 

In \textbf{functional ANOVA} (F-ANOVA), each component is defined by
\begin{align}\label{eq:FANOVA}
    \hat{f}_A(t|\mathbf{X}) = \int_{X_{-A}} \left(\hat{f}(t|\mathbf{X}) - \sum_{B \subset A}\hat{f}_B(t|\mathbf{X})\right) \, dX_{-A} \; \text{.}
\end{align}
To compute a component $\hat{f}_A(t|\mathbf{X}_A)$, the components of  all subsets of $A$ ($B \subset A$) are subtracted from the expected value of the prediction function with regard to all features not in set $A$ ($X_{-A}$), assuming that all features follow a uniform distribution over their respective range of values. By subtracting the lower-order components, their corresponding effects are removed, if a higher-order effect is to be computed, as well as it ensures the estimated effect to be centered. In this way, the computation is iterative, since lower-order components are required for the computation of higher-order components.

Integrating over a uniform distribution when features are dependent, leads to new datapoints that deviate from the joint distribution and makes predictions for observations with improbable feature combinations. This issue of extrapolation, similar to those discussed in previous sections, can cause functional ANOVA results to be biased. To control for the effect of dependence between input features, \textbf{generalized functional ANOVA} (GF-ANOVA)\cite{hooker2007} was introduced. This method can also be easily adapted for survival analysis by employing appropriate prediction functions.

\textbf{Accumulated local effects} also constitute a functional decomposition method, since it holds that
\begin{align}\label{eq:ALE-decomposition}
    \hat{f}(t|\mathbf{x}) = \sum_{A \subseteq\{1,\dots,P\}} \hat{f}_{A,ALE}(t|\mathbf{x}_A) \; \text{.}
\end{align}
The components obtained from ALE decomposition satisfy the pseudo-orthogonality property, instead of hierarchical orthogonality as imposed by (generalized) functional ANOVA. Apley and Zhu\cite{apley2020} argue that pseudo-orthogonality is preferable to hierarchical orthogonality, since it does not mix marginal feature effects. Additionally, ALE has the advantage of allowing for a hierarchical estimation of components, avoiding the computational complexities associated with an estimation of the joint distribution.

The \textbf{statistical regression model} approach presents a bottom-up approach for model decomposition, by enforcing simplifying constraints in the modelling process. In its simplest form, regression models, such as the CoxPH model, or generalized additive models (GAMs) with varying degrees of interactions are fitted to the data. Since any likelihood-based regression model, such as the CoxPH model, can be transformed to a GAM, as introduced by Hastie et al.\cite{hastie1986}, this approach is readily transferable to survival analysis. Hastie and Tibshirani\cite{hastie1995} and Bender et al.\cite{bender2018} further provide detailed tutorials on the use of GAMs for time-to-event data. 

Hiabu et al.\cite{hiabu2023} propose a new identification, equivalent to \textbf{q-interaction SHAP}\cite{tsai2023}. Theoretically, their results can be applied to any model, including survival machine learning models and they provide an algorithm for calculation of exact q-interaction SHAP for tree-based models. Their identification requires $\hat{f}(t|\mathbf{X})$ to be decomposed as described in Equation~\ref{eq:ALE-decomposition} such that 
\begin{align}\label{eq:q-int-SHAP-decomp}
    \sum_{C \cap A \neq \emptyset} \int \hat{f}_C(t|\mathbf{X}_C)\hat{p}_A(\mathbf{X}_A) \, d\mathbf{X}_A = 0 \; \text{,}
\end{align}
with $\hat{p}_A$ some estimator of the density $p_A$ of $X_A$. Given any initial estimator $\hat{f}_A^{(0)} \ | \ A \subseteq {1,\dots,p}$, the unique set of functions $\hat{f}^{\star} = \{ \hat{f}^{\star}_A \ | \ A \subseteq {1,\dots,p} \}$ that satisfy Equation~\ref{eq:q-int-SHAP-decomp} are defined by
\begin{align}\label{eq:q-int-SHAP-components}
    \hat{f}^{\star}_A(t|\mathbf{X}_A) = \sum_{B \subseteq A} \left( -1\right)^{\mid A \setminus B\mid} \int \hat{f}^{\text{sum}}(t|\mathbf{X})\hat{p}_{-B}(\mathbf{X}_{-B}) \, d\mathbf{X}_{-B} \; \text{,}
\end{align}
where $\hat{f}^{\text{sum}}(t|\mathbf{X}) \coloneqq \sum_A \hat{f}_A^{(0)}(t[\mathbf{X}_A) = \sum_A \hat{f}_A^{\star}(t|\mathbf{X}_A)$. The major novelty of this decomposition is that it marries local and global explanations. This is because, on the one hand, (interventional) SHAP values can be extracted from the identification as weighted averages of the components. This allows for a decomposition of simple SHAP values into main effects and all involved interaction effects, extending the concept of functional decomposition to a local explainability scale. On the other hand, global partial dependence plots can be described by the main effect terms plus the intercept. This also holds for higher-order PDPs; in this case all feature sets that contain the feature of interest are included. 

The concept of functional decomposition is limited by the dimensionality restriction for higher-order interactions. More specific limitations depend on the specific estimator used. While functional ANOVA suffers from the extrapolation issue in the presence of dependent features, as explained above, in generalized functional ANOVA the components may mix marginal effects of correlated features. This depends on the choice of the weighting function due to the hierarchical orthogonality condition. Furthermore, its estimation process is rather computationally intensive, requiring the solution of a large, weighted linear system. Although possessing computational advantages and potentially more favorable theoretical characteristics in contrast to the (generalized) functional ANOVA as discussed earlier, accumulated local effects fail to offer a variance decomposition. Consequently, there is no assurance that the variance of the functional components aggregates to the total variance of the function, thereby diminishing its interpretability from a statistical perspective. Regression model based decomposition becomes difficult in the presence of feature interactions, since these have to be specified manually. Not properly identifying interactions or the presence of feature correlations may also lead to biased results for main effects due to a failure to comply with underlying assumptions, e.g., in case the CoxPH model is used for decomposition. This is closely linked to the concerns surrounding global surrogate models, discussed in Section~\ref{sec:LIME}.

\subsubsection{Model-Specific Global Methods}

The feature visualization approach\cite{olah2017} delves into identifying inputs closely aligned with the training dataset that elicit heightened activation in specific units within a neural network. Therein, ``unit'' pertains to individual neurons, channels (also referred to as feature maps), or entire layers of the neural network. The foundation of the feature visualization approach is an optimization problem, which can be solved using various optimization and regularization techniques. For example, training images or (combinations of) tabular features that maximize the activation can be selected. Alternatively, new images or feature (combinations) can be generated employing some regularizing constraints. The network dissection approach by Bau et al.\cite{bau2017} connects highly activated areas of convolutional neural net (CNN) channels with human concepts. It measures the CNN channel activations for training images annotated with human-labeled visual concepts and quantifies the activation-concept alignment with the intersection over union (IoU) score. It is built on the assumption that CNN channels learn disentangled features. The testing with concept activation vectors (TCAV) method\cite{kim2018} goes one step further by training a binary classifier separating the activations generated by a human-labeled concept dataset from those generated by the random dataset. The model's sensitivity to specific concepts is gauged through the directional derivative of the prediction in the direction of the coefficient vector of the trained binary classifier - the so-called concept activation vector (CAV)\cite{kim2018}. A statistical test for the overall conceptual sensitivity of an entire class can be defined using the ratio of inputs with positive conceptual sensitivities to the number of inputs for a class for multiple random datasets. To date, none of the mentioned approaches has been employed within the domain of survival neural networks. Zhong et al.\cite{zhong2022} approach global interpretability for survival neural networks by using a linear surrogate model for a  subset of features and then model the rest of the features by an arbitrarily complex neural network. However, it is crucial to note that this methodology does not inherently address the broader challenge of enhancing the interpretability of neural networks. A majority of methods for global neural network interpretability are either predominantly constrained to image data or insufficiently explored for tabular data. In the context of survival data, this is only relevant if images serve as inputs for the survival prediction. 

While particularly popular for neural networks, model-specific feature attribution methods likewise exist for tree-based machine learning models. Here, however, the contribution score of individual features is usually understood globally rather than locally, quantifying the effects on the overall predictions made by the ensemble model. One common technique for tree ensemble attribution is feature importance, as many of the popular feature importance measures, such as the permutation feature importance (PFI) or mean decrease of accuracy (MDA) and impurity importance or mean decrease of impurity (MDI) have originally been developed in the contexts of random forests\cite{breiman2001}. Later, they have been extended to work for a large class of models, which is why they are discussed in Section~\ref{feature_importance}. Another global tree interpretability technique based on the idea of surrogate trees is to simplify a complex ensemble of decision trees to a few or one representative tree. The core idea behind representative tree algorithms typically involves evaluating tree similarity using a distance metric.\cite{banerjee2012} Distance measures can encompass various aspects of similarity, e.g. predictions, clustering in terminal nodes, the selection of splitting features, or the level and frequency at which features are chosen in both individual trees and ensembles. The representative tree is typically the one demonstrating the highest mean similarity to all other trees. Alternatively, clustering algorithms can be applied to identify multiple representative trees from distinct clusters.\cite{laabs2023}

\subsection{Interpretable Models}\label{interpretable_models}

\subsubsection{Statistical Survival Models}
Intrinsically interpretable survival models allow to extract humanly interpretable insights regarding factors affecting survival probabilities or hazard rates of individual or groups of instances over time.
(Semi-)Parametric statistical survival models usually provide interpretable coefficients or parameters. In the parametric Weibull model, for example, the scale parameter represents the rate of event occurrence. Accelerated failure time models\cite{kalbfleisch2011}, such as the log-normal or the log-logistic model, specify directly how the survival time is affected by features. Decision tree-based models are also widely recognized for their high interpretability. This is due to their ability to divide data into segments based on feature values, resulting in discernible decision rules that make the resulting branches and nodes easily explainable. 

The most-frequently used interpretable survival model is certainly the semi-parametric Cox regression model\citep{cox1972} (see Section~\ref{survival_analysis}). Grounded in its historic and widespread use in clinical and epidemiological research, most survival researchers and practitioners would presumably argue in favor of the CoxPH model's intrinsic ease of interpretability. Indeed, due to the linearity assumption for $\varphi(\mathbf{x},\mathbf{b})$ (see Equation~\ref{eq:coxph3}), which implies the proportional hazard assumption, the signs and the magnitudes of the coefficients $\mathbf{b}^{\intercal} = (b_1,\dots,b_p)$ can be used to quantify the direction and strength of the relationship between feature and prediction. In general, statistically significant coefficient estimates with large absolute values indicate important features. Yet, information on relative feature importance for a given prediction can only be obtained by comparing the absolute values of the coefficients multiplied by the values of its features $|x^{i}_j \times b_j|$ or by standardizing coefficients, since the coefficients can be of different scales. A large part of the popularity of the Cox regression model is rooted in the hazard ratio interpretation. The hazard ratio of a feature $j$ can be computed as $\text{HR}_j = \exp(b_j)$, representing the relative change in the hazard rate associated with a one-unit change in feature $j$, holding all other features constant. The proportional hazard assumption implies constant hazard ratios over time, further simplifying the interpretation. It must be taken into account, however, that the hazard ratio for a distinct feature is a measure conditional on $T \geq t$. This complicates the interpretation even for randomized studies in which the proportional hazard assumption holds, as subjects are no longer comparable for larger $t$ values, if other factors beside the feature of interest influence the time to event. Moreover, in practice, the strong assumptions of the Cox regression model are often not met, particularly the assumption of proportional hazards or no time-varying effects of features.\cite{rulli2018} Further, the CoxPH model suffers from the noncollapsibility problem.\citep{martinussen2013} If an unobserved covariate exerts influence on the time-to-event outcome, the marginal model, conditioned solely on the observed covariates, deviates from a Cox regression model. Consequently, this deviation introduces bias in the estimates of crude hazard ratios when compared to adjusted hazard ratios. More recently, Martinussen\cite{martinussen2022} has further provided mathematical proof, that hazard ratios can not be interpreted causally unless they are unity (no causal effect) or the unrealistic and untestable assumption of potential outcomes under exposure vs. no exposure being independent of each other is satisfied. These factors show that the CoxPH's models presumed ease of interpretation is a fallacy of ease of applicability. Additional drawbacks of the Cox regression model include a decline in performance when handling datasets with high dimensions or correlated features. The CoxPH model further fails to capture complex interactions or non-linear feature effects if these are not manually specified. In summary, this highlights the need for developing and deploying a broader and universally applicable spectrum of models and explainability methods, suitably tailored to address the challenges posed by censored data. This pertains not solely to the realm of complex machine learning models but extends to encompass classical statistical approaches for the purpose of transcending and augmenting conventional modes of comprehending outputs, such as the hazard ratios.

\subsubsection{Interpretable Machine Learning Survival Models}
In general, all survival models that produce some form of output parameters that are linked in some manner to a human-intelligible representation of temporal progression within time-to-event data, can be categorized as interpretable. Discussing them all is beyond the scope of this paper. Given our primary emphasis on the interpretability of machine learning model outputs, we concentrate on particularly salient model categories in this context.

In the field of neural networks and deep learning in recent years several interpretable survival models have been developed, which we choose to highlight due to their aim of amalgamating the predictive capabilities of neural networks and deep learning models with outcomes that are amenable to interpretation. A group of deep learning approaches strives to achieve interpretability by adhering closely to the traditional Cox regression model. For instance, Xie \& Yu\cite{xie2021} suggest employing a neural network to fit a mixture cure rate model. This model maintains the proportional hazards structure for uncured subjects. With \textit{WideAndDeep}, Pölsterl et al.\cite{polsterl2020} combine the output of a neural network analyzing image data with the output of a Cox regression model handling tabular clinical data. Thus, the conventional Cox regression interpretability is retained for the tabular features. 

An alternative group of deep learning methodologies draws inspiration from the techniques used to induce explainability in other semi-parametric survival models. Within this context, an appealing category of models are additive survival models. Source of their inherent intrinsic interpretability are feature level contribution functions, quantifying the relationship between feature values and prediction functions at different timepoints. They can be readily extracted due to the additive nature of the models and aggregated to overall feature importance scores. These scores are measured as mean individual feature contributions on the survival probability or hazard at pre-specified timepoints. Central additive statistical models are generalized additive models\cite{hastie1986,hastie1995,bender2018}, which are discussed further in the context of functional decomposition (Section~\ref{sec:functional_decomposition}). Wu \& Witten\cite{wu2019} additionally introduce an additive CoxPH model, wherein each additive function is derived by means of trend filtering. Several machine learning models incorporate additive model structures as a means to augment their interpretability. \textit{SURVLSSVM}, an additive survival extension of least-squares support vector machines, was proposed by Van Belle et al.\cite{vanbelle2010}. It uses componentwise kernels, allowing to estimate and visualize feature effects in functional forms. Xu \& Guo\cite{xu2023} (\textit{CoxNAM}) and Rahman \& Purushotham\cite{rahman2021,rahman2022} (\textit{PseudoNAM} and \textit{Fair PseudoNAM}) have developed neural additive models (NAMs) for survival outcomes based on the idea of Agrawal et al.\cite{agarwal2021}. They combine some of the expressivity of deep neural networks, particularly their ability to capture non-linearity with the inherent interpretability of generalized additive models.  In similar vein, in \textit{DeepPAMM}, Kopper et al.\cite{kopper2022} embed piecewise exponential additive mixed models (PAMMs)\cite{bender2018} into a highly flexible neural network, which can accommodate multimodal data, time-varying effects and features, the inclusion of mixed effects to model repeated or correlated data, as well as different survival tasks. The predictor modelled by \textit{DeepPAMM} contains a structured additive part corresponding to a PAMM model representation and an unstructured part, capturing high-dimensional interactions between the features from the structured part using a customizable neural network. The structured part estimates inherently interpretable feature effects, providing classical statistical interpretability, e.g., through feature effect plots. Furthermore, many deep learning models explore dimensions of interpretability that extend beyond the ideas of existing semi-parametric methods. 

Chen\cite{chen2022} proposes \textit{survival kernet}, a deep kernel survival model, in which a kernel function is learned, which is used to predict the survival probability for unseen observations based on the most similar training observations. More specifically, kernel netting\cite{kpotufe2017} is used to induce interpretability, by partitioning the training data into clusters and then representing unseen instances as a weighted combinations of a few clusters. The clusters can be interpreted classically by comparing the Kaplan-Meier survival curves across clusters. Alternatively, a heatmap visualization of the fraction of points in a cluster with specific feature values can be employed to deduct feature-based characteristics for each cluster. Since kernel weights determine the contribution of each cluster to a new data instance, local interpretations of the relative importance of different clusters for the hazard function prediction can be extracted. 

In the domain of multi-omics data analysis, a collection of applications has emerged, wherein interpretability is achieved through the attribution of biological significance to nodes. In these methodologies, the pursuit of interpretability is a fundamental design objective. Consequentially, significant attention is directed towards assessing the interpretable outcomes in a comprehensive, sensitive and diverse manner. 
Hao et al.\cite{hao2019CoxPAS} developed a pathway-based sparse deep survival neural network, named \textit{Cox-PASNet}. Its architecture is key to the intrinsic interpretability.A pathway layer incorporates prior knowledge on pathways characterizing biological processes, with the values indicating the active/inactive status of a single pathway in a biological system. Hidden/integrative layers show the interactive (nonlinear and hierarchical) effects of multiple biological pathways. The observations are separated into risk groups by prognostic index outcome of the tuned network. For the risk groups, layer-wise feature importance scores are assigned comparing the average absolute partial derivatives, visualized by heatmaps. A log-rank test and the comparison of Kaplan-Meier survival curves allows the classification of top-ranked features as prognostic factors. Additionally, important pathway and integrative nodes are visualized by t-SNE, to illustrate hierarchical and nonlinear combinations of pathways. 
Hao et al.\cite{hao2019PAGE} extend these intrinsic means of interpretation to a biologically interpretable, integrative deep learning model that integrates histopathological images and genomic data (\textit{PAGE-Net}). For model interpretation, a focus is put on the pathology-specific layers and the genome-specific layers. For the genome-specific layers, pathway-based ranking based on partial derivatives is applied and analyzed analogously to \textit{Cox-PASNet} to identify genetic patterns that cause different survival rates between patients. From pathology-specific layers, pre-identified histopathological patterns associated to cancer survival are displayed by means of survival-discriminative feature maps. The feature maps reveal whether the model successfully identifies the morphological patterns that are of interest to pathologists. Furthermore they can uncover previously undetected relationships between any specific morphological patterns and survival prognosis. 
Like Hao et al.\cite{hao2019CoxPAS}, Zhao et al.\cite{zhao2021} aim to integrate multi-omics data into an interpretable survival deep learning framework with \textit{DeepOmix}. \textit{DeepOmix} is a feed-forward neural network.
Again, key to the intrinsic interpretability of the model is the incorporation of prior biological knowledge of gene functional module networks in a functional module layer. Such functional module networks may for instance be tissue networks, gene co-expression networks, or prior biological signaling pathways. While the hidden layers are able to capture the nonlinear and hierarchical effects of biological pathways associated with survival time, purposefully, low-dimensional representations are learned in the functional module layer, such that nodes represent groups of patients. Significant modules corresponding to prognostic prediction results are extracted by ranking them based on p-values obtained by a Kolmogorov-Smirnov test. It assesses whether the distribution of the patients in the gene modules is significantly different from previously defined high-risk/low-risk subgroups based on predicted survival time. 
In \textit{Cox-nnet}\cite{ching2018},
interpretability is derived by first selecting the most important nodes based on the highest variances in contribution. Contribution is calculated as the output value of each hidden node weighted by the corresponding coefficient of a Cox regression output layer. A t-SNE plot visualization reveals the hidden nodes' ability to reduce dimensionality of the initial omics data while effectively distinguishing samples based on their prognostic index. Then, to explore the biological relevance of the previously discerned important hidden nodes significantly enriched gene pathways are identified using Pearson’s correlation between the log transformed gene expression input and the output score of each node across all patients.
To obtain an importance score of each input gene for survival outcome prediction, the average partial derivative of the output of the model with respect to each input gene value across all patients is computed. 

For other methods, interpretability serves as an incidental consequence of an intrinsic design element within the novel neural network architecture proposed and is only briefly addressed in a peripheral manner.
\textit{SurvTRACE}\cite{wang2022}, a transformer-based deep learning model for competing risk survival analysis, uses multi-head self-attention in the encoder module allowing the model to automatically learn high-order feature interactions. Additionally, the learned attention scores can be utilized to understand feature relevance, interactions and importance. \textit{SurvNode}\cite{groha2020}, a deep learning framework for multi-state modelling, utilizes latent variable extensions to model uncertainty. The resulting latent space is divided into clusters, ideally providing a meaningful differentiation between survival outcomes and state transitions. Groha et al. choose to plot the state occupation probabilities, estimated using the non-parametric Aalen–Johansen estimator, for each cluster separately over time. An alternative method for acquiring feature effects linked to each cluster is introduced, fitting a logistic surrogate regression model within each cluster.

\section{Data Application Example}\label{sec:tutorial}

The primary motivation behind the methodologies discussed in this paper lies in the generation of explanations for different types of survival models. These explanations play a crucial role in interpreting and comprehending pivotal model decisions, ideally contributing to uncovering novel universal domain specific knowledge.  In order to demonstrate a potential use, we employ several of the discussed interpretable machine learning techniques on four models trained with real-world data from the well-known German Breast Cancer Study Group (GBSG2) dataset\cite{schumacher1994}. 
The focal objective is to predict cancer recurrence, leveraging a set of 7 demographic and clinical features and one treatment feature for hormonal therapy. A descriptive overview table detailing the features of interest can be found in Appendix~\ref{app1:data_example}) (see Table~\ref{tab:features}.

\subsection{Data Preprocessing and Exploratory Analysis}
The German Breast Cancer Study Group recruited 720 participating patients with primary node-positive breast cancer into a comprehensive cohort study between 1984 and 1989. The study investigated the effectiveness of three versus six cycles of chemotherapy, as well as the additional benefit of hormonal treatment with tamoxifen, using a 2x2 factorial design. After a median follow-up period of nearly five years, 312 patients had experienced at least one recurrence or had died. We obtain the analysis data from the \texttt{pec}\cite{mogensen2012} package in \texttt{R}, which records the recurrence-free survival time (in days) for 686 patients (with 299 events), all of whom had complete data for key standard factors such as age, tumor size, number of positive lymph nodes, progesterone and estrogen receptor status, menopausal status, and tumor grade.

\begin{figure}[htb!]
\centering
\begin{subfigure}{0.4\textwidth}
    \subcaption{}
    \includegraphics[width=1.0\textwidth]{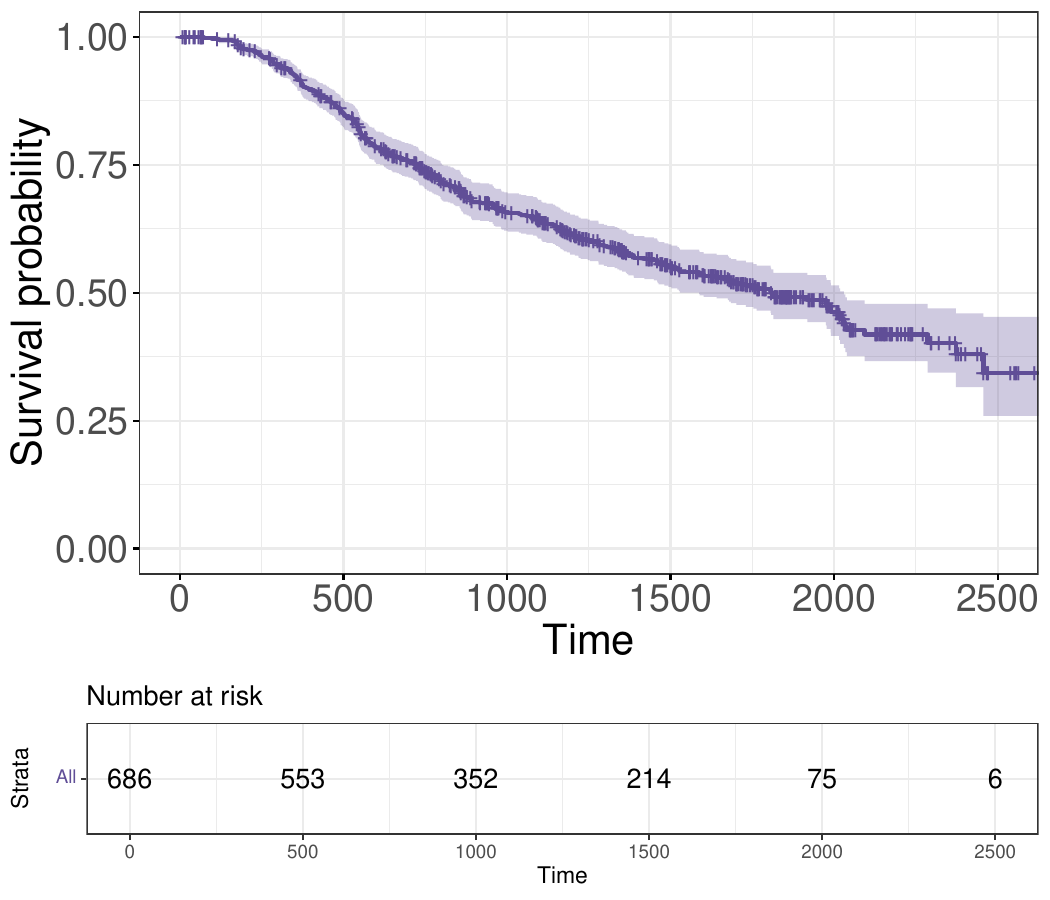}
    \label{fig:KM}
\end{subfigure}
\qquad
\begin{subfigure}{0.4\textwidth}
    \subcaption{}
    \includegraphics[width=1.0\textwidth]{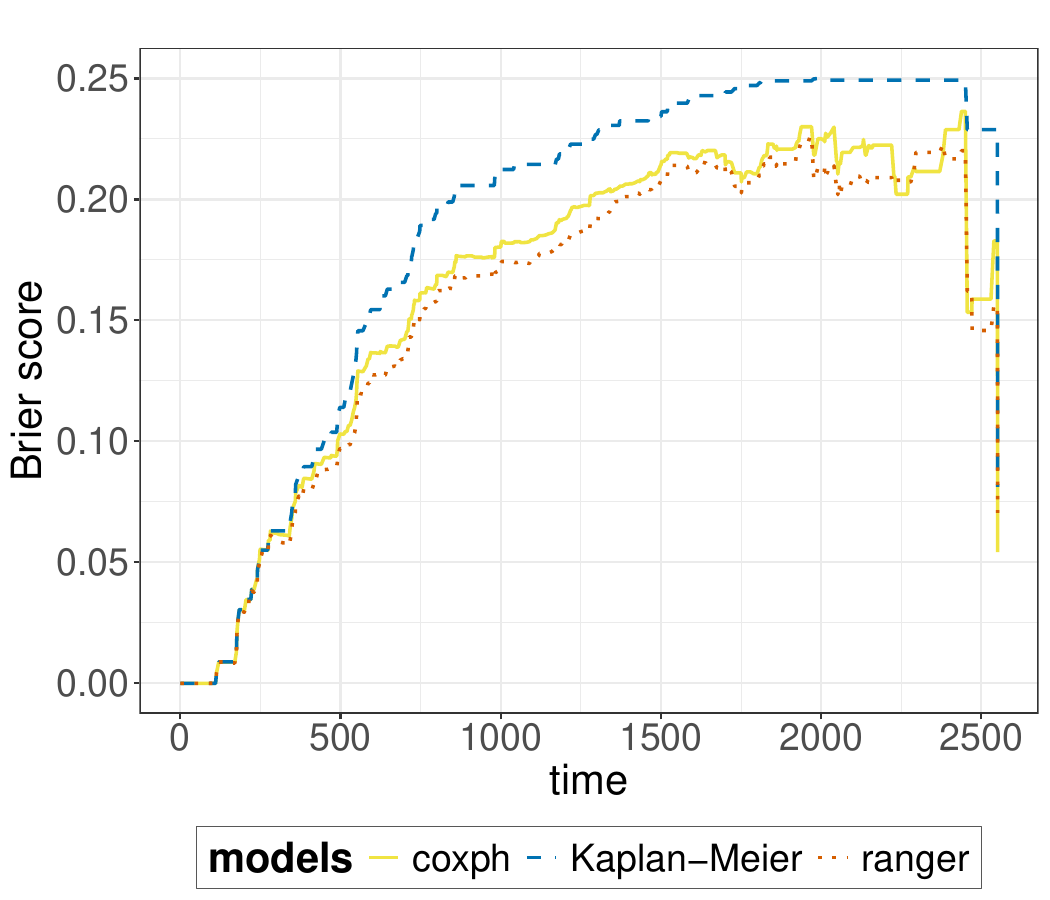}
    \label{fig:brier_real}
\end{subfigure}
\caption{Introductory analysis of GBSG2 dataset, (a) depicts the Kaplan-Meier survival curve and risk sets (b) Brier score performance plot.}
\label{fig:exploratory}
\end{figure}

\subsection{Model Fitting and Evaluation}\label{tut:models}

The data is split into training and test set and a CoxPH model is fit to the training data as a baseline and compared to a random survival forest model (\texttt{ranger} from \texttt{ranger} package\cite{wright2017ranger}). The performance is then evaluated and the explainability techniques are applied to the test data. Aggregated performance measures show that the \texttt{ranger} model slightly outperforms the \texttt{coxph} model in terms of integrated Brier score (0.16 vs. 0.156), C-index (0.716 vs. 0.74) and integrated C/D AUC (0.651 vs. 0.664). This is confirmed by the Brier score performance plot over time in Figure~\ref{fig:brier_real}. Overall, both models perform reasonably well on the data, outperforming the benchmark Kaplan-Meier estimator for any timepoints after 500 days. 

\subsection{Application of Interpretable Machine Learning Methods}

In the following, the interpretable machine learning methods discussed in the methodological section of the paper are applied to the two models estimated on GBSG2 data (see Section~\ref{tut:models}). The purpose of this section is to provide exemplary illustrations of the methods as a basic guideline for researchers aiming to gather insights into (machine learning) survival models. Most explanations were generated using the \texttt{survex}\cite{spytek2023} package in \texttt{R}, the \href{https://github.com/sophhan/IMLSA_2024}{code to reproduce} the results can be found on Github: \url{https://github.com/sophhan/IMLSA_2024}.

\subsubsection{Permutation Feature Importance}\label{sec:tut_pfi}

\begin{figure}[htb!]
  \centering
  \includegraphics[width = 0.9\linewidth]{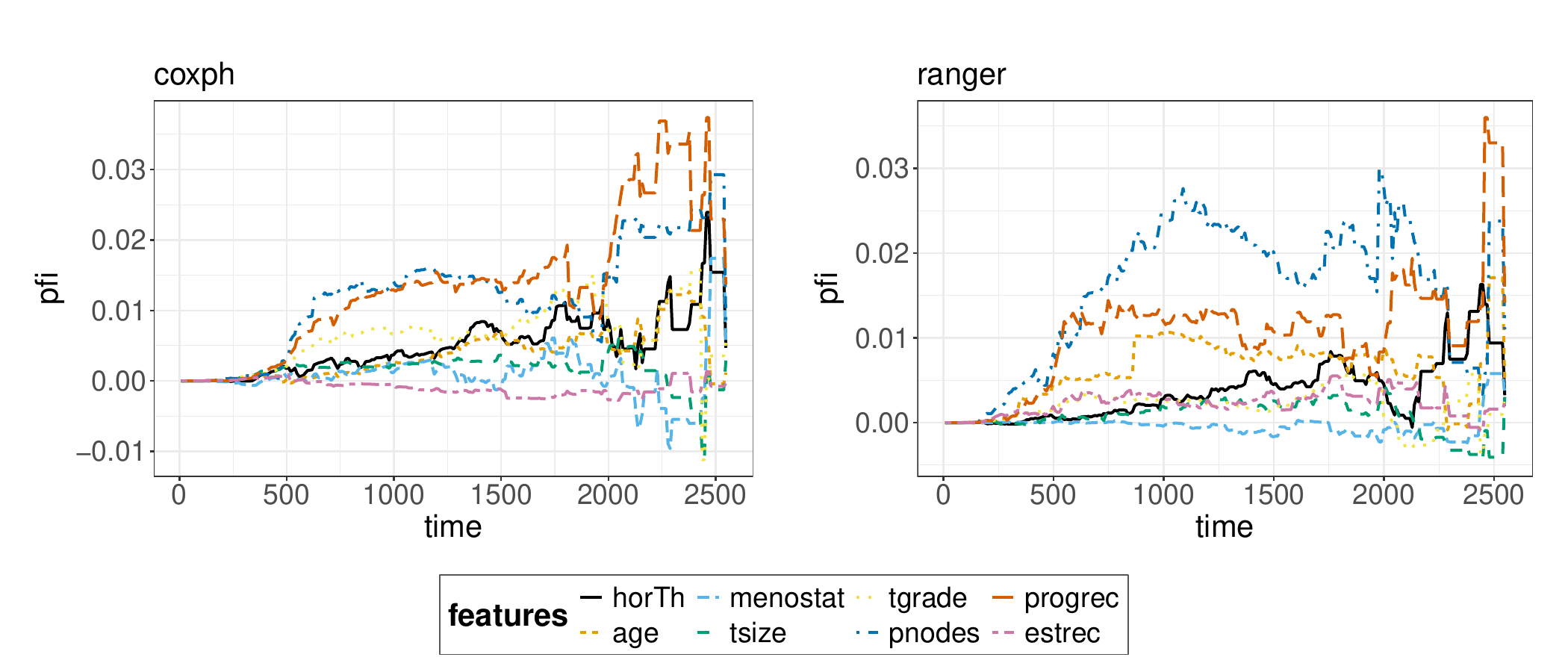}
  \caption{Permutation feature importance for the \texttt{coxph} model (left) and the \texttt{ranger} model (right). The y-axis denotes the Brier score loss after permutations of each separate feature with the loss of the full model including all features without permutations subtracted. Both models rely mostly on the \texttt{progrec}, \texttt{pnodes}, \texttt{tgrade}, \texttt{age} and \texttt{horTh} feature in particular for predicting survival times between 500 and 2200 days. Predictive power is distributed fairly evenly over  features.}
  \label{fig:pfi}
\end{figure}

Permutation feature importance is used to provide highly compressed, global insights into a models’ behavior. The plots in Figure~\ref{fig:pfi} show the increase in Brier score loss at every timepoint $t=0,\dots,2600$ when replacing a feature with its permuted version for the \texttt{coxph} and the \texttt{ranger} model, respectively. For both models, \texttt{progrec} and \texttt{pnodes} are the two most important features. While for later timepoints ($t>2000$ for the \texttt{coxph} model and $t>2200$ for the \texttt{ranger} model) the performance of both models relies heaviest on the \texttt{progrec} feature, for $t>500$ the \texttt{ranger} model depends the strongest on the the \texttt{pnodes} feature, with a visible gap in error increase between \texttt{progrec} and \texttt{pnodes}. For the \texttt{coxph} model we observe roughly three different clusters of features with similar importance levels for $t \in [500, \ldots,2200]$, with \texttt{progrec} and \texttt{pnodes} causing a similar decrease in the Brier scores, followed in magnitude by \texttt{tgrade}, \texttt{horTh} and \texttt{age} and \texttt{estrec}, \texttt{tsize} and \texttt{menostat} having little to no influence on the performance, with \texttt{tsize} and \texttt{menostat} even lowering the Brier score, when replaced by random permutations. For the \texttt{ranger} model \texttt{age} and \texttt{horTh} are more clearly identifiable as the third and fourth most important features for $t \in [500, \ldots,2200]$, while \texttt{estrec}, \texttt{tsize} and \texttt{menostat} are less relevant for improving the prediction performance. Due to the few patients at risk and events after 2000 days, the feature importance values become more confusing and should only be interpreted with care. The PFI values being very close to zero for $t<500$ for both models is in line with the the Brier score plot over time in Figure~\ref{fig:brier_real}, indicating that the featureless model performs equally well for those early timepoints.
\subsubsection{Individual Conditional Expectation and Partial Dependence Plots}
\begin{figure}[htb!]
  \centering
  \includegraphics[width = 0.9\linewidth]{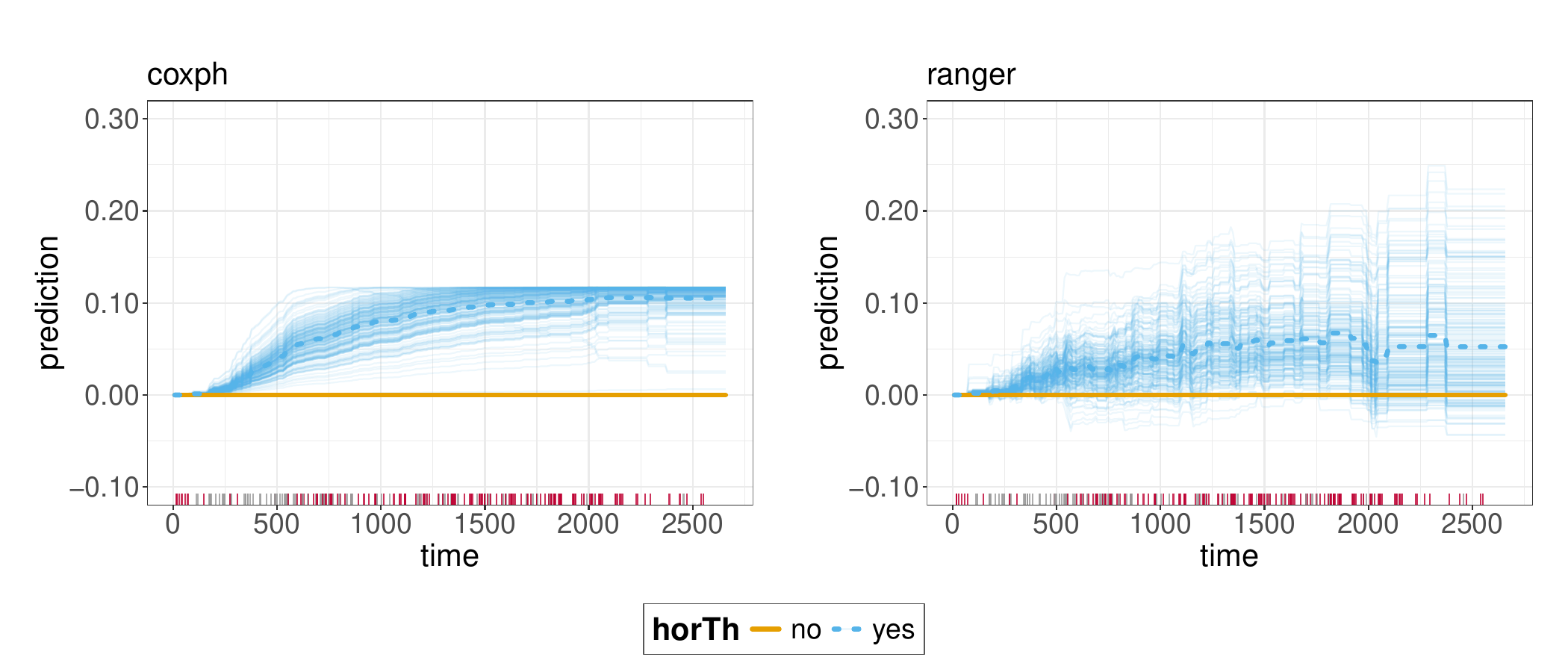}
  \caption{c-ICE and c-PDP curves for the \texttt{coxph} model (left) and the \texttt{ranger} model (right) for the \texttt{horTh} feature. c-ICE curves are depicted as thin colored lines, while c-PDPs are thick colored lines with white dashed lines in the center. Different colors indicate different treatment arms. The rug on the x-axis shows the survival time distributions (grey indicates recurrence, red censoring). The c-ICE (c-PD) curves depict the (average) predicted probability of survival for changes in the values of the feature of interest over time, with the time values denoted on the x-axis. In both models, \texttt{horTh} = 1 is in general associated with a higher predicted probability of recurrence free survival, with the effect amplifying over time.}
  \label{fig:pdp}
\end{figure}

Now centered individual conditional expectation (c-ICE) and partial dependence plots (c-PDP) are jointly plotted to visualize the influence of the \texttt{horTh} feature on the probability of breast cancer recurrence in the test set (see Figure~\ref{fig:pdp}). To compute the c-ICE curves, a random sample of 100 patients is taken from the test set, to ensure discernability of the depicted c-ICE curves by avoiding over-cluttering. The c-PDP curves are computed on the full test set. Centering with regard to the minimum observed feature value is employed to facilitate the assessment of the degree of heterogeneity in the ICE curve shapes to determine whether there are interactions between \texttt{horTh} and other features. For both the \texttt{coxph} and \texttt{ranger} models, the degree of heterogeneity in the c-ICE curves is low to moderate comparing the curves of different individuals across the same model. This suggests a low to moderate degree of interaction between the \texttt{horTh} feature and other features in both presented models. Considering the c-PDPs, the average predicted probability of recurrence free survival is found to increase in both models if patients receive hormonal therapy compared to not receiving therapy. In both models, this effect becomes with increasing recurrence free survival time. The estimated effect magnitude is similar in both models with $[0, 0.1]$.

\subsubsection{H-statistics}\label{sec:h_stat_real}

\begin{figure}
  \centering
  \includegraphics[width = 0.9\linewidth]{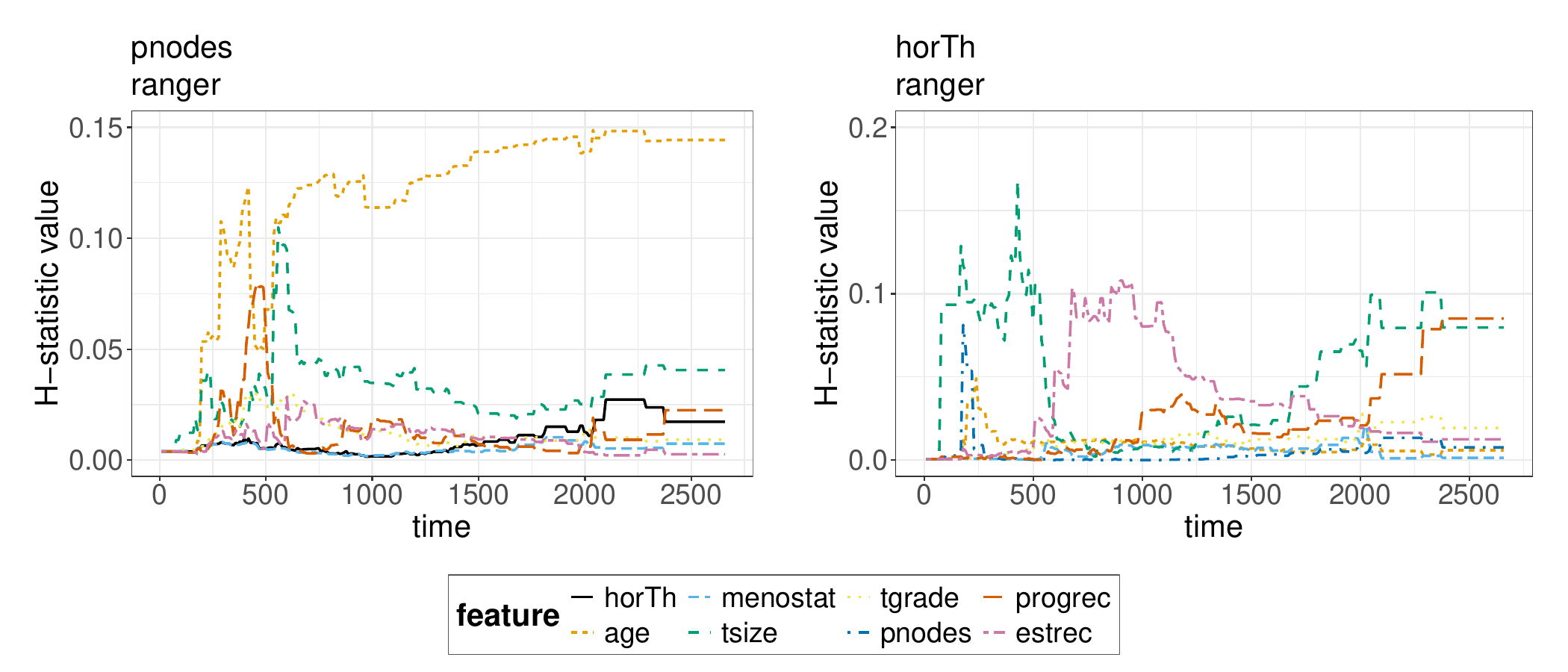}
  \caption{The 2-way interaction strengths ($\text{H}_{jk}(t)$-statistics) curves for the \texttt{ranger} model for the \texttt{pnodes} (left) and the \texttt{horTh} (right) feature. The interaction at each timepoint is the proportion of variance explained by the interaction. The both features are found to interact with other features on a moderate to low level. A consistent but moderate in overall degree interaction is suggested between \texttt{pnodes} and \texttt{age}.}
  \label{fig:h-stat}
\end{figure}

Complementary to the previous results, we calculate the two-way interaction statistics for the \texttt{horTh} and the \texttt{pnodes} feature in the \texttt{ranger} model, respectively (see Figure~\ref{fig:h-stat}). As deducted from the low to moderate degree of heterogeneity visible in the c-ICE curves generated from the \texttt{ranger} model for the \texttt{horTh} feature in Figure~\ref{fig:pdp}, the values of the $\text{H}_{jk}(t)$-statistics are overall low. The strongest interactions for earlier timepoints are observed for \texttt{tsize} (for $t<500$) and for \texttt{estrec} (for $t \in [500, 1200]$); for later time points some level of interaction between \texttt{horTh} and \texttt{tsize} and \texttt{progrec} is suggested. While the values of the $\text{H}_{jk}(t)$-statistics are likewise low overall for the \texttt{pnodes} feature, a consistent interaction between \texttt{pnodes} and \texttt{age} is suggested. Generally speaking, the H-statistic values can always only be an indication of interaction strength; it is difficult to say when it is large enough to consider an interaction strong, relevant interactions need to be assessed using alternative methods to determine type and direction of interaction. 

\subsubsection{Accumulated Local Effects}

\begin{figure}[!htb]
  \centering
  \includegraphics[width = 0.9\linewidth]{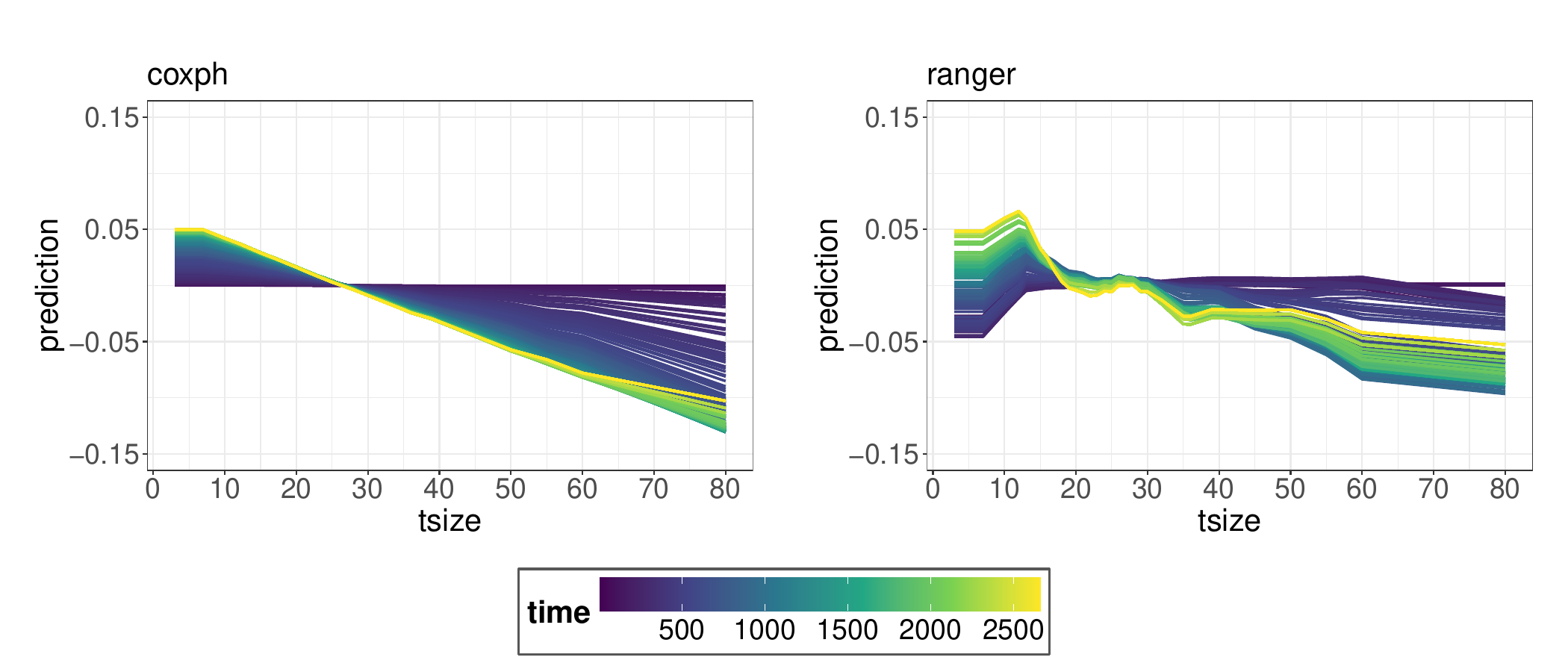}
  \caption{Accumulated local effects curves for the \texttt{tsize} feature in the \texttt{coxph} model (left) and the \texttt{ranger} model (right). The accumulated local effects are plotted as unbiased estimates of the main effects of the different values of the continuous \texttt{tsize} feature, which are denoted on the x-axis. Different timepoints are indicated by color gradients. Both models predict a decrease in average probability of recurrence free survival for women larger tumors.}
  \label{fig:ale}
\end{figure}

Based on the observations from Section~\ref{sec:h_stat_real} interactions with other features cannot be ruled out in the machine learning models for the \texttt{tsize} feature, it is interesting to consider accumulated local effects (ALE) plots instead of PDPs, as they can be biased. The uncentered ALE main effect curves for the \texttt{tsize} feature computed on the full test set for the \texttt{coxph} and \texttt{ranger} model are shown in Figure~\ref{fig:ale}. Their interpretation is analogous to uncentered PDPs, except that the effect of different feature values on the average predicted survival probability at different survival timepoints is computed conditional on a given value. Both models predict the average probability of recurrence free survival of patients to decrease with larger tumor size. This effect becomes stronger for later timepoints, which simply reflects the natural shape of the survival curve, also observed in the Kaplan-Meier plot in Figure~\ref{fig:KM}. The results of the different models only differ in the shapes of the ALE curves, which reflect the assumptions of the underlying models. Parallel, straight-line ALE curves are observed for the \texttt{coxph} model due to the proportional hazard assumption, while the \texttt{ranger} allows for more flexibility in the feature effects.  

\subsubsection{SurvLIME}

\begin{figure}[htb!]
  \centering
  \includegraphics[width = 0.9\linewidth]{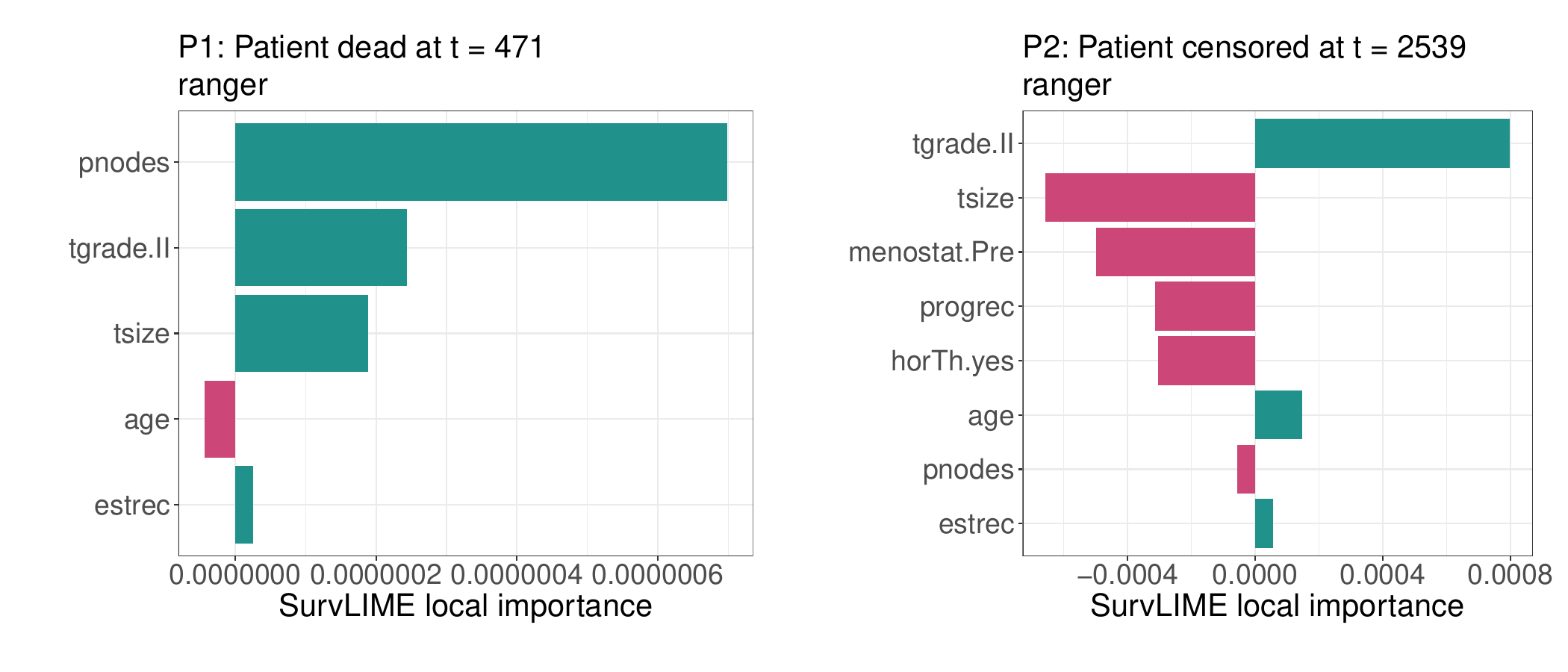}
  \caption{The local feature importance values from \emph{SurvLIME} for two selected observations computed on the \texttt{ranger} model ranked by absolute magnitude. The local importance values of each feature are obtained as the product of the absolute values of the coefficients estimated by the respective surrogate CoxPH model and the feature values observed for the specific observation. Categorical features are one-hot encoded, with the positive value chained after the feature name.}
  \label{fig:lime}
\end{figure}

After exploring the global effects of features and their interactions on the predicted recurrence-free survival of the breast cancer patients, we will now demonstrate how different local methods can be utilized to explore the effects of certain features on one specific instance. For this purpose two exemplary patients are chosen from the test set, whose characteristics are denoted in Table~\ref{tab: lime_children}. For these patients \emph{SurvLIME} is applied to the black box survival curves predicted by the \texttt{ranger} model. These curves are shown in Figure~\ref{fig:surv_grid} alongside the survival curve predictions of the Cox proportional hazard surrogate model in the neighborhood\footnote{By default the neighborhood size is set to $g=100$.} of the respective observation of interest. Figure~\ref{fig:lime} shows the local importance values estimated based on the surrogate model, ordered based on absolute effect size, with zero coefficients excluded. For both patients the locally important features deviate significantly from the globally important features permutation feature importance (see Figure~\ref{fig:pfi}) or global \emph{SurvSHAP(t)} importance (see Figure~\ref{fig:shap_agg}). The local importance values can for instance be interpreted in the following way: The tumor size positively influences the predicted probability of survival of the surrogate model for patient P1, while it negatively affects the predicted probability of survival for patient P2; the opposite holds for the \texttt{age} feature in both cases. It must be brought to the fore, that the \emph{SurvLIME} explanation survival function diverges strongly from the black box survival prediction for P1, indicating a rather poor fit of the surrogate model (compare in Figure~\ref{fig:surv_grid}); while the divergence is less strong for the curves of P2, the fit is still subpar. Additionally, the estimated local importance values are very close to zero impeding a meaningful interpretation. In conclusion, the lack of a theoretical foundation, as well as the many practical problems in \emph{SurvLIME} estimation, such as the correct definition of the sampling neighborhood, the overall instability of explanations and the inability to capture time-dependent effects arguably hinders practical usability of \texttt{SurvLIME} explanations.

\subsubsection{SurvSHAP(t)}

\begin{figure}[htb!]
  \centering
  \includegraphics[width = 0.9\linewidth]{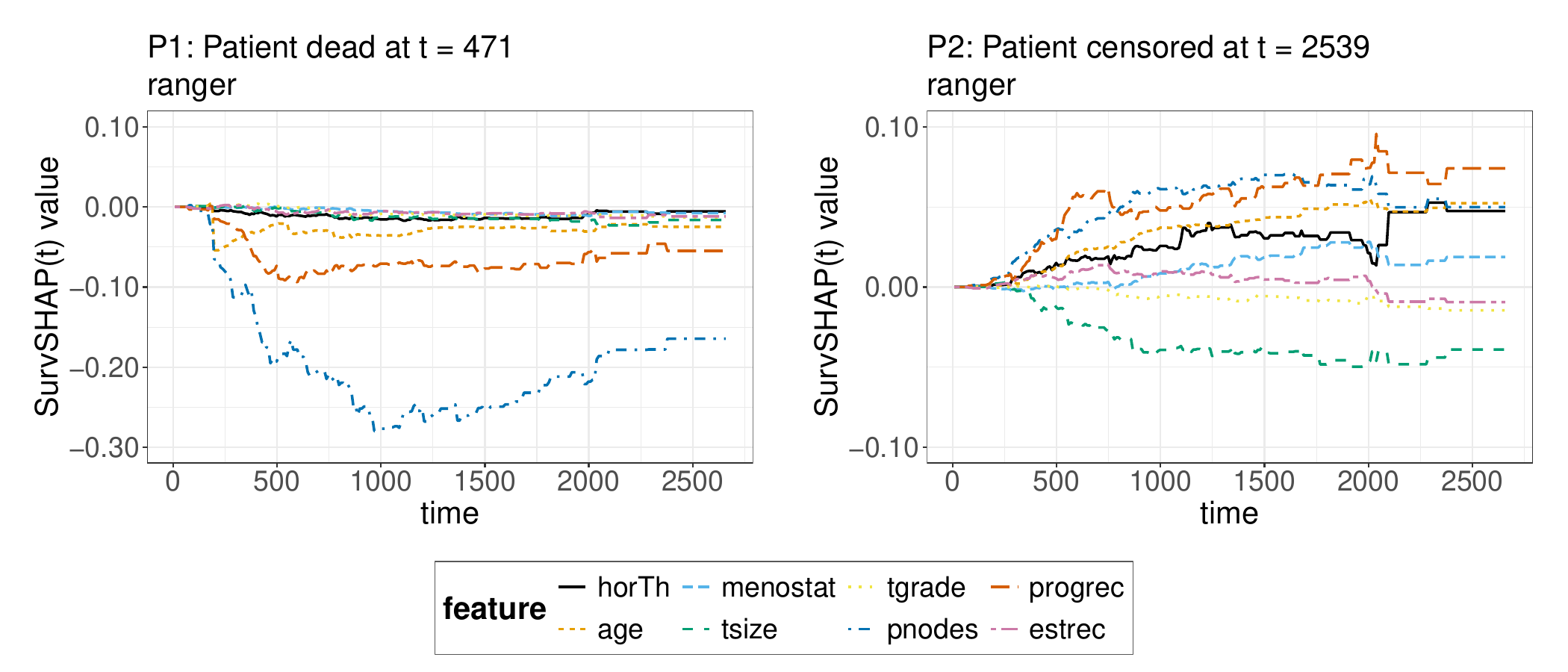}
  \caption{\emph{SurvSHAP(t)} for two selected observations computed on the \texttt{ranger} model. The \emph{SurvSHAP(t)} values plotted at different timepoints $t$ quantify the contribution of each feature to the difference between the black box survival function predicted for the particular observation and the average predicted survival function of the black box; hence all curves summed up over the different timepoints added to the average predicted survival curve of the model, will output the predicted survival curve of the chosen observation.}
  \label{fig:shap}
\end{figure}

In contrast to \emph{SurvLIME}, \emph{SurvSHAP(t)} offers insight into how each feature influences the black box survival function at different timepoints. For comparison purposes the same patients as described in Table~\ref{tab: lime_children} are considered. A general juxtaposition of locally important features identified by \emph{SurvLIME} (Figure~\ref{fig:lime}) vs. \emph{SurvSHAP(t)} (Figure~\ref{fig:shap}) shows that they do not generally align, with the exception of the \texttt{pnodes} and \texttt{tsize} feature for both patients. Furthermore, \emph{SurvSHAP(t)} further highlights the inandequacy of \emph{SurvLIME} as an explanation method, as the plots in Figure~\ref{fig:shap} show that the relative importance of different features is variable over time, even changing in relative order in some instances. In general, the \emph{SurvSHAP(t)} values at different timepoints quantify the contribution of each feature to the difference between the black box survival function for the particular patient and the average predicted survival function of the black box. For the patient dead at $t=471$ (P1), the patient having 30 positive lymph nodes, a progesterone receptor level of zero and being 80 years old, negatively contribute to the difference between the average predicted survival curve and the black box survival curve estimated by the \texttt{ranger} model shown in Figure~\ref{fig:surv_grid}. In contrast, for the patient censored at $t=2539$ (P2), the patient having only two positive lymph nodes, a higher progesterone receptor level of 264, being 47 years old and having received hormonal therapy the four features most important for explaining why the predicted survival curve for this patient exceeds the average predicted survival curve of the black box model. 

\begin{figure}[htb!]
  \centering
  \includegraphics[width = 0.9\linewidth]{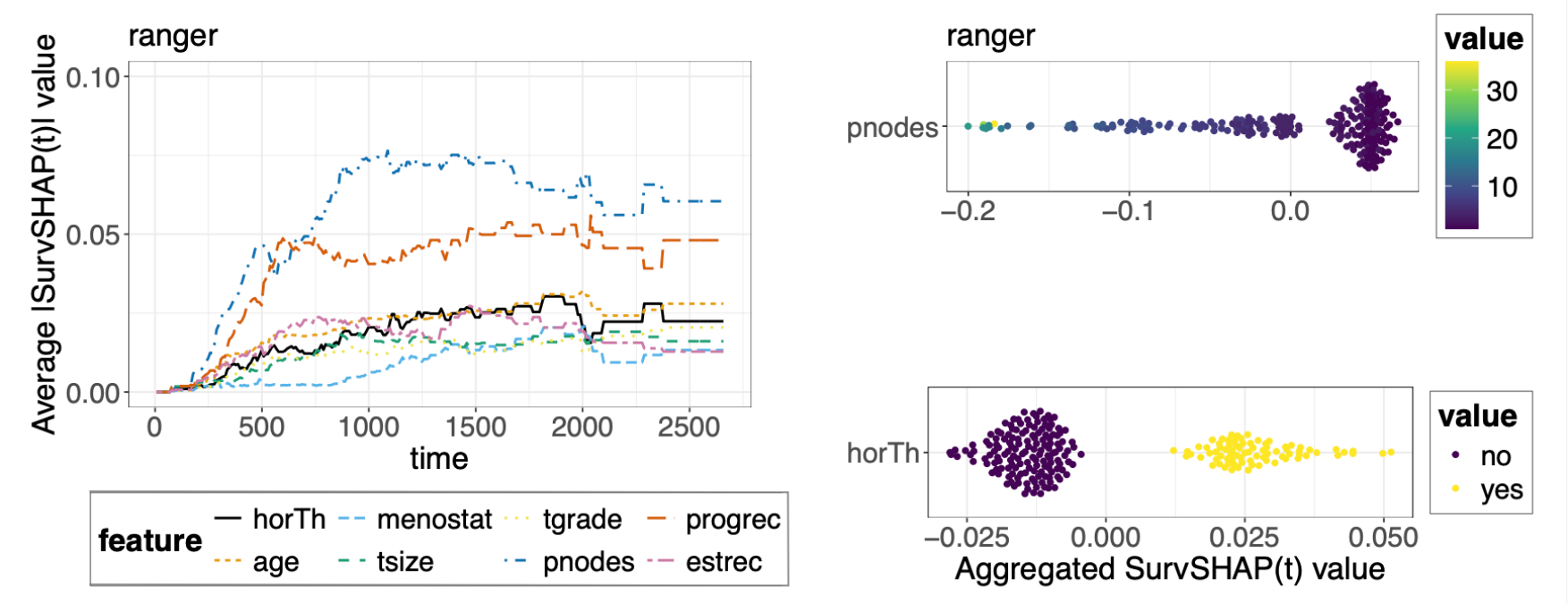}
  \caption{Aggregated \emph{SurvSHAP(t)} results computed on the \texttt{ranger} model. The left plot shows the average absolute \emph{SurvSHAP(t)} values computed over all observations in the test set plotted at different timepoints to quantify the global importance of each feature over time. The two right-hand side figures are beeswarm plots of the \emph{SurvSHAP(t)} values of every observation aggregated over time for a selected subset of features to summarize the global effect of individual features on the predicted survival probability.}
  \label{fig:shap_agg}
\end{figure}

\emph{SurvSHAP(t)} values can also help to reveal the impact of each feature across the entire dataset, highlighting which features consistently contribute more or less to the predictions. On the left hand side Figure~\ref{fig:shap_agg} shows the average absolute \emph{SurvSHAP(t)} contributions over time for the \texttt{ranger} model; \texttt{pnodes} is the most important feature at every timepoint except $t \in [500, 600]$, followed by \texttt{progrec}. All other features, except for \texttt{menostat}, are roughly of equal importance. The average absolute \emph{SurvSHAP(t)} contributions are increasing up to time $t=1000$ and then approximately constant. In line with the results of PFI, features are of low importance for the prediction for $t<500$, consistent with the Brier score of the featureless Kaplan-Meier learner compared to \texttt{coxph} and \texttt{ranger} model in Figure~\ref{fig:KM}. In order to understand both importance and direction of feature contributions aggregated over time for all observations in the test set beeswarm plots, as depicted on the right-hand side of Figure~\ref{fig:shap_agg}, can be used as information-dense summaries. For less important features, such as \texttt{horTh} the dots representing the aggregates \emph{SurvSHAP(t)} values are closely scattered around zero ($\in [-0.025, 0.05]$), whereas for more influential features like \texttt{pnodes}, the dots are spread out along the x-axis ($\in [-0.2, 0.1]$). Color is used to display the original value of a feature, to reveal their association with the aggregated \emph{SurvSHAP(t)} values. For instance, observations with low values of \texttt{pnodes} are in associated with positive aggregated \emph{SurvSHAP(t)} contributions, meaning they make a positive contribution to the predicted probability of survival of each instance.

\section{Limitations and Challenges}\label{sec:limits}

Many of the IML techniques discussed share limitations and challenges that must be addressed to enhance the effectiveness and applicability of survival model explanations. While this section focuses on model-agnostic methods, the absence of model-specific methods can be considered a distinct challenge requiring a solution. Identified issues include the general pitfalls associated with model-agnostic XAI methods detailed by Molnar et al.\cite{molnar2020general} and Molnar\cite{molnar2022}. Recurring and particularly noteworthy are the inherent computational complexity of many methods (e.g., KernelSHAP, H-statistics, generalized functional ANOVA or q-interaction SHAP), the inability to properly capture higher-order feature interdependencies (e.g., ICE, PDP or ALE) and the extrapolation issue (e.g., ICE, PDP, H-statistics or SHAP). The extrapolation issue arises as a consequence of the innate assumption of feature independence underlying many IML methods. These methods commonly involve the process of averaging over the marginal distribution of the feature set $-A$, which in the presence of correlated features, may take unrealistic data points, outside the margins of the observed values into consideration. Such extrapolation, in turn, leads to unreliable predictions. This concern is not limited solely to regions beyond the boundaries of observed values; it also pertains to areas with a significant separation between neighboring values of the feature of interest $A$, potentially resulting in an inappropriate and volatile relationship with the target variable, particularly in the presence of outliers.\cite{goldstein2015} In summary, this phenomenon leads to biased feature effect estimates in case of ICE or PDP, biased feature attributions in case of SHAP, even when using interventional SHAP values,\cite{aas2021,olsen2022,olsen2023} as well as unreliable H-statistics interaction values, since they are calculated based on partial dependence functions. Beside the widely acknowledged general pitfalls, the main emphasis of this section pertains to showcasing overarching limitations and challenges of the methods discussed, peculiar to the survival nature of the explanations, this constitutes: 

\textbf{Complex explanations.} When explaining survival models that produce functions as outputs, raw explanations can become increasingly complex. Traditional means of making survival models with functional outcomes more interpretable involves collapsing results into single numbers, such as hazard ratios, restricted mean survival time or median survival time. This is arguably an attempt of making their complexity comprehensible for users, at the apparent cost of a loss of important information. For real transparency and responsibility, a way must be found in which complex results are not simply reduced, but can be presented such that they deliver insights not erasing their nuances, while remaining comprehensible for end users and practitioners, e.g., in the medical domain. A proposed solution involves aggregating atomic explanations, which, in turn, complicates the explanation process. Such an approach might require integrating over time, introducing the need for selecting a suitable weight function. This choice of weight function can significantly impact the final results and potentially oversimplify or even hide the true findings. Alternatively, one might opt to retain explanations in the time domain, a concept named time-dependent explainability, requiring at least one additional dimension, compared to their counterparts in a classic, non-functional output setting. However, these explanations can be overly complex for end-users to grasp and come with their own set of issues discussed below. It is worth to note that this challenge of complex explanations is not limited to survival models and is a broader concern affecting any model producing outputs beyond simple scalar values.

\textbf{Complex visualizations.} Consequential to complex explanations are challenging visualizations. Visualization and interpretability are inseparably intertwined, as it is an indispensable tool to deliver human-understandable results. Balancing a comprehensive presentation of time-dependent, high-dimensional  explanations with transparency and cognitive intelligibility for humans, is a central challenge for future research in the field of interpretability for survival analysis or interpretability of functional output problems in general. Ideally, the challenges of complex explanations and complex visualizations should be jointly addressed by developing and deploying a broader and universally applicable spectrum of explainability methods for survival data, as well as more acute and insightful visualizations of existing methods and results.

\textbf{Computational complexity.} In addition to the inherent computational complexity of many IML methods outside of the survival context, time-dependent methods, integral for understanding survival predictions, often exhibit added computational complexity, given that explanations are often computed independently for each timepoint. For instance, in the standard machine learning setting PDPs and ICE plots for one feature (if computed for all $n$ observations) have a computational complexity of $\mathcal{O}(n \cdot g)$, as for each of the $g$ grid points, predictions are produced for $n$ observations. In the survival setting complexity is increased to $\mathcal{O}(n \cdot g \cdot t^{\max})$, since predictions are additionally required for each timepoint. ALE plots reduce the computational complexity to $\mathcal{O}(n)$ in the standard setting and $\mathcal{O}(n \cdot t^{\max})$, as for each observation (and timepoint in the survival setting) only one prediction is made for the original observation, the upper and the lower interval limit. 

\textbf{Noisy results.} Another challenge related to time-dependent methods is the inherent noise in their results, particularly when applied to flexible models like random survival forests. The \emph{SurvSHAP(t)} method, for instance, often outputs attribution functions with considerable variability, which may make it challenging to derive meaningful insights from the explanations. Introducing constraints or restrictions to achieve smoother outcomes might enhance cognitive readability. Additionally, these constraints could accelerate the calculation process.

\textbf{Timepoints selection.} The process of selecting timepoints for discretization is critical not only in time-dependent methods but also in approaches that rely on aggregations over time intervals. The number and values of these timepoints can significantly impact the explanation results, especially for methods that aggregate over time intervals. For example, selecting timepoints at the tail of the distribution, where the support may be limited, can introduce noise and irrelevant information to the results. Unfortunately, a lack of reporting of the selection criteria for these timepoints compounds the issue. Therefore, developing robust and transparent guidelines for selecting timepoints is crucial for further development.

\textbf{Applicability to different survival scenarios.} All of the discussed explanation methods are predominantly tailored to the prevalent scenario of right-censoring and time-constant features, reflecting the popularity of these conditions within survival analysis. However, such a focus potentially constrains their applicability in more diverse and complex scenarios, which may involve competing risks or time-varying features, which are also supported by many machine learning survival models. This limitation underscores the need for methodological developments to address the broader spectrum of conditions encountered in real-world survival analysis applications.

\textbf{Applicability to different data modalities.} While most of the discussed explainability methods are mainly designed to accommodate tabular data, the scope of survival analysis extends beyond this traditional data modality. For instance, in medical research, the analysis of medical imaging data, such as CT scans and X-rays, holds great promise for survival analysis.\cite{ranschaert2019,baniecki2023hospital} Developing approaches that can effectively explain models working with such data types is essential for advancing the utility of IML in survival analysis. 

\textbf{Relation to data-generating process.} 
Scientific inference is the process of rationally deriving conclusions about real world phenomena from data.\cite{freiesleben2022} This can build the basis of scientific explanations or knowledge about said phenomena, which are often of primary interest for scientists and practitioners. However, there is a critical difference between simply understanding how the model makes predictions and understanding the underlying data generating process (DGP). In contrast to statistical models, machine learning approaches often lack a mapping between model parameters and properties of the DGP.\cite{molnar2023} Recent work has shown how some model-agnostic IML methods can be used as statistical estimators of ground truth properties in the DGP in classification and regression contexts.\cite{freiesleben2022,molnar2023,ewald2024} Establishing equally clear connections with the DGP for explainability methods in the survival context, is a pivotal aspect of ensuring transparency and reliability of survival machine learning models. 

\section{Conclusion}

This comprehensive review and tutorial paper aims to serve as a fundamental reference for researchers delving into survival interpretable machine learning. Emphasis is put on model-agnostic methods for their applicability in a broad modelling context and their consequential ease of adaption into accessible software. We review existing work on model-agnostic interpretable machine learning methods specifically developed for survival analysis, including survival counterfactuals, \emph{SurvLIME} and its extensions, \emph{SurvSHAP} and \emph{SurvSHAP(t)}. Immediate research gaps identified are addressed, as other popular model-agnostic methods, which prior to this paper have not yet been modified to work in the survival context, are formally adapted to survival analysis. This concerns methods, whose scope of adjustment do not warrant uniquely dedicated pieces of work, such as ICE, PDP, ALE, H-statistic or different feature importance measures. By unifying the formalization of existing and newly adapted survival explainability methods, a concise referential basis is created, on which new work can be built upon. Furthermore, it facilitates highlighting limitations known from the non-functional output setting, as well as additional deficiencies resulting from the unique properties of the survival setting. In the process, opportunities for future research are identified and outlined. In addition to model-agnostic methods, the paper explores existing model-specific interpretability methods, particularly in the domains of survival tree ensemble learners and survival neural networks, overall offering a thorough overview of the current state of research in the field of interpretable machine learning for survival analysis.

Moreover, the paper includes an application of some of the reviewed and newly-adapted interpretability methods to real data, for the purpose of gaining insights into a set of statistical and machine learning models predicting cancer recurrence for patients in the German Breast Cancer Study Group. This part is meant to serve as a tutorial or guide for researchers, on how the methods can be utilized in practice to facilitate understanding of why certain decisions or predictions are made by the models. The computational results are obtained using \texttt{survex}\cite{spytek2023}, a user-friendly, state-of-the-art \texttt{R} software package, which allows to produce explanations for survival models from numerous of the most popular survival analysis related packages in \texttt{R}.\cite{spytek2023} Since the analysis is heavily simplified for the sake of clarity and the models are explicitly framed as prediction tools, conclusions about the underlying data must be drawn with utmost care. Particularly, it is crucial to highlight, that statements about causal relationships can in general not be made from explanation results, since the underlying models, like most statistical learning procedures, merely analyze correlation structures between features instead of the true inherent causal structure of the data generating process.\cite{molnar2020conf} In summary, it needs to be stressed, that interpretable machine learning methods are useful to discover knowledge, to debug or justify, as well as control and improve models and their predictions, but not to make causal interpretations of the data.\cite{molnar2020conf} 

Future work could improve existing interpretability methods with large potential in the survival field, like survival counterfactuals, functional decomposition or Shapley values, as discussed in their respective sections in this paper. This paper primarily aims to synthesize, standardize and adapt existing IML methods to survival, therefore a deeper investigation into how the discussed quantities behave across a wide range of settings (e.g., varying effect strengths, sample sizes, evaluation metrics and censoring levels) in form of a dedicated benchmark study would offer a valuable avenue for future research. Additionally, we encourage researchers to rethink interpretability in the context of censored data or data with functional outcomes, developing novel methods addressing the limitations and challenges discussed in Section~\ref{sec:limits} and targeting the unique needs and requirements in these areas.  

\section*{Acknowledgements}
MNW was supported by the German Research Foundation (DFG), Grant Numbers: 437611051, 459360854. MK was supported from the SONATA BIS grant number 2019/34/E/ST6/00052 funded by the Polish National Science Centre (NCN). PB was supported by INFOSTRATEG-I/0022/2021-00 grant funded by Polish National Centre for Research and Development (NCBiR). Scientific work subsidized from the state budget within the Polish Ministry of Education and Science program ''Pearls of Science'' project number PN/01/0087/2022 (grantholder HB).

\section*{Conflict of Interest}
The authors declare no conflicts of interest.

\section*{Data Availability Statement}
The data that support the findings of this study are available from the Demographic and Health Surveys (DHS) Program. Restrictions apply to the availability of these data, which were used under license for this study. Data are available at \url{https://dhsprogram.com/} with the permission of the DHS Program.

\clearpage
\appendix

\section{Appendix}\label{app1}

\subsection{Simulating under a Weibull model with time-dependent effects for ICE, PDP and PFI}\label{app1:sim1}
Survival times are generated under a Weibull model with time-dependent effects for $N=3000$ observations. For a chosen observation from the dataset the hazard function is of the form:
\begin{align}\label{sim_pfi_ice_pdp}
h(t) = 1.5 \cdot 0.1 \cdot t^{(0.1 - 1)} \exp(-2.5 \cdot \texttt{treatment} + 5 \cdot t \cdot \texttt{treatment} + 0.7 \cdot \texttt{x}_1)
\end{align}
where $\texttt{Treatment} \sim \mathcal{B}(1, 0.5)$, $X_1 \sim \mathcal{N}(0, 1)$ and $X_2 \sim \mathcal{N}(0, 1)$. The generation of survival times $T^i$ follows the procedure outlined by Crowther and Lambert \cite{crowther2013}. First, the hazard function in Equation~\ref{sim_pfi_ice_pdp} is numerically integrated to obtain the cumulative hazard function, which is then transformed into the survival function using Equation~\ref{eq:survival-function}. Next, Brent's iterative root-finding method is applied to the equation $g(t) = S(t, \mathbf{x}) - U$, where $U \sim U(0, 1)$. The root of this equation represents the true (latent) survival time $T^i$ for a given vector of variables $\mathbf{x}^i$. Random censoring is simulated by drawing from a Binomial distribution $\Delta \sim \mathcal{B}(1, 0.2)$.
\begin{figure}[htb!]
  \centering
  \includegraphics[width = 0.9\linewidth]{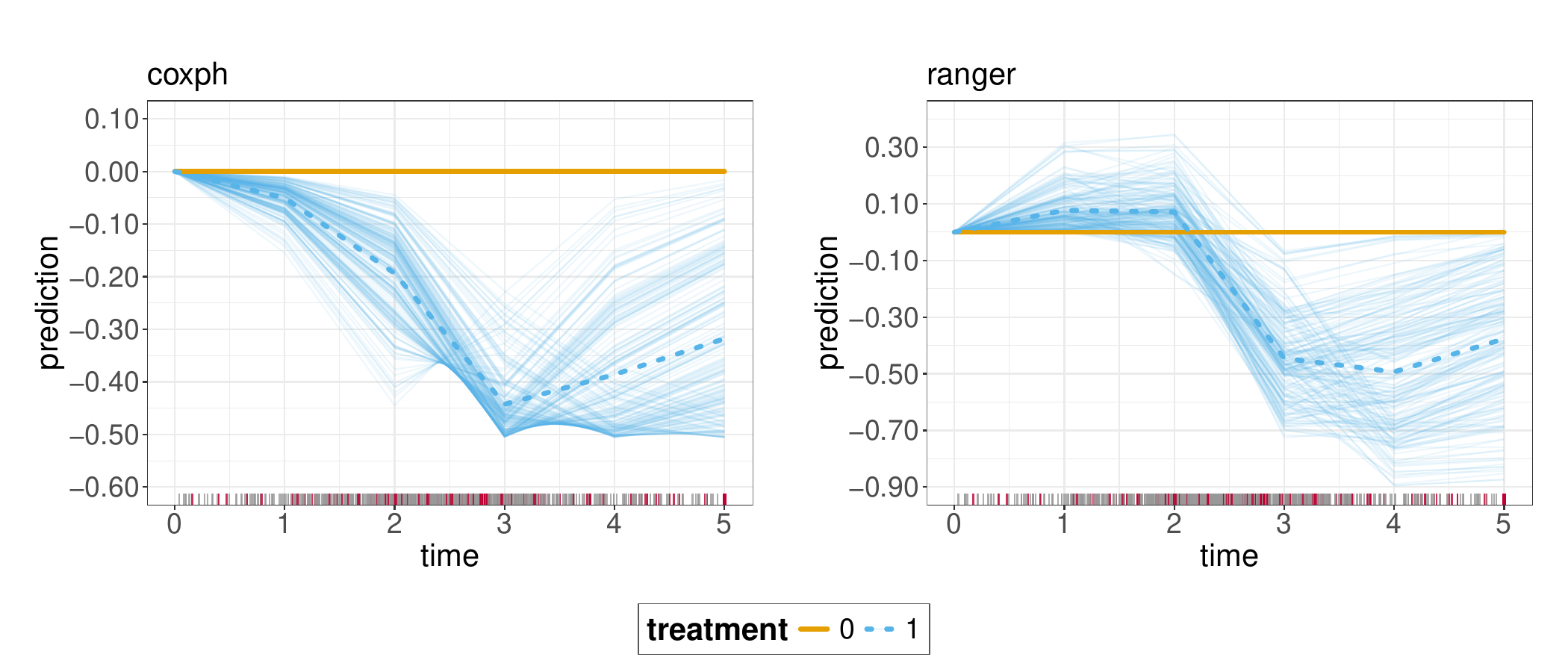}
  \caption{c-PDPs and c-ICE curves for the \texttt{coxph} model (left) and the \texttt{ranger} model (right) for the time-dependent treatment feature. ICE curves are depicted as thin colored lines, while PDPs are thick colored lines. The different line colors correspond to different treatment strategies (0 = no \texttt{treatment}, 1 = \texttt{treatment}). The PDPs depict the average predicted survival probability of patients having received \texttt{treatment} relative to those not having received treatment. The rug on the x-axis shows the survival time distribution with the grey bars indicating observed survival times and the red bars indicating censoring.}
  \label{fig:pdp_uc_sim}
\end{figure}
\begin{figure}[htb!]
\centering
\begin{subfigure}{0.4\textwidth}
    \subcaption{}
    \includegraphics[width=1.0\textwidth]{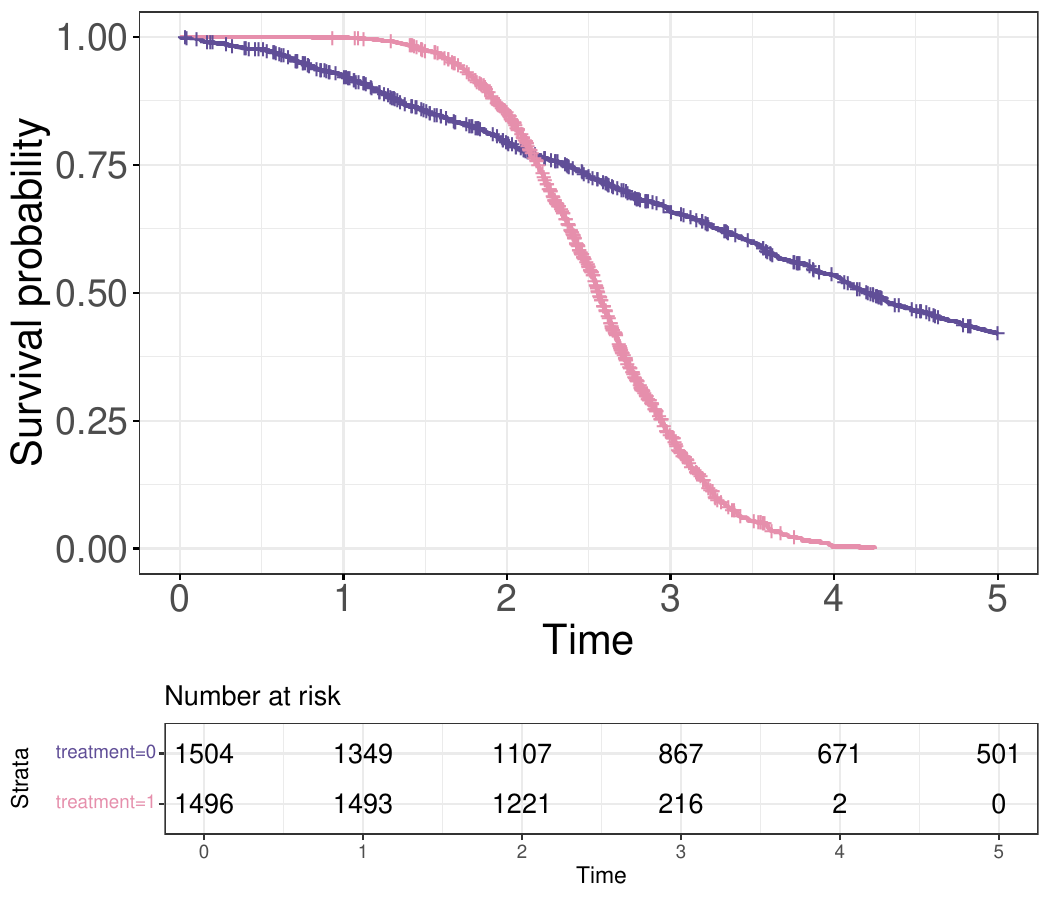}
    \label{fig:KM_sim}
\end{subfigure}
\qquad
\begin{subfigure}{0.4\textwidth}
    \subcaption{}
    \includegraphics[width=1.0\textwidth]{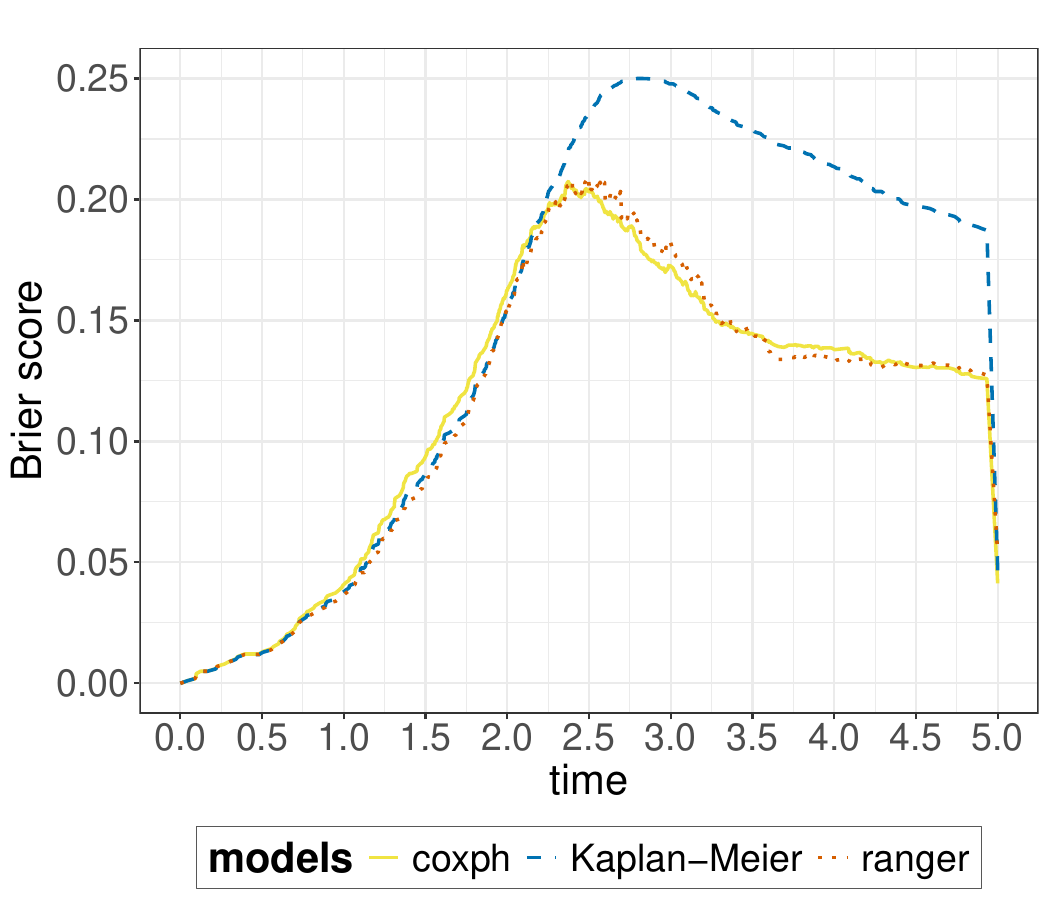}
    \label{fig:brier_sim}
\end{subfigure}
\caption{Introductory analysis of simulated dataset, (a) depicts the Kaplan-Meier survival curves and risk sets for different treatment groups (b) Brier score performance plot.}
\end{figure}

\subsection{Simulating under a Weibull model with time-independent, highly correlated features for PDP and ALE}\label{app1:sim2}

Survival times are generated under a Weibull model with time-dependent effects for $N=3000$ observations. For a chosen observation from the dataset the hazard function is of the form:
\begin{align}\label{sim_ale}
h(t) = 1.5 \cdot 0.1 \cdot t^{(0.1 - 1)} \exp(-4.5 \cdot \texttt{x}_1 + 4.5 \cdot \texttt{x}_2)
\end{align}
where $X \sim \mathcal{N}(0, 1)$, $X_1 \sim X + \mathcal{N}(0, 0.01)$ and $X_2 \sim X + \mathcal{N}(0, 0.01)$. The generation of survival times $T^i$ follows the procedure outlined by Crowther and Lambert \cite{crowther2013}. First, the hazard function in Equation~\ref{sim_pfi_ice_pdp} is numerically integrated to obtain the cumulative hazard function, which is then transformed into the survival function using Equation~\ref{eq:survival-function}. Next, Brent's iterative root-finding method is applied to the equation $g(t) = S(t, \mathbf{x}) - U$, where $U \sim U(0, 1)$. The root of this equation represents the true (latent) survival time $T^i$ for a given vector of variables $\mathbf{x}^i$. Random censoring is simulated by drawing from a Binomial distribution $\Delta \sim \mathcal{B}(1, 0.2)$.

\subsection{Simulating under an exponential survival model with time-independent features with interaction}\label{app1:sim3}

Survival times are generated under an exponential model with time-dependent effects for $N=3000$ observations. For a chosen observation from the dataset the hazard function is of the form:
\begin{align}\label{sim_inter}
h(t) = 0.1 \cdot t^{(0.1 - 1)} \exp(-0.5 \cdot \texttt{x}_1 - 0.5 \cdot \texttt{x}_2 + 3 \cdot \texttt{x}_1 \cdot \texttt{x}_2 - 0.5 \cdot \texttt{x}_4)
\end{align}
where $X_1 \sim \mathcal{N}(0, 1)$, $X_2 \sim \mathcal{N}(0, 1)$ and $X_3 \sim \mathcal{N}(0, 1)$. The generation of survival times $T^i$ follows the procedure outlined by Crowther and Lambert \cite{crowther2013}. First, the hazard function in Equation~\ref{sim_pfi_ice_pdp} is numerically integrated to obtain the cumulative hazard function, which is then transformed into the survival function using Equation~\ref{eq:survival-function}. Next, Brent's iterative root-finding method is applied to the equation $g(t) = S(t, \mathbf{x}) - U$, where $U \sim U(0, 1)$. The root of this equation represents the true (latent) survival time $T^i$ for a given vector of variables $\mathbf{x}^i$. Random censoring is simulated by drawing from a Binomial distribution $\Delta \sim \mathcal{B}(1, 0.2)$.

\newpage
\subsection{Data Application Example - Supplementary Information}\label{app1:data_example}

\begin{table}[htb!]
    \centering
        \caption{Description of the features selected for modelling the cancer recurrence of patients from the German Breast Cancer Study Group from 1984 to 1989. The feature type is either numerical (num) or categorical (cat). In the values column, for numerical features the observed ranges are given; for categorical features the different categories are denoted with their observed frequencies in brackets.}
    \begin{tabularx}{\textwidth}{llXX}
        \hline
        \textbf{Name} & \textbf{Type} & \textbf{Description} & \textbf{Values} \\
        \hline
        \texttt{horTh} & cat & Patient received hormonal therapy & yes (246); no (440)\\
        \texttt{age} & num & Age of the patient in years & 21 - 80\\
        \texttt{menostat} & cat & Menopausal status of the patient & Post(menopausal) (396); Pre(menopausal) (290)\\
        \texttt{tsize} & num & Size of the tumor in millimeters & 3 - 120\\
        \texttt{tgrade} & cat & Histological grade of the tumor & I (Low grade (well-differentiated)) (81); II (Intermediate grade (moderately differentiated)) (444); III (High grade (poorly differentiated)) (161)\\
        \texttt{pnodes} & num & Number of positive axillary lymph nodes & 1 - 51\\
        \texttt{progrec} & num &  Progesterone receptor levels in fmol & 0 - 2380\\
        \texttt{estrec} & num & Continuous variable measuring the estrogen receptor levels in fmol & 0 - 1144\\
        \hline
    \end{tabularx}
    \label{tab:features}
\end{table}

\begin{table}[htb!]
\centering
\caption{Observations used to compute \emph{SurvLIME} and \emph{SurvSHAP(t)} with corresponding feature values.}
\begin{adjustbox}{max width=\textwidth}
\begin{tabular}{lll}
\hline
\textbf{Individual} & \textbf{P1: Patient dead at t = 471} & \textbf{P2: Patient censored at t = 2539} \\
\hline
\texttt{time} & 471 & 2539 \\

\texttt{status} & 1 & 0 \\

\texttt{horTh} & no & yes \\

\texttt{age} & 80 & 47 \\

\texttt{menostat} & Post & Pre \\

\texttt{tsize} & 39 & 45  \\

\texttt{tgrade} & 2 & 2 \\

\texttt{pnodes} & 30 & 2 \\

\texttt{progrec} & 0 & 264 \\

\texttt{estrec} & 59 & 59 \\
\hline
\end{tabular}
\end{adjustbox}
\label{tab: lime_children}
\end{table}

\begin{figure}[t]
  \includegraphics[width = 1\linewidth]{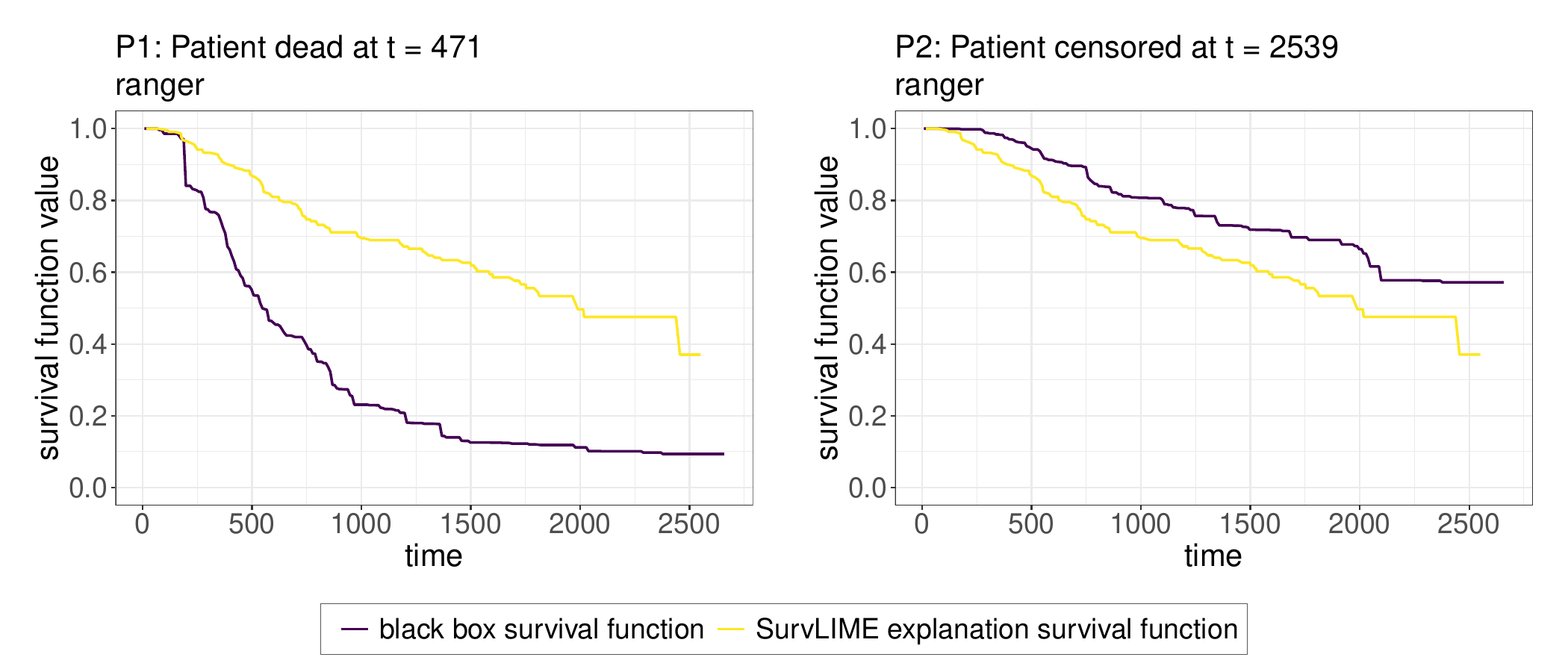}
  \caption{Black box survival functions predicted by the \texttt{ranger} model and \texttt{SurvLIME} explanation survival function for a patient dead at $t=471$ (left) and a patient censored at $t=2539$.}
  \label{fig:surv_grid}
\end{figure}

\clearpage
\bibliography{sample.bib}

\end{document}